\documentclass[10pt,reqno,twoside]{article}

\usepackage{natbib} 
    \renewcommand{\bibsection}{\subsubsection*{References}}
\usepackage{mathtools} 
\usepackage{booktabs} 
\usepackage{tikz} 
\usepackage[utf8]{inputenc}
\usepackage{authblk}
\usepackage{enumitem}  
\usepackage{amsthm}
\usepackage{amsmath}
\usepackage{bm}

\usepackage{setspace} 
\usepackage{comment}

\usepackage[colorinlistoftodos,prependcaption,textsize=tiny]{todonotes}
\allowdisplaybreaks

\usepackage{amssymb}
\usepackage{amsfonts}
\usepackage{comment}
\usepackage{colortbl}
\usepackage{graphicx}
\usepackage{placeins}
\usepackage{caption}
\usepackage{subcaption}

\newtheorem{innercustomgeneric}{\customgenericname}
\providecommand{\customgenericname}{}
\newcommand{\newcustomtheorem}[2]{%
  \newenvironment{#1}[1]
  {%
   \renewcommand\customgenericname{#2}%
   \renewcommand\theinnercustomgeneric{##1}%
   \innercustomgeneric
  }
  {\endinnercustomgeneric}
}

\newcustomtheorem{customthm}{Theorem}
\newcustomtheorem{customlemma}{Lemma}
\newcustomtheorem{customassump}{Assumption}

\usepackage{sectsty} 

\usepackage[margin=1.2in]{geometry}

\newcommand\fy[1]{{\color{magenta}Fanny: ``#1''}}
\newcommand\nr[1]{{\color{green}Nick: ``#1''}}

\newcommand{\zs}{\mathbf{Z}_s}
\newcommand{\zc}{\mathbf{Z}_c}

\newcommand{\tzc}{\mathbf{\tilde{Z}}_c}
\newcommand{\Y}{\mathbf{Y}}
\newcommand{\X}{\mathbf{X}}
\newcommand{\Z}{\mathbf{Z}}

\newcommand{\ProbXY}{\mathbb{P}_{\X,\Y}}
\newcommand{\ProbY}{\mathbb{P}_{\Y}}
\newcommand{\fhat}{\hat{f}}

\newcommand{\R}{\mathbb{R}}
\newcommand{\tildeZ}{\tilde{\Z}}

\newcommand{\tildeD}{\tilde{D}}

\newcommand{\ProbXcY}{\mathbb{P}_{\X|\Y}}
\newcommand{\pstarX}{p^\star({\X})}
\newcommand{\pstarXcY}{p^\star({\X|\Y})}
\newcommand{\id}{\mathrm{Id}}
\newcommand{\ktrue}{{k}^\star}

\DeclareTextFontCommand{\bgreen}{\color{green}\bfseries}
\DeclareTextFontCommand{\bred}{\color{red}\bfseries}
\DeclareTextFontCommand{\byellow}{\color{yellow}\bfseries}

\newcommand\independent{\protect\mathpalette{\protect\independenT}{\perp}}
\def\independenT#1#2{\mathrel{\rlap{$#1#2$}\mkern2mu{#1#2}}}
\usepackage{amsmath}
\newcommand{\methName}{\emph{CLAP}}
\newcommand{\suppmat}{Supp. Mat.}
\newcommand{\oursection}{Sec.}

\newcommand{\distequal}{\stackrel{\text{dist}}{=}}

\newcommand{\chest}{Chest X-ray}

\usepackage{titlesec}
\titlespacing*{\section}{0pc}{5pt}{5pt}
\titlespacing{\section}{0pc}{5pt}{5pt}
\titlespacing{\subsection}{0pc}{5pt}{5pt}
\titlespacing{\subsubsection}{0pc}{5pt}{5pt}
\DeclareMathOperator*{\argmax}{argmax} 
\DeclareMathOperator*{\argmin}{argmin} 

\title{Provable concept learning for interpretable predictions using variational autoencoders
}

\author[1]{Armeen Taeb}
\author[2 3]{Nicol\`o Ruggeri}
\author[4]{Carina Schnuck}
\author[3]{Fanny Yang}
\affil[1]{%
   Seminar for Statistics\\
    ETH\\
    Z\"urich, Switzerland
}
\affil[2]{%
    Max-Planck-Institute for Intelligent Systems\\
    T\"ubingen, Germany
}
\affil[3]{%
    Computer Science Dept.\\
    ETH\\
    Z\"urich, Switzerland
}
\affil[4]{%
    Mathematics Dept.\\
    ETH\\
    Z\"urich, Switzerland
}

\usepackage[
            CJKbookmarks=true,
            bookmarksnumbered=true,
            bookmarksopen=true,
            colorlinks=true,
            citecolor=blue,
            linkcolor=black,
            anchorcolor=red,
            urlcolor=blue,
            ]{hyperref}
\usepackage{cleveref}

\date{}
\begin{document}
\maketitle
\begin{abstract}

In safety-critical applications, practitioners are reluctant to trust neural networks when no interpretable explanations are available. Many attempts to provide such explanations revolve around pixel-based attributions or use previously known concepts. In this paper we aim to provide explanations by provably identifying \emph{high-level, previously unknown ground-truth concepts}. To this end, we propose a probabilistic modeling framework to derive (C)oncept (L)earning and (P)rediction (CLAP) -- a VAE-based classifier that uses visually interpretable concepts as predictors for a simple classifier. Assuming a generative model for the ground-truth concepts, we prove that \methName{} is able to identify them while attaining optimal classification accuracy. Our experiments on synthetic datasets verify that CLAP identifies
  distinct ground-truth concepts on synthetic datasets and yields promising results on the medical Chest X-Ray dataset.

\end{abstract}
\section{Introduction}
\label{sec:introduction}
Suppose a hospital aims to deploy a model that classifies diseases $\Y$ from medical images $\X$ and informs the doctor about relevant predictive features. There may be multiple diseases such as lung atelectasis and lung infiltration and multiple \emph{interpretable} ground-truth \emph{features} (or \emph{concepts}) $\Z_c$, such as lung or heart shape, that are relevant for predicting each disease. Ideally, in addition to identifying and utilizing these interpretable features, the model should perform prediction in an interpretable manner itself. The domain expert can then check whether the model is reasonable and also potentially make new scientific discoveries -- i.e. discover new factors relevant for prediction.

Thus, in this paper, we seek an interpretable predictive model that uses the ground-truth features for prediction. But what makes a predictive model interpretable from a practical perspective? 
Even though the definite answer depends on the application domain, practitioners often agree on the following desiderata:
first of all, the model should be \emph{simple} -- e.g. additive in the predictive features with a small number of relevant features. Simplicity allows us to interpret the relevance of each variable \citep{Rudin2018PleaseSE}, and ensure that the  interpretation is robust to small changes to the input \citep{AlvarezMelis2018OnTR,alvarez2018towards}. Furthermore, the model ideally assigns \emph{global and local} importance to the features used for prediction \citep{Reyes2020OnTI,tiglic2020InterpretabilityOM}; 
in the context of medical imaging for example, the former corresponds to the population-level importance, the latter to the patient-level one.


While there have been many works on interpretable predictions, none of them provide a prediction model that identifies and uses these previously unknown ground-truth features (see relate works for more discussion). 
This paper tries to go bottom-up, starting from a generative model to derive a procedure based on variational inference that satisfies all the desiderata. Our proposed framework i) mathematically formalizes concept learning and ii) provably identifies the ground-truth concepts and provides an accurate and simple prediction model using these discovered concepts. 

More concretely, we view the recovery of the ground-truth concepts as a latent variable estimation problem.
We start by assuming an explicit graphical model for the joint distribution of $(\X,\Z,\Y)$. 
Here, the latent variables $\Z$ include all ground-truth latent features, as well as others irrelevant for prediction. Together, the latent variables $\Z$ generate the raw observation $\X$. 
The task of concept learning can then be mathematically thought of as obtaining identifiability and performing inference on the latent factors. 
Using a VAE-based architecture, we enable both visualization (and thus facilitate human interpretation) of the learned concepts, as well as prediction based on these.
In summary, we make the following contributions:

\begin{enumerate}
 \item  We present a framework to model ground-truth latent features $\zc$ (\emph{\oursection{}~2}), and derive C(oncept) (L)earning and (P)rediction (in short \methName{}), an inherently interpretable prediction framework based on variational autoencoders 
 (\emph{\oursection{}~3})
 \item We prove that \methName{} enables identification of the ground-truth concepts underlying the data and learns a simple optimal prediction model based on these. Importantly, our framework does not require knowing the number of latent features
 (\emph{\oursection{}~4})
 \item We validate \methName{} on various multi-task prediction scenarios on synthetic (MPI3D, Shapes3D and SmallNorbs) datasets that yield encouraging results 
 on domain-specific application of the framework on real data 
 (\emph{\oursection{}~5})
\end{enumerate}
 
We believe that our theoretical framework  is a useful step for formalizing interpretable predictions. In particular, 
         in settings where it's reasonable to assume that the ground-truth features are themselves
         interpretable by a domain expert, \methName{} provably provides an end-to-end interpretable prediction model. 
         Even when the assumption does not hold,
         we can still guarantee that \methName{} finds a simple and accurate prediction model using ground-truth features.

\vspace{-0.1in}
\subsection{Related work}
\label{sec:related_work}
In this section, we compare existing interpretable prediction methods with \methName{} in detail, with a concise summary provided in Table~\ref{table:comparisons}.
Previous methods proposed in the context of explainable/interpretable AI can be broadly divided into two categories:  (i) providing post-hoc explanations for black-box prediction models and (ii) designing interpretable models that explicitly incorporate transparency into the model design, where the explanation is learned during training.
\begin{table*}[htp]
\centering
\resizebox{0.95\textwidth}{!}{%
\begin{tabular}{|c|c|c|c|c|c|}
\hline
 &  \multicolumn{3}{|l|}{Post-hoc explanations} & \multicolumn{2}{|l|}{Inherently interpretable}\\\hline
 \shortstack{{}\\Desiderata\\{}}&\shortstack{{}\\pixel attribution+ \\counterfactual}&\shortstack{{}\\pre-defined\\{concepts}} &\shortstack{{}\\StyleGANs\\{}}&\shortstack{{}\\existing VAEs/ \\ autoencoders}&\shortstack{{}\\\methName{}\\{}}
 \\\hline
 \shortstack{{}\\Learning {visually distinct features}} & \shortstack{{}\\\large{$\bm{\times}$}\\{}} & \shortstack{{}\\\large{{$\bm{\times}$}}\\{}} & \shortstack{{}\\\large{\checkmark}\\{}} & \shortstack{{}\\\large{ ${\checkmark}^\star$}\\{}} & \shortstack{{}\\\large{\checkmark}\\{}} \\\hline
 \shortstack{{}\\Global importance of {predictive features}} & \shortstack{{}\\\large{$\bm{\times}$}\\{}} & \shortstack{{}\\\large{\checkmark}\\{}} & \shortstack{{}\\\large{$\bm{\times}$}\\{}} & \shortstack{{}\\\large{$\bm{\times}$}\\{}} & \shortstack{{}\\\large{\checkmark}\\{}} \\\hline
 \shortstack{{}\\Guarantees: concept {learning+prediction}} & \shortstack{{}\\\large{$\bm{\times}$}\\{}} & \shortstack{{}\\\large{{$\bm{\times}$}}\\{}} & \shortstack{{}\\\large{$\bm{\times}$}\\{}} & \shortstack{{}\\\large{$\bm{\times}$}\\{}} & \shortstack{{}\\\large{\checkmark}\\{}} \\\hline
\end{tabular}
}
\caption{\small{Comparison of \methName{} with post-hoc explanation methods and other inherently interpretable techniques. The symbol ${\checkmark}^\star$ highlights that for learning visually distinct features, existing predictive VAEs require strong knowledge of the latent variables or auxiliary variables (in addition to labels).}}
\label{table:comparisons}
\end{table*}

{\bf{Post-hoc methods}} The majority of work on interpretability so far has focused on (i), providing post-hoc explanations for a given prediction model. 
These include pixel attribution methods \citep{Bach15Pixel,Selavarju17,Simonyan2014DeepIC}, counterfactual explanations \citep{antoran2021getting, Chang2019ExplainingIC}, explanations based on pre-defined concepts \citep{Kazhdan2020NowYS,pmlr-v32-rezende14,Yeh2020OnCompleteness}, and recently developed StyleGANs \citep{Lang2021ExplainingIS,Wu2021StyleSpaceAD}. Post-hoc methods have a number of shortcomings given our desired objectives: First, it is unclear whether post-hoc explanations indeed reflect the black-box model's true "reasoning" \citep{Kumar2020ProblemsWS,Rudin2018PleaseSE}. Even if an expert deems the output of the explanation model as unreasonable, one is unable to determine whether the explanation method or the original model is at fault. Furthermore, by construction, post-hoc methods cannot come with statistical inference guarantees and ensure that the learned concepts align with the ground-truth features. Finally, post-hoc methods are typically used to explain complex classifiers; as a result, they are unable to provide meaningful global and local importance of features for prediction.

{\bf{VAE-based methods for inherently interpretable prediction}} 
Our procedure \methName{} is an inherently interpretable prediction model and similar in spirit to VAE-based prediction techniques.
On a high level, existing procedures either are unable to identify the ground-truth latent features or require additional
labels. Therefore, they are not applicable in the traditional supervised learning setting considered in this paper (where only $\X,\Y$ are available). Further, none of the existing methods provide simultaneous guarantees for learning the underlying concept and obtaining optimal predictions using these learned features. We provide more specific comparisons next.

Unsupervised VAEs \citep{KingmaSemi} can easily be used for prediction tasks by training a classifier on the latent features. A massive literature proposes various structural adjustments to improve disentanglement \citep{AnnealedVAE,chen2018isolating,Higgins2017betaVAELB,kim2019disentangling,Kumar17Disentangle}.
However, \cite{Locatello-challenges} empirically and theoretically demonstrate that these methods generally do not successfully identify the ground-truth latent features.
Recently proposed VAE methods address the issue of non-identifiability by assuming access to additional data and improve identifiability. However, they either require the label as direct input \citep{ICLR-VAE2021}, or labels for auxiliary variables that contain information about the ground-truth latent factors \citep{VAE-ICA,ID-VAE2021} or the ground-truth factors themselves \citep{Locatello-challenges}. None of these scenarios are applicable to the traditional supervised learning setting in our paper. 

{\bf{Other works}}
With respect to model architecture, our method is similar to Self-Explaining Neural Networks (SENN) \citep{alvarez2018towards} which decomposes a complex prediction model into learning interpretable concepts (using an autoencoder) and a simple (linear) predictor. 
More broadly, methods based on contrastive learning or multi-view data (e.g. \citep{Gresele2019TheIR,Hyvrinen2019NonlinearIU,Locatello_weakly_supervisedd,Shu2020WeaklySD,vonKgelgen2021SelfSupervisedLW}) can identify underlying latent features, albeit with access to pairs of images that share similar sources. Furthermore, the focus of these methods is on representation learning rather than interpretable predictions.
\section{Modeling interpretable and predictive concepts}
\label{sec:model}
We present a probabilistic graphical model that statistically relates the ground-truth latent features $\zc$ to the labels and observed variables; our proposed method later uses this model to learn the latent concepts as well as a simple classifier based on these features. We remark that, although the methodology in this paper is presented under a specific generative model, the framework is general and flexible to other modeling choices. 

Let $\X$ be raw observations and $\Y \in \mathcal{Y}$ be the associated label vector taking a finite collection of values. In general, $\X$ is comprised of \emph{style factors} $\zs$, that should not be relevant for prediction, and high-level \emph{core factors} $\zc$ that are the desired ground-truth concepts. For example, in the context of medical imaging, $\Y$ are various disease labels such as the presence of lung atelectasis and lung infiltration. Core factors $\zc$ that one can see in the X-ray image $\X$, such as heart and lung shapes, are typically direct consequences of a patient contracting the disease. Style factors $\zs$ such as physiological characteristics of the subject or specialities of the scanner are also factors that appear in the image but are not related to the disease.

\begin{figure}[t]
\vspace{-1cm}
\centering
\begin{subfigure}[b]{0.43\linewidth}
    \includegraphics[width=\textwidth]{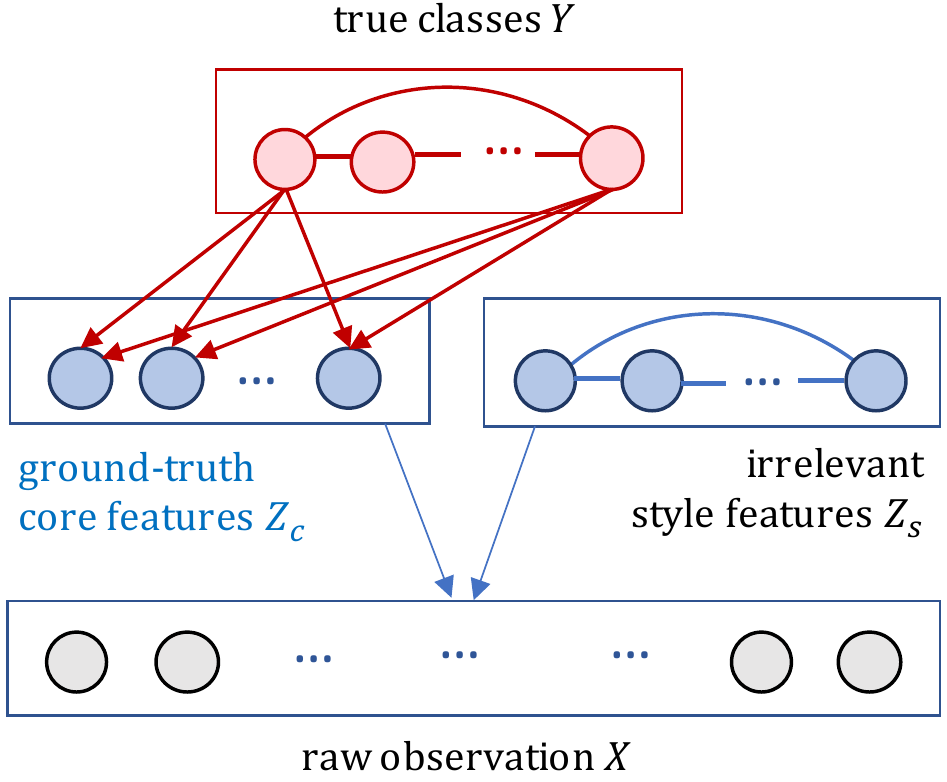}
    \caption{Generative graphical model}
    \label{fig:graph_models}
\end{subfigure}
\begin{subfigure}[b]{0.43\linewidth}
    \includegraphics[width=\textwidth]{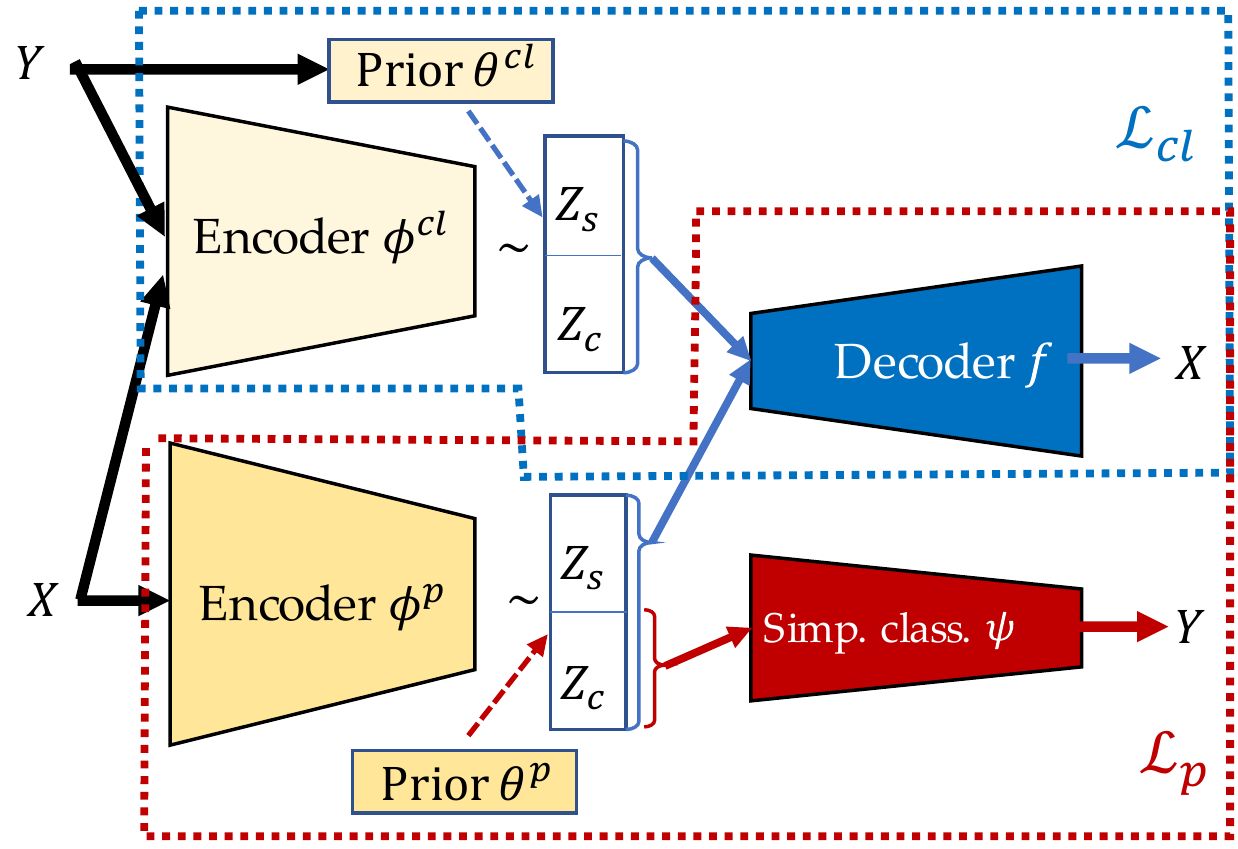}
    \caption{VAE architecture of \methName}
    \label{fig:VAE_architecture}
\end{subfigure}
\caption{\small{The graphical model in (a) describes how the desired high-level core latent features $\zc$ are related to the remaining variables $\Y,\X,\zs$. The VAE architecture in (b) is derived by lower-bounding the evidence values $p(\X, \Y)$ and $p(\X |\Y)$ and incorporating the generative assumptions from (a) (see main text). We utilize two separate encoders, correspondent to the $\mathcal{L}_{cl}$ and $\mathcal{L}_{p}$ terms of objective \eqref{eqn:CbPVAE}, and impose sharing of the decoder. The two encoders define two different sets of latents $\Z=(\zc, \zs)$, which are separately passed through $f$ to get the relative reconstructions. The two resulting objectives $\mathcal{L}_p$ and $\mathcal{L}_{cl}$ are then summed in the full objective $\mathcal{L}_{\methName}$. A simple classifier based on $\zc$ is trained as part of the model inside $\mathcal{L}_p$.}}
\label{fig:overall}
\vspace{-0.2in}
\end{figure}

A natural model for settings such as the one above is to assume an \emph{anti-causal} model as in Fig.~\ref{fig:graph_models}, where $\zc$ is a child of $\Y$, and combines with $\zs$ to produce the raw observation $\X$. We assume $\zc$ to be independent conditionally on $\Y$, as in the X-ray example, they may often vary independently (across patients) given a disease label.
We instead allow arbitrary dependencies within $\zs$ and $\Y$.

 Aggregating style and core factors in the vector $\Z = (\zc,\zs)$, we impose the following structural equation model on the graph in Fig.~\ref{fig:graph_models}:
\begin{equation}
\begin{aligned}
&\X = f^\star({\Z}) + \epsilon~~\text{where }\epsilon \independent \Z,\Y \text{ and for all }y \in \mathcal{Y}:\\
&\Z|\Y\hspace{-0.03in}=\hspace{-0.03in}y \sim \mathcal{N}\left(\begin{pmatrix}\mu^\star_y \\\mu^\star \end{pmatrix},\begin{pmatrix}D^\star_{y} & 0 \\ 0 & G^\star \end{pmatrix}\right); D^{\star}_{y}\hspace{0.03in}\text{diagonal} \, ,
\end{aligned}
        \label{eqn:model}
\end{equation}
for some continuous one-to-one function $f^\star$, vectors $\mu^\star_y,\mu^\star$, and positive-definite matrices $D_y^\star,G^\star$. 
The model \eqref{eqn:model} encodes the conditional independence relationships in Fig.~\ref{fig:graph_models}: the covariance of the distribution $\zc|\Y$ is diagonal; 
the mean and covariance corresponding to $\zs$ are not a function of $y$ and the noise $\epsilon$ is independent of $\Y$ so that $\X \independent \Y | \zc$ and $\zs\independent\Y$.



\vspace{-0.1in}
\section{\methName{}: interpretable predictions using ground-truth\\ concepts}
\label{sec:interpretablepredictions}
Given data of $\X$ and $\Y$ arising from the graphical model in Fig.~\ref{fig:graph_models}, our objective is to identify the ground-truth concepts and learn a simple classifier that uses these to accurately predict $\Y$. Additionally, to facilitate human interpretability, we aim to enable experts in the loop to visually interpret the learned concepts. For concreteness, we specialize our exposition to images, although our framework can in principle be used on other types of data. 

Our proposed framework is based on variational autoencoders (VAEs) \cite{KingmaSemi,pmlr-v32-rezende14}. VAEs offer a number of favorable properties for our objectives. First, they can be derived in a principled manner from the underlying data generating mechanism. Second, the encoder/decoder pair in VAEs provide an effective approach to visualize and thus interpret the learned latent features via latent traversals (see \oursection~\ref{sec:stage2} for more details). 

{In that light, a natural first approach that might come to mind would be to train a VAE that uses the estimated latent features for prediction. In \oursection{}~\ref{sec:vanilla} we derive such a model, and show why, in its vanilla version, it can perform prediction but cannot identify the ground-truth core concepts.}
In \oursection{}~\ref{sec: clap_as_solution}, we overcome these challenges by introducing a novel VAE architecture \methName{} shown in Fig.~\ref{fig:VAE_architecture}. Our proposed method combines the predictive VAE structure from earlier with a second VAE which helps with identifying the underlying ground-truth concepts.



\subsection{Vanilla predictive VAE and its shortcomings}
\label{sec:vanilla}


A natural first attempt at learning a predictive VAE procedure is to maximize the following ELBO of the log-evidence of $(\X, \Y)$: 
\begin{equation}
\label{eqn:predVAE}
\log{p}(\X,\Y) 
    \ge \mathbb{E}_{q_{{\phi}^p}(\Z|\X)}\log \frac{p_{f}(\X|\Z)p_\psi(\Y|\Z_c)p_{{\theta}^p}(\Z)}{q_{\phi^p}(\Z|\X)} 
    =: \mathcal{L}_p(\phi^p,{\theta}^p,f,{\psi};\X,\Y) \, .
\end{equation}
The objective $\mathcal{L}_p$ corresponds to the VAE architecture in the red box in Fig.~\ref{fig:VAE_architecture}. Here, $q$ is the approximate posterior with encoder parameters $\phi^p$, $\psi$ parameterizes a simple classifier, $f$ is the decoder's parameters, and $\theta^p$ the prior distribution's parameters. Specifically, from the data generating mechanism \eqref{eqn:model}, the prior $p_{\theta^p}(\Z)$ is a density of a Gaussian mixture distribution with $|\Y|$ (number of labels) components, where the covariance corresponding to the core features for each mixture component is diagonal. 
The ELBO~\eqref{eqn:predVAE} is derived in a classical fashion by using Jensen's inequality $
\log{p}(\X,\Y) \ge \mathbb{E}_{q(\Z|\X, \Y)}\log \frac{p(\X, \Y|\Z) p(\Z)}{q(\Z|\X, \Y)}$ 
and leveraging the assumed generative model \eqref{eqn:model} to simplify the right-hand side.

The model learned by maximizing the objective $\mathcal{L}_p$ naturally yields a classifier $p_\psi(\Y|\Z_c)$ based on core features extracted from the encoder $q_{\phi^{p}}(\Z|\X)$, which should approximate the ground-truth ones. Since the encoder does not rely on $\Y$ as an input, {we can readily use it for end-to-end classification during test time.}
In fact, under a regularity condition, we show in \suppmat{
} \oursection{} \ref{sec:proof_lemma_predictive} that this architecture is optimal for prediction. However, it \emph{does not} guarantee that the estimated core features $\hat{\Z}_c$ correspond to the ground-truth factors $\zc$. 
In fact, they can be arbitrary linear transformations of $\zc$ 
without sacrificing prediction performance \cite{Locatello-challenges} (see ablation studies in \oursection{} \ref{sec:experiments}), thus not satisfying our desired properties. 
In addition, as the dimensionality of the core features $\zc$
is typically unknown, a conservative choice for the number of latent features (over-parameterized setting)
may wrongly include style features or redundant core features in the prediction model (see ablation study in \oursection~\ref{sec:experiments}). 
In the next section, we propose our framework \methName{} that mitigates the {aforementioned issues}: it learns a prediction model using the ground-truth core concepts (even in the over-parameterized setting), without sacrificing classification accuracy. 
\vspace{-0.07in}
\subsection{CLAP to overcome shortcomings}
\label{sec: clap_as_solution}
To overcome the aforementioned challenges, we augment the objective $\mathcal{L}_p$ with two additional terms to arrive at our proposed objective function for \methName:
\begin{equation}
\mathcal{L}_{\methName} := \mathcal{L}_p + \mathcal{L}_{cl} - \lambda_n \rho \,.
\label{eqn:CbPVAE}
\end{equation}
On a high level, the additional component  $\mathcal{L}_{cl}$ ensures 
identifiability of the ground-truth concepts $\zc$ (\emph{concept learning})
and 
the regularization term $\lambda_n\rho$ 
helps to identify a minimal number of ground-truth concepts in an over-parameterized latent space.
In the following, we formalize each term.

\textbf{Concept-learning component $\mathcal{L}_{cl}$} While the objective $\mathcal{L}_p$ is designed to maximize the full likelihood of image data $\X$ and target labels $\Y$, the term $\mathcal{L}_{cl}$ maximizes the likelihood of $\X$ conditioned on $\Y$.
The fact that the labels act as additional input data in this likelihood objective, plays a central role in provably obtaining identifiability.
Furthermore, the conditional independence of $\zc$ given $\Y$ 
can be more naturally captured when $\Y$ is considered as an input.  
Similarly to above, for any posterior $q$, we can lower-bound the conditional log-evidence as 
$
    \log{p}(\X|\Y) \geq \mathbb{E}_{q(\Z|\X,\Y)}\log\frac{p(\X|\Z,\Y)p(\Z|\Y)}{q(\Z|\X,\Y)} \, ,
$
and incorporate the generative assumptions in \eqref{eqn:model} to obtain the final ELBO objective: 
\begin{equation}
\label{eqn:cbVAE}
\log p(\X|\Y) \geq \mathbb{E}_{q_{\phi^{cl}}(\Z|\X,\Y)}\log\frac{p_f(\X|\Z)p_{{\theta^{cl}}}(\Z|\Y)}{q_{{\phi^{cl}}}(\Z|\X,\Y)} 
    :=\mathcal{L}_{cl}({\phi}^{cl},{\theta}^{cl},f;\X,\Y) \, .
\end{equation}
The component of \methName{} corresponding to $\mathcal{L}_{cl}$ is highlighted in blue in Fig.~\ref{fig:VAE_architecture}. Here, $\phi^{cl}$ are the parameters of the encoder, and $f$ those of the decoder. Appealing to the data generating mechanism \eqref{eqn:model},
we can further factorize the prior in the form $p_{{\theta^{cl}}}(\Z|\Y) = p(\zc|\Y)p(\zs)$. Here, $p(\zc|\Y)$ is a Gaussian density function with diagonal covariance and different parameters for different $\Y$ while we model the prior $p(\zs)$ as a standard Gaussian distribution without loss of generality. We aggregate all these parameters in $\theta^{cl}$.

In general, maximizing the ELBO or even the true log-evidence would not allow for of identification the true concepts. However, a simple
heterogeneity assumption can alleviate this issue,
formally stated in \suppmat{} \oursection{} \ref{sec:assumptions}.
\begin{customassump}{1}[Concept learning, informal]
\label{ass:cl} The functions $f,f^\star$ satisfy a regularity condition and the distribution of core features change `enough' when conditioned on different realizations of $\Y$.
\end{customassump}
\vspace{-0.07in}
We now utilize these assumptions to prove the following result.
\begin{customlemma}{1}[Maximizing $\mathcal{L}_{cl}$ identifies the ground-truth concepts]  Suppose the data is generated according to the model in \eqref{eqn:model} with no noise, i.e. $\epsilon \equiv 0$ and Assumption~\ref{ass:cl} holds. Suppose $\mathcal{L}_{cl}$ is maximized in the infinite data limit with the correct number of latent features included in the model. Then, the posterior samples $\hat{\Z}_c$ obtained from the encoder $q_{\hat{\phi}^{cl}}$ are equal to the ground-truth features $\zc$ up to permutation and scaling. 
\label{lemma:cl}
\end{customlemma}
We prove this lemma in \suppmat{} \oursection{}~\ref{sec:proof_lemma_concept}, and also extend to the noisy setting in \suppmat~\oursection{}~\ref{thm:noisy}. Theoretical results for identifibiality were previously established in \cite{VAE-ICA}. We note that our guarantees differ substantially and refer to \suppmat{} \oursection~\ref{sec:comparison_with_khemakhem} for more details.
Despite the concept-learning capabilities, a model trained only on $\mathcal{L}_{cl}$ \emph{cannot} be used for prediction since it requires the labels as input to the encoder $q_{\phi^{cl}}(\Z | \X, \Y)$. 

Therefore, we combine the objectives $\mathcal{L}_p$ and $\mathcal{L}_{cl}$ by utilizing the {same decoder} $f$ in \eqref{eqn:predVAE} and \eqref{eqn:cbVAE}, as represented in Fig.~\ref{fig:VAE_architecture}.
This coupling via a shared decoder is crucial, as it forces the $\mathcal{L}_p$ architecture to also perform concept learning.
To see why, first note that in joint training, the two encoders of $\mathcal{L}_{cl}$ and $\mathcal{L}_{p}$ learn approximately the same latent space. In fact, we show in Theorem~\ref{thm:PDVAE} that the latent spaces align in the infinite data limit. \footnote{Informally speaking, the reason for this is that the latent features in each architecture reconstruct the image via the same decoder. Since the common decoder defines a generative model , the posteriors (i.e. the different encoders) need to be similar as well.}Since $\mathcal{L}_{cl}$ provably identifies
the ground-truth features in the latent space, it then follows that 
the estimated core features obtained by the encoder of $\mathcal{L}_p$ closely align with $\zc$. 
Thus, after training the combined objective $\mathcal{L}_{p} + \mathcal{L}_{cl}$, the trained VAE architecture corresponding to $\mathcal{L}_p$ provides an interpretable prediction model: an input image is mapped to accurate ground-truth core features, which are then used on top of a simple classifier to predict the target label $\Y$. We refer the reader to \oursection~\ref{sec:stage2} for more discussion on how the trained \methName{} is used at test time.

\textbf{Sparsity penalty $\rho$ to account for overparameterized latent space} 
We add a regularization term $\lambda_n\rho(f,\psi)$ to impose simultaneous group sparsity on the prediction weights and decoder weights -- this ensures that if an estimated core feature feature is predictive, it has non-negligible effect in the reconstruction of the image and vice versa. In particular, let ${k}_c,{k}_s$ be the conservative choice on the dimensionality of the core and style features in our VAE model, respectively. Further, let ${k} = {k}_c + {k}_s$ be the total number of latent variables. We consider the following parameterization for the decoder $f = f' \circ B,\ B \in \mathbb{R}^{{k} \times {k}}$ and classifier $\psi = \psi' \circ C,\ C \in \mathbb{R}^{k_c \times {k}_c}$, where $|\Y|$ is the number of labels to be predicted and $f',\psi'$ are one-to-one and continuous. Then, the sparsity inducing penalty $\rho(f,\psi)$ {in the combined objective function \eqref{eqn:CbPVAE}}
takes the form:
\begin{equation}
    \rho(f,\psi) := \sum_{i = 1}^{{k}_c} \mathbb{I}\left[\left\|\begin{pmatrix}B_{:,i}^T & C_{:,i}^T\end{pmatrix}\right\|_2 > 0 \right] + \sum_{i = {k}_c+1}^{{k}}\mathbb{I}\left[\left\|B_{:,i}^T\right\|_2 > 0 \right],
    \label{eqn:rho_defn_theoretical}
\end{equation}
where the indicator function $\mathbb{I}[\cdot]$ counts the number of latent features effectively utilizes by the model. {Note that the nonzero columns of $C$ correspond to core features in the model with predictive power, and the nonzero columns of $B$ correspond to core and style features that are used for reconstruction with the decoder $f$.} For practical considerations, we consider the following convex surrogate {in our experiments}: $\rho(f,\psi) = \sum_{i = 1}^{{k}_c} \left\|\begin{pmatrix}B_{:,i}^T & C_{:,i}^T\end{pmatrix}\right\|_2$.


\vspace{-0.07in}
\subsection{Theoretical guarantees for \methName{}} 
\label{sec:theory}
In \oursection{}~\ref{sec: clap_as_solution}, we described how after the training of \methName{}, the component corresponding to $\mathcal{L}_p$ can be used as an interpretable prediction model. We next provide guarantees that this prediction model is optimal {in terms of accuracy} and is based on high-level features that align with the ground-truth concepts. In the sequel, we denote ${k_c, k_s}$ to be the number of core and style features chosen in the VAE architecture and $\ktrue_c, \ktrue_s$ to be the dimensions of the true features of the generative model in Fig.~\ref{fig:graph_models}. Further, we use $q_{\hat{\phi}^{p}}$, $q_{\hat{\phi}^{cl}}$ to denote the encoders obtained by maximizing the objective in \eqref{eqn:CbPVAE} in the infinite data limit and let $\hat{\Z}$ be the posterior samples obtained from $q_{\hat{\phi}^{p}}$. {Finally, we denote the trained classifier as $\hat{\psi} = \hat{\psi}' \circ \hat{C}$, and the core features $\hat{\Z}_c$ 
are specified as the elements corresponding to nonzero columns of $\hat{C}$}. 

Our theory requires Assumption \ref{ass:cl} for concept learning as well as an assumption about a simple classifier being optimal:
\begin{customassump}{2}[optimal classifier] The Bayes optimal classifier for predicting $\Y$ using $\zc$ belongs to the set of simple classifiers used in \methName{}.
\end{customassump}
\vspace{-0.07in}
We utilize Assumptions 1 and 2 to prove the following result.

\begin{customthm}{1}[\methName{} learns an optimal prediction model using interpretable ground-truth features] Consider the same setup as Lemma~\ref{lemma:cl}. Suppose $k_c \geq \ktrue_c$, $k_s \geq \ktrue_s$, and that Assumptions 1 and 2 hold. Then, the posterior samples $\hat{\Z}$ obtained from the encoder $q_{\hat{\phi}^p}$ are identical to the posterior samples obtained from the encoder $q_{\hat{\phi}^{cl}}$. Furthermore, 
the core features $\hat{\Z}_c$ are 1) optimally predictive: $\Y|\hat{\Z}_c\stackrel{\text{dist}}{=} \Y|\X$, and 2) aligned with the ground truth: $\hat{\Z}_c$ is equal to $\zc$ up to scaling and permutation. 
\label{thm:PDVAE}
\end{customthm}
\vspace{-0.07in}
The proof of Theorem~\ref{thm:PDVAE} is presented in \suppmat{}  \oursection~\ref{sec:proof_corollary}. Our guarantees in Theorem~\ref{thm:PDVAE} ensure that the prediction model obtained by \methName{} is optimal. Furthermore, the core features $\hat{\Z}_c$ align with the ground-truth concepts. Finally, the number of predictive factors equals to the number of ground-truth concepts; that is, our model obtains the minimal set of predictive features. 


\vspace{-0.07in}
\subsection{Visualizing and evaluating \methName{}'s output for interpretation}
\label{sec:stage2}
\begin{figure}[t]
    \vspace{-1cm}
    \centering
        \includegraphics[width=0.9\textwidth]{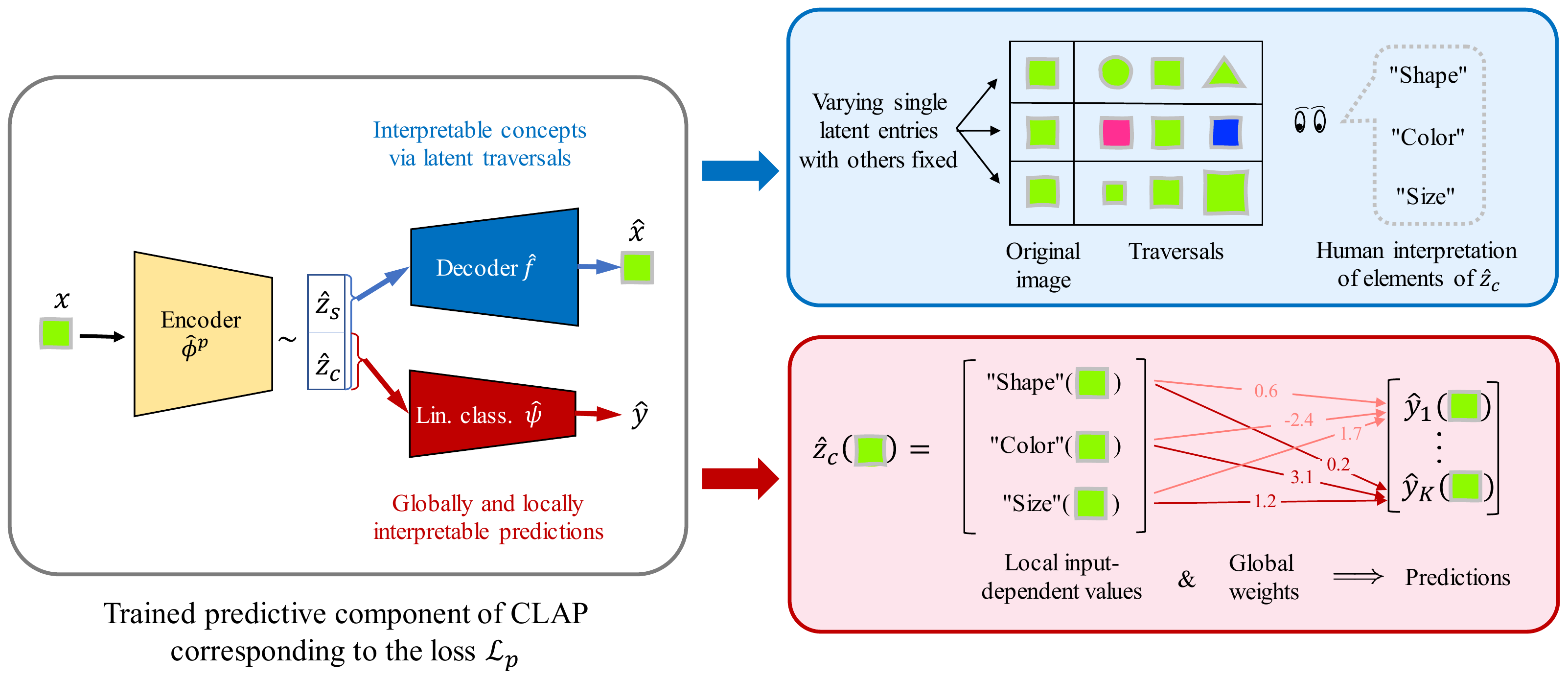}
    \caption{\small{We present how the prediction model obtained by training \methName{} can be used and interpreted at test time. Supplying a test images $x$ to the component $\mathcal{L}_p$ of \methName{}, we learn core features $\hat{\Z}_c$. These features are visualized using latent traversals and interpreted by a human, who assigns them to high-level concepts. Furthermore, the estimated linear classifier predicts a label and provides global (population wise) and local (instance wise) importance for the interpreted concepts. }
    }
    \label{fig:teaser}
    \vspace{-0.2in}
\end{figure}

We now discuss how \methName{}'s trained model can be used to produce an end-to-end interpretable prediction model pipeline, which we represent in Fig.~\ref{fig:teaser}. 

At inference time, the part of \methName{}'s model corresponding to $\mathcal{L}_p$ is utilized, since it does not require a label as an input (Fig.~\ref{fig:teaser} left). As we describe in detail next, the learned concepts are visualized using latent traversals; to conclude the pipeline, a human expert visually inspects these traversals and assigns a meaning to the relative latent variables.

\textbf{Interpretations via latent traversals} Generally, the visual explanations provided by the model need to be evaluated by a human expert (see \oursection{}~\ref{sec:introduction}). As is customary for VAE models, we provide such visualizations via latent traversals. Specifically, let $x$ be an input image. The core concepts associated to $x$ are obtained via the posterior mean $\hat{\mu}(x) := \mathbb{E}_{q_{\hat{\phi}^p}(\hat{\Z}_c|x)}[\hat{\Z}_c]$. 
The semantics of $\hat{\Z}_c$ are then discovered by performing latent traversals. In these, we change one component of $\hat \mu (x)$ at a time, while keeping the others fixed, and observe the reconstructions obtained through the decoder $\hat{f}$.
Owing to the concept-learning capabilities of \methName{}, the traversals on the core latent features will produce distinct changes in the reconstructed images corresponding to the different discovered ground-truth concepts, which will allow the human expert to assign them with a semantic meaning. This procedure is represented in the top-right of Fig.~\ref{fig:teaser}. There, for example, upon visual inspection, the first latent is assigned the meaning of \emph{"Shape"} from the expert, the second \emph{"Color"}, and so on.

\textbf{Interpretable predictions using learned concepts} We note here that in our experiments, we found a linear classifier to be well-performing across all datasets. For this reason, the following description assumes $\psi$ to simply be the linear weights of the corresponding linear classifier $p_{\psi}(\Y | \hat{\Z}_c)$.
For each concept, we provide both a global and local relevance for prediction, as depicted in the bottom right of Fig.~\ref{fig:teaser}. The global relevance represents the importance of a concept for prediction at a population level (i.e. across images) and is thus directly encoded in the entries of $\hat{\psi}$. The local relevance is instead image-specific, and is observed in the summands of the linear combination $\langle \hat{\mu}(x),\hat{\psi}\rangle$. These two measures allow the practitioner to transparently assess the decision process of the model, as they assign a prediction weight to human interpretable features.

\label{sec:experiments}

\section{Experiments: using CLAP for interpretable predictions}
\label{sec:experiments}
\begin{figure}[t]
\centering
\begin{subfigure}[b]{0.49\textwidth}
\centering
\includegraphics[width = 1\textwidth]{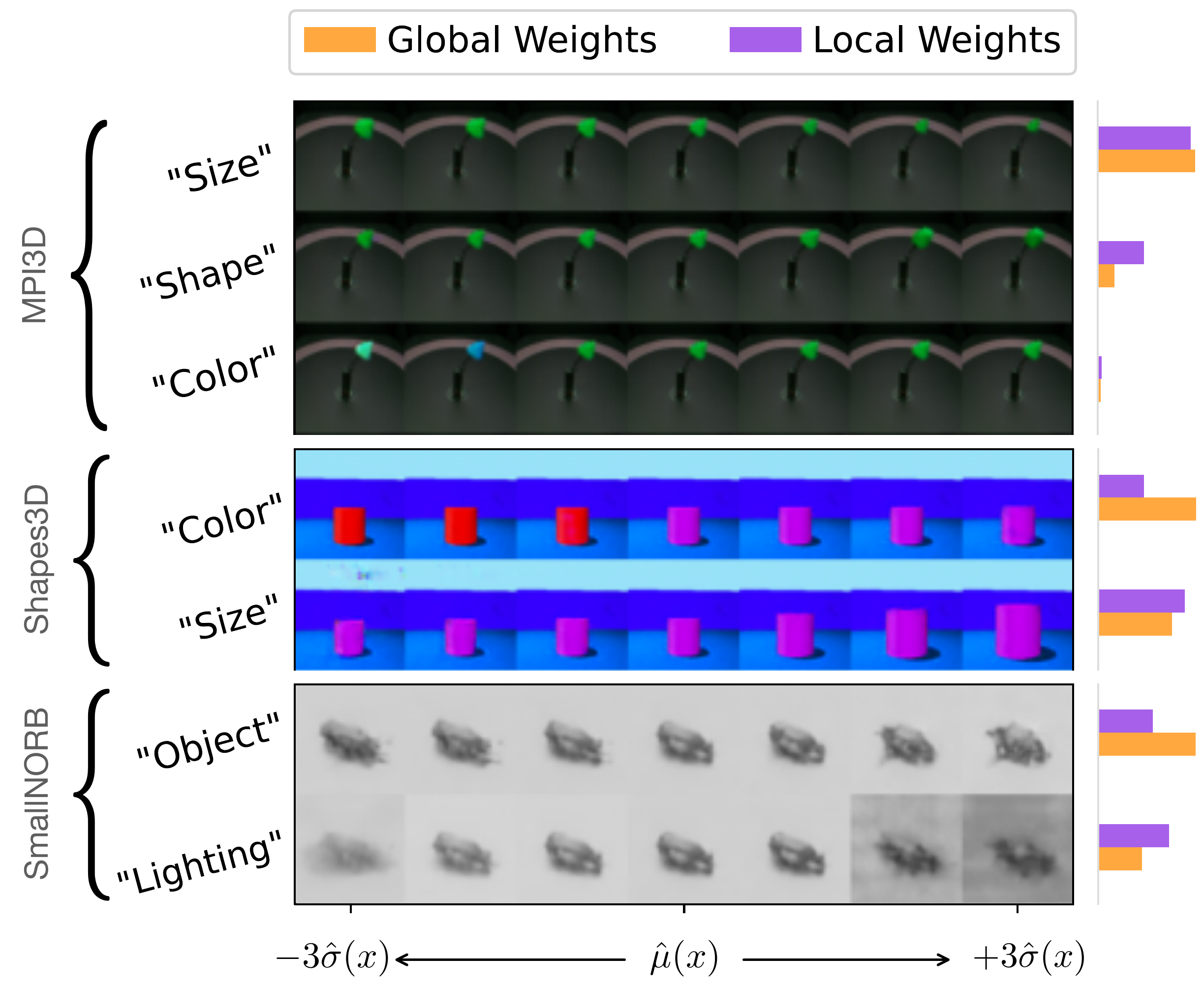}
    \caption{\methName{} traversals and interpretations}
     \label{fig:CLAP_traversals_synthetic}
\end{subfigure}
\begin{subfigure}[b]{0.49\textwidth}
\centering
\includegraphics[scale = 0.455]{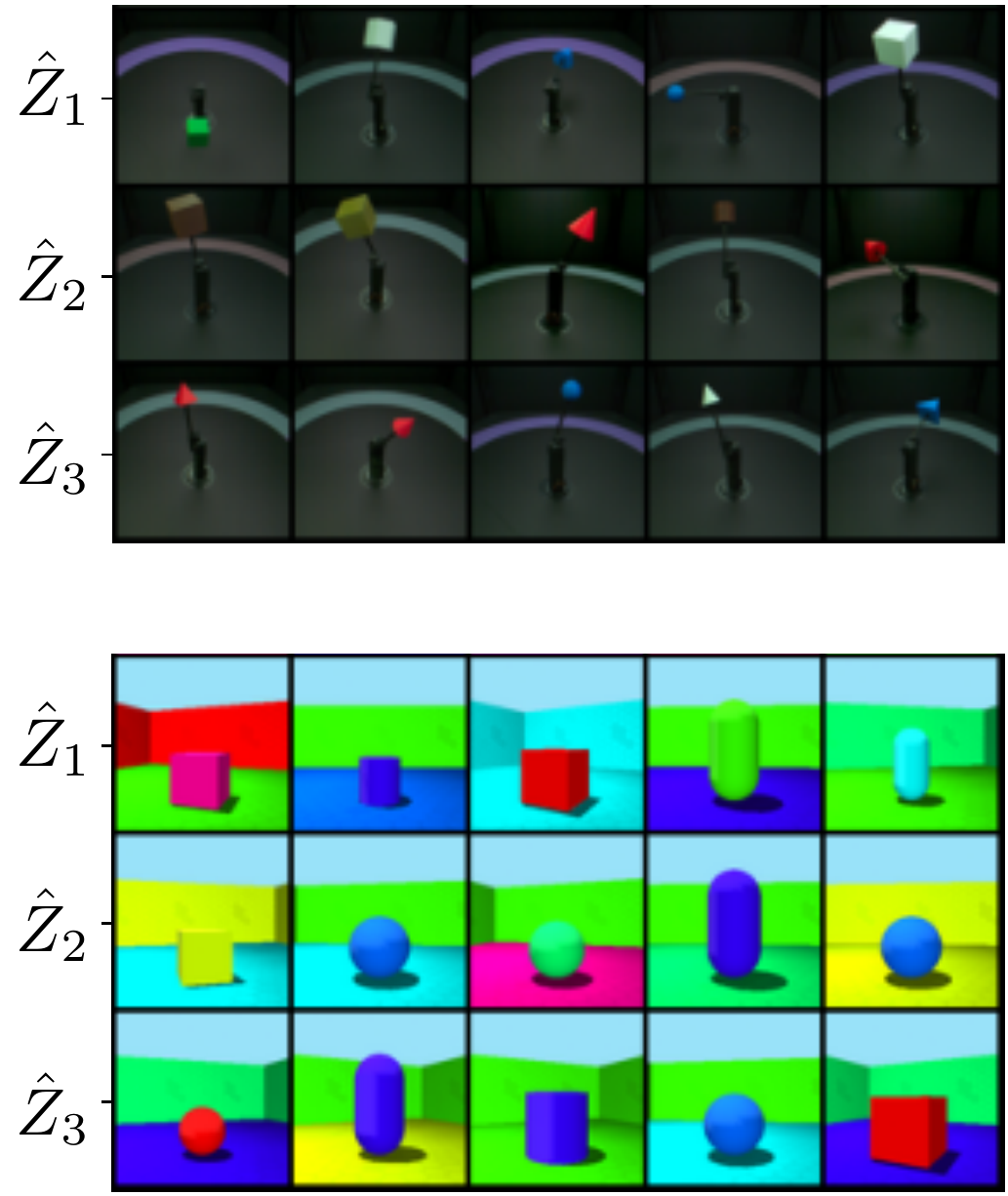}
    \caption{SENN prototypes}
\label{fig:mpi_senn}
\end{subfigure}
\caption{\small{a) \methName{} traversals on (in order) the MPI3D, Shapes3D and SmallNORB datasets, and b) SENN prototypes on (in order) the MPI3D and Shapes3D datasets.}}
\label{fig:traversals_clap}
\end{figure}
We next present experiments on synthetic data to corroborate our theoretical results, and evaluate the ability of \methName{} to learn an accurate prediction model using the ground-truth features. 
Since in most real-world datasets, ground-truth factors are unknown but necessary to verify whether \methName{} can work in practice,
we resort to three standard "disentanglement" datasets MPI3D \citep{gondal2019transfer}, Shapes3D \citep{3dshapes18} and SmallNORB \citep{smallnorb}.
These datasets consist of collections of objects generated synthetically according to some ground-truth factors of variation. The images are a priori unlabeled; thus, we select some of the ground-truth factors, which represent the concepts $\zc$ to be discovered, and generate artificial binary labels $\Y$. The ground-truth factors $\zc$ are object shape, size and color for MPI3D, object color and size for Shapes3D and object type and lighting for SmallNORB (see \suppmat{} \oursection~\ref{sec: details synthetic}). For all the experiments and baselines in \oursection{}~\ref{sec:experiments}, details on training and architectures employed are deferred to \suppmat{} \oursection~\ref{sec:impdetails}\footnote{Our code is publicly available at \url{https://github.com/nickruggeri/CLAP-interpretable-predictions}}. In general, for all methods, we used neural network architectures comparable in complexity to those utilized in \citep{ICLR-VAE2021, qiao2019disentanglement}. 

As explained in \oursection{}~\ref{sec:stage2}, we proceed with the evaluation of \methName{} by first generating latent traversals.
The goal is to determine whether the discovered concepts have a one-to-one correspondence with the ground-truth $\zc$ that we used to generate the data. In Fig.~\ref{fig:CLAP_traversals_synthetic}, every row corresponds to the traversal for one latent feature. As can be observed, the estimated core features indeed represent the ground-truth ones; this means that the model identifies the ground-truth concepts underlying the data generating mechanism. Importantly, we remark that the concept names assigned to the single rows (e.g. \emph{"Size", "Shape"}) are obtained by visual inspection; the model doesn't have direct access to them, but only to the images $\X$ and labels $\Y$.  

Finally, the discovered $\zc$ are also fully predictive, as \methName{} achieves classification accuracy above $0.99$ on all the datasets. 
We include additional traversals in \suppmat{} \oursection{}~\ref{sec:mpitraversals}; there, we also show that, due to the sparsity regularization penalty $\rho(f,\psi)$, the model accurately assigns negligible global and local weights (i.e. no predictive value) to the remaining latent features included in the model. This is in contrast to the concepts shown in Fig.~\ref{fig:CLAP_traversals_synthetic} that have non-negligible global and local weights. In other words, in line with our theory, estimated core features that have prediction power align with the ground-truth concepts. 

\textbf{Comparison with baselines} We compare the outputs of \methName{} with those of SENN \citep{alvarez2018towards} and CCVAE \citep{ICLR-VAE2021}, two prediction models in the existing literature that are closest to \methName{}.
To explain its predictions and visualize the learned concepts, SENN uses prototypes -- a set of training images that ``best represent'' every latent variable. In Fig.~\ref{fig:mpi_senn}, we depict the prototypes relative to some of these features.
Similarly to \methName{}, human inspection is needed to describe the concepts that such latents encode. However, the task here is substantially more difficult: for any of the latents, we can observe many different changes, e.g in the first row objects of different colors and shapes are observed, and from different camera angles. This indicates that not only SENN is not able to identify the ground-truth $\zc$, thus hindering interpretability, but also mixes them with non-predictive style features $\zs$. We also apply CCVAE on synthetic data and observe that its learned latent features do not align with the ground-truth ones; due to space constraints, we show these results in \suppmat{} \oursection{}~\ref{sec:ccvae on shapes3d appendix}. 

\textbf{Ablation studies}
In order to demonstrate the importance of each of our design choices, we also perform various ablation studies on the MPI3D dataset, presented in \suppmat{} \ref{sec:ablations}. 
Firstly, we show that if the sparsity penalty $\lambda_n\rho(f,\psi)$ is removed from the learning objective, the resulting model utilizes separately some latent variables for visualization, and some others for prediction. On the other hand, with the use of $\lambda_n\rho(f,\psi)$, \methName{} ensures correspondence between features utilized for prediction and visualization. Furthermore, we show latent traversals for a model trained only on $\mathcal{L}_p$. As explained in \oursection{}~\ref{sec:vanilla}, the learned features are fully predictive, but do not correspond to the ground-truth one. In fact, it can be observed that various ground-truth features change jointly within one single traversal. Further, we empirically confirm that
the concept-learning capabilities of \methName{} rely on the labels $\Y$ being informative enough, as highlighted by the assumptions in \oursection{}~\ref{sec: clap_as_solution}. Practically, this means that multiple labels help with more accurate recovery of the ground-truth $\zc$;
we show that the concept learning capabilities of \methName{} indeed decrease on a dataset where only one label is available.

\section{Future Outlook}
\label{sec:future work}
\vspace{-0.1in}

So far, we have evaluated \methName{} in synthetic scenarios where we know the ground-truth data generating mechanism and the core factors are easy to recognize for a {layperson}. For many scientific scenarios such as the example in the introduction, evaluating whether learned concepts correspond to the "ground-truth" can only be done by domain experts.  Nevertheless, we provide the outputs of \methName{} for some challenging real datasets to highlight
some of its favorable properties compared to other competing methods. 

\begin{figure}[t]
\vspace{-1cm}
\centering
\begin{subfigure}[b]{1\linewidth}
\centering
    \includegraphics[scale = 0.3]{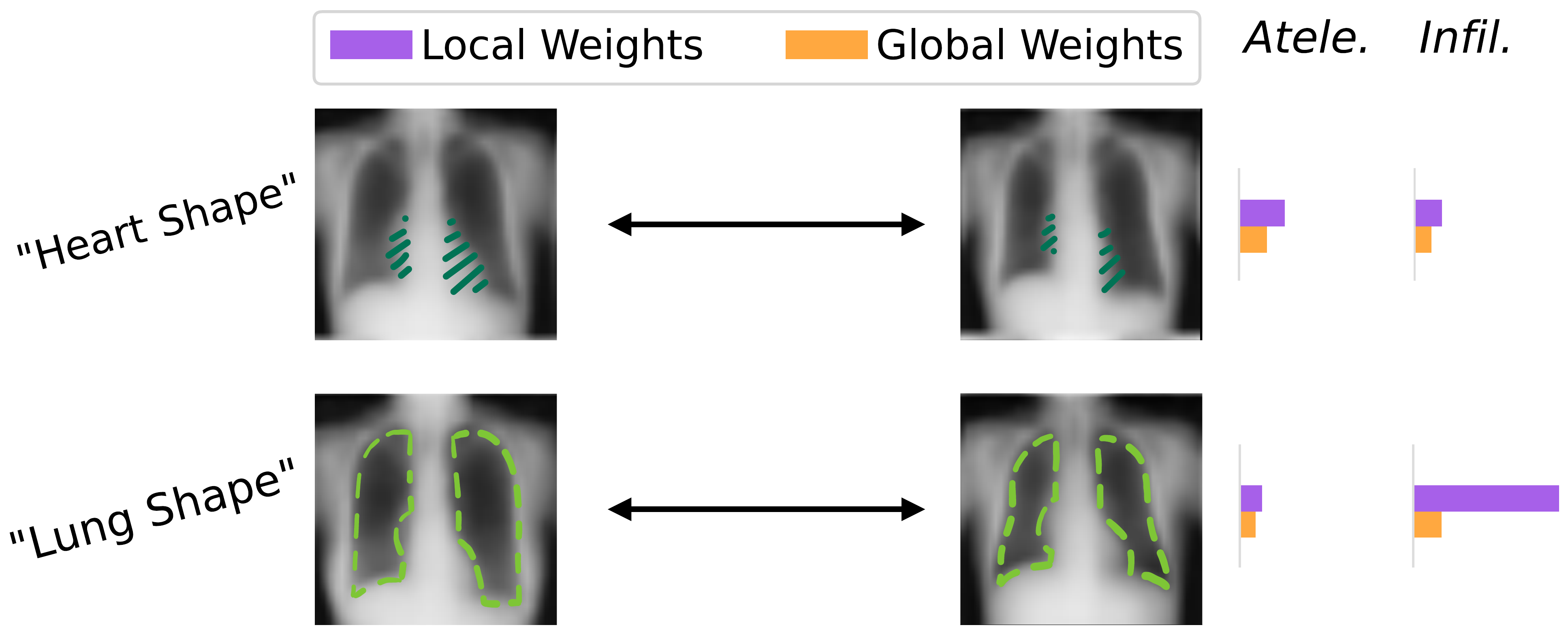}
    \caption{\methName{} traversals and interpretations}
    \label{fig:traversals_chestxray_clap}
\end{subfigure}
\vspace{0.03in}
\hspace{-0.15in}\begin{subfigure}[b]{1\linewidth}
\centering
    \includegraphics[scale = 0.3]{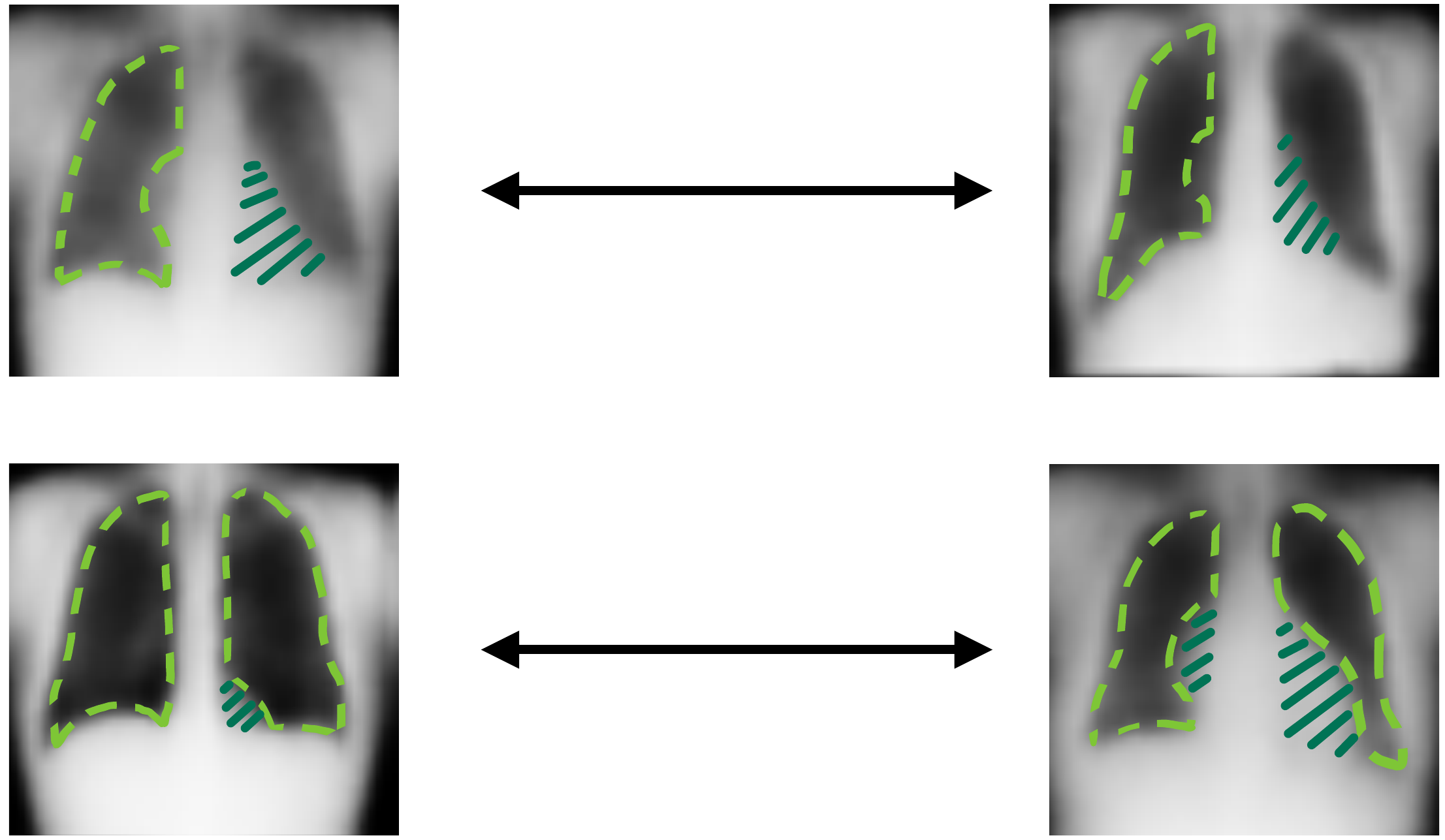}
    \caption{CCVAE traversals}
\label{fig:traversals_chestxray_ccvae}
\end{subfigure}
\caption{\small{Output of \methName{} and traversals of CCVAE for the \chest{} dataset. In (a), we present the weights for both the atelectasis and lung infiltration disease predictions, as well as the human interpretations of the discovered concepts. For better visual comparison, we only show the images obtained at the extremes of the latent traversals. Moreover, we highlight the changes that occur during the traversals. We include magnified figures with full traversals in \suppmat{} \oursection{}~\ref{sec:chestxray_traversals}, as well as a glossary on how to read the results.}}
\label{fig:chestxray}
\vspace{-0.2in}
\end{figure}

In this section, we present results on the \chest{} dataset, and defer additional experiments on the PlantVillage dataset \citep{leafdisease} to \suppmat{} \oursection~\ref{sec:leavestraversals}. The \chest{} dataset \cite{wang2017chestxray} consists of radiography images; each image has 14 associated binary disease labels. 
We emphasize  
that only the disease labels may be used to learn the underlying concepts and no additional supervision is available. As explained in \oursection~\ref{sec:related_work}, many inherently interpretable models cannot be applied successfully in this setting, since they generally assume further information on the ground-truth factors. 
Due to the negative results for SENN in \oursection{}~\ref{sec:experiments}, we only compare our method with CCVAE.

Both \methName{} and CCVAE attain similar classification accuracies of 0.903 and 0.898, respectively. In Fig.~\ref{fig:chestxray}, we compare the traversals obtained by both methods.
First we observe that \methName{} manages to learn concepts that are localized in the X-Ray image, corresponding to {separate properties},
such as \emph{"Heart shape"} and \emph{"Lung shape"}. 
Instead, in both the traversals presented for CCVAE, characteristics that can be associated to both the heart and lung shapes vary together. Thus, while CCVAE finds similar concepts for prediction, they do not appear distinctly as separate components of $\zc$.
For this reason, it is harder for a human expert to uniquely label the learned concepts and, consequently, interpret the model's output.

Another desirable characteristic of \methName{} is that the global and local weights reflect the importance of the concepts in predicting different diseases.
For example, compared to atelectasis, the concept "lung shape" has higher weight (both global and local) in determining the presence of lung infiltration. Since lung infiltration is a condition related to dense substances in the lungs, the concept "lung shape" learned by \methName{} is natural and indicative.
Further, we remark that the discovered concepts manifest through very nuanced traversals. 
This is sensible, as it is to be expected that real life examples come with subtle and less pronounced features than synthetic and commonly used datasets. In conclusion, these experiments show the advancement and potential of \methName{} compared to existing methods for providing real-life interpretable predictions.


There are a number of exciting future directions that can further improve \methName{} for broader and more effective use in real-world scenarios.
For example, the visualizations of the VAE are not optimally sharp compared to the status quo for GANs.
Hence, it would be interesting to explore
whether one can obtain provable concept learning when the
VAE is replaced by a GAN structure.
Further, in many scientific applications, the number of available images can be quite small. An interesting avenue for future research could be to develop solutions for the small data regime, e.g. via transfer learning.

{\bf{Acknowledgements}}
This project has received funding from the European Research Council (ERC) under the
European Union’s Horizon 2020 research and innovation programme (grant agreement No.
786461). We thank Christina Heinze-Deml, Laura Manduchi and Ricards Marcinkevics for the useful discussions and feedback on our work. We thank Pietro Spolettini for helping in the interpretation of the \chest{} medical data.

\bibliography{bibliography}

\begin{thebibliography}{37}
\providecommand{\natexlab}[1]{#1}
\providecommand{\url}[1]{\texttt{#1}}
\expandafter\ifx\csname urlstyle\endcsname\relax
  \providecommand{\doi}[1]{doi: #1}\else
  \providecommand{\doi}{doi: \begingroup \urlstyle{rm}\Url}\fi

\bibitem[Alvarez-Melis and Jaakkola(2018{\natexlab{a}})]{AlvarezMelis2018OnTR}
David Alvarez-Melis and Tomi Jaakkola.
\newblock On the robustness of interpretability methods.
\newblock In \emph{arXiv preprint arXiv:1806.08049}, 2018{\natexlab{a}}.

\bibitem[Alvarez-Melis and Jaakkola(2018{\natexlab{b}})]{alvarez2018towards}
David Alvarez-Melis and Tomi Jaakkola.
\newblock Towards robust interpretability with self-explaining neural networks.
\newblock In \emph{Neural Information Processing Systems}, 2018{\natexlab{b}}.

\bibitem[Antoran et~al.(2021)Antoran, Bhatt, Adel, Weller, and
  Hern{\'a}ndez-Lobato]{antoran2021getting}
Javier Antoran, Umang Bhatt, Tameem Adel, Adrian Weller, and Jos{\'e}~Miguel
  Hern{\'a}ndez-Lobato.
\newblock Getting a {CLUE}: A method for explaining uncertainty estimates.
\newblock In \emph{International Conference in Learning Representations}, 2021.

\bibitem[Bach et~al.(2015)Bach, Binder, Montavon, Klauschen, M{\"u}ller, and
  Samek]{Bach15Pixel}
Sebastian Bach, Alexander Binder, Gr{\'e}goire Montavon, Frederick Klauschen,
  Klaus-Robert M{\"u}ller, and Wojciech Samek.
\newblock On pixel-wise explanations for non-linear classifier decisions by
  layer-wise relevance propagation.
\newblock \emph{PLOS ONE}, 10\penalty0 (7):\penalty0 1--46, 07 2015.

\bibitem[Burgess and Kim(2018)]{3dshapes18}
Chris Burgess and Hyunjik Kim.
\newblock {3D Shapes Dataset}.
\newblock https://github.com/deepmind/3dshapes-dataset/, 2018.

\bibitem[Burgess et~al.(2018)Burgess, Higgins, Pal, Matthey, Watters,
  Desjardins, and Lerchner]{AnnealedVAE}
Christopher Burgess, Irina Higgins, Arka Pal, Lo{\"i}c Matthey, Nicholas
  Watters, Guillaume Desjardins, and Alexander Lerchner.
\newblock Understanding disentangling in $\beta$-{VAE}.
\newblock In \emph{arXiv preprint arXiv:1804.03599}, 2018.

\bibitem[Chang et~al.(2019)Chang, Creager, Goldenberg, and
  Duvenaud]{Chang2019ExplainingIC}
Chun-Hao Chang, Elliot Creager, Anna Goldenberg, and David~Kristjanson
  Duvenaud.
\newblock Explaining image classifiers by counterfactual generation.
\newblock In \emph{International Conference in Learning Representations}, 2019.

\bibitem[Chen et~al.(2018)Chen, Li, Grosse, and Duvenaud]{chen2018isolating}
Tian~Qi Chen, Xuechen Li, Roger Grosse, and David~Kristjanson Duvenaud.
\newblock Isolating sources of disentanglement in variational autoencoders.
\newblock In \emph{Neural Information Processing Systems}, 2018.

\bibitem[Gondal et~al.(2019)Gondal, W{\"u}thrich, Miladinovi{\'c}, Locatello,
  Breidt, Volchkov, Akpo, Bachem, Sch{\"o}lkopf, and Bauer]{gondal2019transfer}
Muhammad~Waleed Gondal, Manuel W{\"u}thrich, DJ~Miladinovi{\'c}, Francesco
  Locatello, Martin Breidt, Valentin Volchkov, Joel Akpo, Olivier Bachem,
  Bernhard Sch{\"o}lkopf, and Stefan Bauer.
\newblock On the transfer of inductive bias from simulation to the real world:
  a new disentanglement dataset.
\newblock In \emph{Neural Information Processing Systems}, 2019.

\bibitem[Gresele et~al.(2019)Gresele, Rubenstein, Mehrjou, Locatello, and
  Sch{\"o}lkopf]{Gresele2019TheIR}
Luigi Gresele, Paul Rubenstein, Arash Mehrjou, Francesco Locatello, and
  Bernhard Sch{\"o}lkopf.
\newblock The incomplete {R}osetta stone problem: Identifiability results for
  multi-view nonlinear {ICA}.
\newblock In \emph{Uncertainty in Artificial Intelligence}, 2019.

\bibitem[Higgins et~al.(2017)Higgins, Matthey, Pal, Burgess, Glorot, Botvinick,
  Mohamed, and Lerchner]{Higgins2017betaVAELB}
Irina Higgins, Lo{\"i}c Matthey, Arka Pal, Christopher~P. Burgess, Xavier
  Glorot, Matthew~M. Botvinick, Shakir Mohamed, and Alexander Lerchner.
\newblock beta-{VAE}: Learning basic visual concepts with a constrained
  variational framework.
\newblock In \emph{International Conference in Learning Representations}, 2017.

\bibitem[Hughes et~al.(2015)Hughes, Salath{\'e}, et~al.]{leafdisease}
David Hughes, Marcel Salath{\'e}, et~al.
\newblock An open access repository of images on plant health to enable the
  development of mobile disease diagnostics.
\newblock In \emph{arXiv preprint arXiv:1511.08060}, 2015.

\bibitem[Hyv{\"a}rinen et~al.(2019)Hyv{\"a}rinen, Sasaki, and
  Turner]{Hyvrinen2019NonlinearIU}
Aapo Hyv{\"a}rinen, Hiroaki Sasaki, and Richard Turner.
\newblock Nonlinear {ICA} using auxiliary variables and generalized contrastive
  learning.
\newblock In \emph{International Conference on Artificial Intelligence and
  Statistics}, 2019.

\bibitem[Joy et~al.(2021)Joy, Schmon, Torr, Siddharth, and
  Rainforth]{ICLR-VAE2021}
Tom Joy, Sebastian Schmon, Philip Torr, N~Siddharth, and Tom Rainforth.
\newblock Capturing label characteristics in {VAEs}.
\newblock In \emph{International Conference in Learning Representations}, 2021.

\bibitem[Kazhdan et~al.(2020)Kazhdan, Dimanov, Jamnik, Li{\`o}, and
  Weller]{Kazhdan2020NowYS}
Dmitry Kazhdan, Botty Dimanov, Mateja Jamnik, Pietro Li{\`o}, and Adrian
  Weller.
\newblock Now you see me ({CME}): Concept-based model extraction.
\newblock In \emph{International Conference on Information and Knowledge
  Management}, 2020.

\bibitem[Khemakhem et~al.(2020)Khemakhem, Kingma, Monti, and
  Hyv{\"a}rinen]{VAE-ICA}
Ilyes Khemakhem, Ricardo Kingma, Pio Monti, and Aapo Hyv{\"a}rinen.
\newblock Variational autoencoders and nonlinear {ICA}: A unifying framework.
\newblock In \emph{International Conference on Artificial Intelligence and
  Statistics}, 2020.

\bibitem[Kim and Mnih(2019)]{kim2019disentangling}
Hyunjik Kim and Andriy Mnih.
\newblock Disentangling by factorising.
\newblock In \emph{International Conference on Machine Learning}, 2019.

\bibitem[Kingma et~al.(2014)Kingma, Rezende, Mohamed, and Welling]{KingmaSemi}
Diederik Kingma, Danilo Rezende, Shakir Mohamed, and Max Welling.
\newblock Semi-supervised learning with deep generative models.
\newblock In \emph{Neural Information Processing Systems}, 2014.

\bibitem[Kumar et~al.(2017)Kumar, Sattigeri, and
  Balakrishnan]{Kumar17Disentangle}
Abhishek Kumar, Prasanna Sattigeri, and Avinash Balakrishnan.
\newblock Variational inference of disentangled latent concepts from unlabeled
  observations.
\newblock In \emph{International Conference in Learning Representations}, 2017.

\bibitem[Kumar et~al.(2020)Kumar, Venkatasubramanian, Scheidegger, and
  Friedler]{Kumar2020ProblemsWS}
Indra~Elizabeth Kumar, Suresh Venkatasubramanian, Carlos~Eduardo Scheidegger,
  and Sorelle~A. Friedler.
\newblock Problems with {S}hapley-value-based explanations as feature
  importance measures.
\newblock In \emph{International Conference in Machine Learning}, 2020.

\bibitem[Lang et~al.(2021)Lang, Gandelsman, Yarom, Wald, Elidan, Hassidim,
  Freeman, Isola, Globerson, Irani, and Mosseri]{Lang2021ExplainingIS}
Oran Lang, Yossi Gandelsman, Michal Yarom, Yoav Wald, Gal Elidan, Avinatan
  Hassidim, William~T. Freeman, Phillip Isola, Amir Globerson, Michal Irani,
  and Inbar Mosseri.
\newblock Explaining in style: Training a {GAN} to explain a classifier in
  {StyleSpace}.
\newblock In \emph{International Conference in Computer Vision}, 2021.

\bibitem[LeCun et~al.(2004)LeCun, Huang, and Bottou]{smallnorb}
Yann LeCun, Fu~Huang, and L{\'e}on Bottou.
\newblock Learning methods for generic object recognition with invariance to
  pose and lighting.
\newblock In \emph{Computer Vision and Pattern Recognition}, 2004.

\bibitem[Locatello et~al.(2019)Locatello, Bauer, Lucic, Raetsch, Gelly,
  Sch{\"o}lkopf, and Bachem]{Locatello-challenges}
Francesco Locatello, Stephan Bauer, Mario Lucic, Gunar Raetsch, Sylvain Gelly,
  Bernhard Sch{\"o}lkopf, and Olivier Bachem.
\newblock Challenging common assumptions in the unsupervised learning of
  disentangled representations.
\newblock In \emph{International Conference in Machine Learning}, 2019.

\bibitem[Locatello et~al.(2020)Locatello, Poole, Raetsch, Sch{\"o}lkopf,
  Bachem, and Tschannen]{Locatello_weakly_supervisedd}
Francesco Locatello, Ben Poole, Gunnar Raetsch, Bernhard Sch{\"o}lkopf, Olivier
  Bachem, and Michael Tschannen.
\newblock Weakly-supervised disentanglement without compromises.
\newblock In \emph{International Conference in Machine Learning}, 2020.

\bibitem[Mita et~al.(2021)Mita, Filippone, and Michiardi]{ID-VAE2021}
Graziano Mita, Maurizio Filippone, and Pietro Michiardi.
\newblock An identifiable double {VAE} for disentangled representations.
\newblock In \emph{International Conference on Machine Learning}, 2021.

\bibitem[Qiao et~al.(2019)Qiao, Li, Xu, Cai, and
  Zhang]{qiao2019disentanglement}
Jie Qiao, Zijian Li, Boyan Xu, Ruichu Cai, and Kun Zhang.
\newblock Disentanglement challenge: From regularization to reconstruction.
\newblock \emph{arXiv preprint arXiv:1912.00155}, 2019.

\bibitem[Reyes et~al.(2020)Reyes, Meier, Pereira, Silva, Dahlweid, von
  Tengg-Kobligk, Summers, and Wiest]{Reyes2020OnTI}
Mauricio Reyes, Raphael Meier, S{\'e}rgio Pereira, Carlos Silva, Fried
  Dahlweid, Hendrik von Tengg-Kobligk, Ronald Summers, and Roland Wiest.
\newblock On the interpretability of artificial intelligence in radiology:
  Challenges and opportunities.
\newblock \emph{Radiology:Artificial intelligence}, 23:\penalty0 e190043, 2020.

\bibitem[Rezende et~al.(2014)Rezende, Mohamed, and
  Wierstra]{pmlr-v32-rezende14}
Danilo Rezende, Shakir Mohamed, and Daan Wierstra.
\newblock Stochastic backpropagation and approximate inference in deep
  generative models.
\newblock In \emph{International Conference in Machine Learning}, 2014.

\bibitem[Rudin(2018)]{Rudin2018PleaseSE}
Cynthia Rudin.
\newblock Please stop explaining black box models for high stakes decisions.
\newblock In \emph{arXiv preprint arXiv:1811.10154}, 2018.

\bibitem[Selvaraju et~al.(2017)Selvaraju, Cogswell, Das, Vedantam, Parikh, and
  Batra]{Selavarju17}
Ramprasaath Selvaraju, Michael Cogswell, Abhishek Das, Ramakrishna Vedantam,
  Devi Parikh, and Dhruv Batra.
\newblock Grad-{CAM}: Visual explanations from deep networks via gradient-based
  localization.
\newblock In \emph{International Conference on Computer Vision}, 2017.

\bibitem[Shu et~al.(2020)Shu, Chen, Kumar, Ermon, and Poole]{Shu2020WeaklySD}
Rui Shu, Yining Chen, Abhishek Kumar, Stefano Ermon, and Ben Poole.
\newblock Weakly supervised disentanglement with guarantees.
\newblock In \emph{International Conference on Learning Representations}, 2020.

\bibitem[Simonyan et~al.(2014)Simonyan, Vedaldi, and
  Zisserman]{Simonyan2014DeepIC}
Karen Simonyan, Andrea Vedaldi, and Andrew Zisserman.
\newblock Deep inside convolutional networks: Visualising image classification
  models and saliency maps.
\newblock In \emph{arXiv preprint arxiv:1312.6034}, 2014.

\bibitem[Stiglic et~al.(2020)Stiglic, Kocbek, Fijavko, Zitnik, Verbert, and
  Cilar]{tiglic2020InterpretabilityOM}
Gregor Stiglic, Primoz Kocbek, Nino Fijavko, Marinka Zitnik, Katrien Verbert,
  and Leona Cilar.
\newblock Interpretability of machine learning‐based prediction models in
  healthcare.
\newblock \emph{Wiley Interdisciplinary Reviews: Data Mining and Knowledge
  Discovery}, 10, 2020.

\bibitem[von K{\"u}gelgen et~al.(2021)von K{\"u}gelgen, Sharma, Gresele,
  Brendel, Scholkopf, Besserve, and Locatello]{vonKgelgen2021SelfSupervisedLW}
Julius von K{\"u}gelgen, Yash Sharma, Luigi Gresele, Wieland Brendel, Bernhard
  Scholkopf, Michel Besserve, and Francesco Locatello.
\newblock Self-supervised learning with data augmentations provably isolates
  content from style.
\newblock In \emph{Neural Information Processing Systems}, 2021.

\bibitem[Wang et~al.(2017)Wang, Peng, Lu, Lu, Bagheri, and
  Summers]{wang2017chestxray}
Xiaosong Wang, Yifan Peng, Le~Lu, Zhiyong Lu, Mohammadhadi Bagheri, and Ronald
  Summers.
\newblock Chest{X}-ray8: Hospital-scale chest {X}-ray database and benchmarks
  on weakly-supervised classification and localization of common thorax
  diseases.
\newblock In \emph{Computer Vision and Pattern Recognition}, 2017.

\bibitem[Wu et~al.(2021)Wu, Lischinski, and Shechtman]{Wu2021StyleSpaceAD}
Zongze Wu, Dani Lischinski, and Eli Shechtman.
\newblock {StyleSpace} analysis: Disentangled controls for {StyleGAN} image
  generation.
\newblock In \emph{Conference on Computer Vision and Pattern Recognition},
  2021.

\bibitem[Yeh et~al.(2020)Yeh, Kim, Arik, Li, Pfister, and
  Ravikumar]{Yeh2020OnCompleteness}
Chih Yeh, Been Kim, Sercan Arik, Chun-Liang Li, Tomas Pfister, and Pradeep
  Ravikumar.
\newblock On completeness-aware concept-based explanations in deep neural
  networks.
\newblock In \emph{Neural Information Processing Systems}, 2020.

\end{thebibliography}
\newpage
\appendix
\section*{\centering Supplementary Material for ``Provable concept learning for interpretable predictions using variational inference"}

\section{Proof of Theoretical results}
For simplicity, we first prove guarantees for \methName{} in the setting where the number of latent variables is specified correctly. Subsequently, we will extend the analysis to the setting where the number of latent variables is miss-specified (i.e. chosen conservatively). 

Throughout, we use the following notation. Let $\mathbb{P}_{\X}$ be the distribution of $\X$, $\ProbY$ be the probability distribution of, $\ProbXY$ be the joint probability distribution of and $(\X,\Y)$, and $\ProbXcY$ be the probability distribution of $\X|\Y$, all with respect to the data generating model. Associated with the probability distributions $\mathbb{P}_{\X}$ and $\ProbXcY$ are the density functions with we denote by $\pstarX$ and $\pstarXcY$. Finally, we use the notation $\tilde{\Z}$ to denote latent variables that specify the VAE model.

\subsection{Formal description of our assumptions}
\label{sec:assumptions}
Let $f'$ be the function from the decomposition $f = f' \circ B$ introduced in relation to the sparsity regularization term ($B$ is the identity matrix when the number of latents is correctly specified). Assumption 1 then is formally described as follows:
\begin{customassump}{1}[Concept learning, formal]
\label{ass:cl_formal} 
\begin{equation*}
\begin{aligned}
&\text{Assumption 1.1.}~~~~ \text{The functions }f',f^\star \text{ are one-to-one and continuous}\\
&\text{Assumption 1.2.}~~~~\text{There exists } (y,\tilde{y}) \in \mathcal{Y} \text{ s.t. } D_{y}^\star({D}_{\tilde{y}}^\star)^{-1} \text{ has distinct diagonal entries not equal to one}.\\[0.05in]
\end{aligned}
\end{equation*}
\end{customassump}
Assumptions 1.1 is rather mild and ensures that the functions mapping from the latent space to the input space are injective. Assumption 1.2 states that variations in the label $Y$ should impact the variance of all the core latent features (hence the exclusion of value ``one") and in a distinct manner. This type of assumption is similar in spirit to requiring ``heterogeneous interventions" in causal structural learning. Specifically, in the context of our anti-causal graphical model \ref{fig:graph_models}, the labels $y \in \mathcal{Y}$ can be viewed as an ``environment" variable where an environment dictates the distribution of the core latent features. A change in an environment can then be viewed as interventions on the core latent features. Thus, in this perspective, Assumption 1.2 requires the impact of the interventions to be on all of the core features and to be sufficiently heterogeneous. Furthermore, we note that Assumption 1.2 requires that changes to the label lead to simultaneous changes to all of the core features. This assumption can be relaxed so that not all of the core features must vary at once with a change in the label, as long as each feature varies for some change to the label. Mathematically, the assumption can be relaxed to the following: for every $i,j \in [k_c], i\neq j$, there exists $y,\tilde{y} \in \mathcal{Y}$ such that $[D^\star_y{D^\star_{\tilde{y}}}^{-1}]_{i,i} \neq [D^\star_y{D^\star_{\tilde{y}}}^{-1}]_{j,j}$ and $[D^\star_y{D^\star_{\tilde{y}}}^{-1}]_{i,i} \neq 1$, $[D^\star_y{D^\star_{\tilde{y}}}^{-1}]_{j,j} \neq 1$. 


\subsection{Analysis with known number of latent features}
\label{sec:proof_of_main}
For building intuition, we first provide an analysis of \methName{} in the setting where the number of latents is correctly specified. Along the way, we prove why maximizing $\mathcal{L}_p$ achieves optimal prediction and maximizing $\mathcal{L}_{cl}$ learns the ground-truth concepts. Throughout, we denote $k^\star_c,k^\star_s$ to be the dimension of the true core and style features and $k^\star = k^\star_c + k^\star_s$ to be the total number of latent features. 

\subsubsection{Maximizing $\mathcal{L}_p$ achieves optimal prediction}
As described in Section~\ref{sec:vanilla}, maximizing the objective $\mathcal{L}_p$ achieves optimal prediction. We formalize this below.
\begin{customlemma}{2}[Maximizing $\mathcal{L}_{p}$ achieves optimal prediction]  Suppose the data is generated according to the model in \eqref{eqn:model} with no noise, i.e. $\epsilon \equiv 0$ and Assumptions~1.1 and 2 hold. Suppose $\mathcal{L}_{p}$ is maximized in the infinite data limit with the correct number of latent features included in the model. Then, the posterior samples $\hat{\Z}_c$ obtained from the encoder $q_{\hat{\phi}^{p}}$ are optimallly prediction: $\Y|\hat{\Z}_c \stackrel{\text{dist}}{=}\Y|\X$. 
\label{lemma:pred}
\end{customlemma}
\begin{proof}[Proof of Lemma \ref{lemma:pred}]
\label{sec:proof_lemma_predictive}
We analyze the following estimator in the infinite data limit:
\begin{equation}
    \begin{aligned}
    \argmax_{\phi^p,\theta^p,f,\psi} {\mathbb{E}}_{\X,\Y \sim \mathcal{P}_{\X,\Y}}[\mathcal{L}_\texttt{p}(\phi^p,{\theta}^p,f,\psi;\X,\Y)]. 
    \end{aligned}
    \label{eqn:pVAE}
\end{equation}
The optimization program \eqref{eqn:pVAE} can be equivalently expressed as: 
\begin{equation}
    \begin{aligned}
        \argmax_{\substack{f,\psi,\phi^p,\theta^p}} ~~~
            &\underbrace{\mathbb{E}_{\X \sim\mathbb{P}_{\X}}\Bigg[ \mathbb{E}_{q_{\phi^p}(\tilde{\Z}|{\X} )}[\log{p_f}( \X|\tilde{\Z})]-\mathrm{KL}\left(q_{\phi^p}(\tilde{\Z}|{\X}),p_{\theta^p}(\tilde{\Z})\right)\Bigg]}_{\text{reconstruction loss}} \\
            & +~\underbrace{\mathbb{E}_{\X,\Y \sim\mathbb{P}_{\X,\Y}}\Bigg[\mathbb{E}_{q_{\phi^p}(\tilde{\Z}|{\X})}[\log{p}_\psi(\Y|\tzc)]\Bigg]}_{\text{classification term}}.
        \end{aligned}
        \label{PVAE_pop}
        \end{equation}  
Here, $\tilde{\Z}$ is an approximation for the true latent variables $\Z$ with $p_f(x|\tilde{\Z}) = \delta_{f(\tilde{\Z})}$. Consider maximizing the reconstruction loss in \eqref{PVAE_pop}. In this setting, the VAE model searches for an approximation $\hat{f}(\tilde{\Z}) \stackrel{\text{dist}}{\approx} \X$ where the parameters of the VAE model (e.g. posterior $\phi^{p}$, prior $\theta^{p}$, $f$) are optimized to yield the best approximation of $\X$. In other words, VAE training approximates the following optimization:
\begin{equation}
    \begin{aligned}
            \argmax_{f,\theta^p}& ~~\mathbb{E}_{\X\sim\mathbb{P}_{\X}}[\log{p_{f, \theta^p}}(\X)],
    \end{aligned}
    \label{eqn:opt_predVAE_alg}
\end{equation}
where the likelihood $p_{f,\theta^p}(\X)$ is defined with respect to the distribution $\tilde{\X} \stackrel{\text{dist}}{=} f(\tilde{\Z})$ with $\tilde{\Z}$ being a Gaussian mixture distribution with parameters $\theta^p$. 
Optimality for \eqref{eqn:opt_predVAE_alg} is achieved if $\tilde{\X} \distequal \X$, i.e. $p_{f,\theta^{p}}(\X) = \pstarX$. In particular, by definition this is achieved for $f  = f^\star$ since $\X \stackrel{\text{dist}}{=} f^\star(\Z)$. Similarly, since $\X\stackrel{\text{dist}}{=} \Bigg[f^\star \circ \begin{pmatrix}PD & 0 \\ 0 & N\end{pmatrix}\Bigg]\begin{pmatrix}PD & 0 \\ 0 & N \end{pmatrix}^{-1}\Z$ and the family of Gaussian mixture distributions is invariant to linear transformations, we can further conclude that $f = f^\star\circ \begin{pmatrix}PD & 0 \\ 0 &N\end{pmatrix}$ for any permutation matrix $P$ and diagonal matrix $D$ is also an optimum of \eqref{eqn:opt_predVAE_alg}.

Since it is hard to maximize over $p_{f,\theta^p}$ directly, 
the VAE training approach uses a surrogate for the density $p$  via the ELBO approximation. 
Specifically, recall that the ELBO is a lower bound for the log-likelihood: 
\begin{equation}
\begin{aligned}
 \log{}(\pstarX) & \geq \mathbb{E}_{q_{\phi^p}(\tilde{\Z}|\X)}[\log{p}_f(\X|\tilde{\Z})]-\mathrm{KL}\left(q_{\phi^p}(\tilde{\Z}|\X),p_{\theta^p}(\tilde{\Z})\right).
 \end{aligned}
 \label{eqn:ineq_2}
\end{equation}
Equality holds if the approximate posterior matches the true posterior, that is in the noiseless case, for any $x \in \mathcal{X}$:
\begin{equation}
\begin{aligned}
q_{\phi^p}(\tilde{\Z}|\X=x) = p_f(\tilde{\Z}|\tilde{\X}=x) = \delta_{f^{-1}(x)}.
    \end{aligned}
    \label{eqn:opt_post_2}
\end{equation} 
Here, we have appealed to one-to-one property of the function f by Assumption 1.1 and that the number of latents is specified correctly. Hence, with the choice of the posterior in \eqref{eqn:opt_post_2}, and setting $\tilde{\X} \stackrel{\text{dist}}{=} \X$, the maximization over $f,\theta^p$ of \eqref{eqn:opt_predVAE_alg} and the reconstruction loss in \eqref{PVAE_pop} are equivalent.

We finally need to verify that $\phi^p$ in \eqref{eqn:opt_post_2} leads to an optimal prediction in the classification term in \eqref{PVAE_pop}. This follows by noting that $\Y | f^{-1}(\X) \stackrel{\text{dist}}{=} \Y|\X$ according to the graphical model in Figure~\ref{fig:graph_models}, {from the optimum \eqref{eqn:opt_post_2} for $q_{\phi^p}$}, and Assumption 2.
\end{proof}

\subsubsection{Proof of Lemma~\ref{lemma:cl}: maximizing $\mathcal{L}_{cl}$ identifies the true concepts}
\label{sec:proof_lemma_concept}
We analyze the following estimator in the infinite data limit
\begin{equation}
    \begin{aligned}
    \argmax_{\phi^{cl},\theta^{cl},f}&\sum_{y \in \mathcal{Y}} \ProbY(\Y = y){\mathbb{E}}_{\X \sim \mathbb{P}_{\X|\Y=y}}[\mathcal{L}_\texttt{cl}(\phi^{cl},{\theta}^{cl},f;\X,\Y)] \, \\
     =\argmax_{\substack{\phi^{cl},\theta^{cl},f}} &\sum_{y\in\mathcal{Y}} \ProbY(\Y = y) \mathbb{E}_{\X  \sim\ProbXcY}\Bigg[ \mathbb{E}_{q_{\phi^{cl}}}(\tilde{\Z}|\X,\Y = y)}[\log{p_{f}(\X|\tilde{\Z})]\\&~~~~~~~~~~~~~~~~~~~~~~~~~~~~~~~-\mathrm{KL}\left(q_{\phi^{cl}}(\tilde{\Z}|{\X},\Y = y),p_{\theta^{cl}}(\tilde{\Z}|\Y = y)\right)\Bigg],
    \end{aligned}
    \label{disVAE_pop}
\end{equation}
where equality follows from the definition of $\mathcal{L}_\texttt{cl}$.
Here, $\tilde{\Z}$ is an approximation for the underlying latent variables $\Z$ with $\tilde{\Z}|\Y = y \sim \mathcal{N}\left(\mu_y,\begin{pmatrix}D_y & 0\\ 0 & G \end{pmatrix}\right)$, for some vector $\mu_y$ and (diagonal) matrix $D_y \in \R^{k_c^\star \times k_c^\star}$ and a general matrix $G \in \R^{k_s^\star \times k_s^\star}$, altogether accumulated in the parameter $\theta^{cl}$. Finally, we have that $p(\X|\tilde{\Z}) = \delta_{f(\tilde{Z})}$.

The proof of Lemma~\ref{lemma:cl} relies on the following lemmas, which we state below and prove later. 
\begin{customlemma}{3}
Let $a, b \in \mathbb{N}_+$. Suppose $Q\tildeD = D{Q}$ for orthogonal matrix $Q \in \mathbb{R}^{(a+b) \times (a+b)}$ and diagonal matrices $D, \tildeD$ where in the first $a$ coordinates, $D$ has unequal diagonal entries with no entry equal to one. Suppose that the last $b$ entries of $D, \tildeD$ are equal to $1$. Then, $Q$ takes the following form: $Q = \begin{pmatrix}
P & 0 \\ 0 & \tilde{Q}
\end{pmatrix}$
where $P \in \mathbb{R}^{a \times a}$ is a permutation matrix and $\tilde{Q} \in \mathbb{R}^{b \times b}$ is an orthogonal matrix. 
\label{lemma:1}
\end{customlemma}

\begin{customlemma}{4} The following two statements are equivalent:
\begin{enumerate}
    \item the parameters $(\hat{\phi}^{cl},\hat{f},\{\hat{\mu}_y,\hat{D}_y,\hat{G}\}_{y \in \mathcal{Y}})$ are optimizers of \eqref{disVAE_pop}.
    \item for all $y \in \mathcal{Y}$, $\X \stackrel{\text{dist}}{=} \hat{f}(\tilde{\Z})~;~\tilde{\Z}|\Y ={y} \sim \mathcal{N}\left(\hat{\mu}_y,\begin{pmatrix}\hat{D}_y & 0\\ 0& \hat{G} \end{pmatrix}\right)~;~ \hat{\phi}= \text{ parameters of }~\tilde{\Z}|\X,\Y=y$.
\end{enumerate}
\label{lemma:2}
\end{customlemma}
With Lemmas~\ref{lemma:1} and \ref{lemma:2} at hand, we are ready to prove Lemma~\ref{lemma:cl}.
\begin{proof}[Proof of Lemma~\ref{lemma:cl}]
Lemma~\ref{lemma:2} states that in the noiseless case $\epsilon =0$, we have $\X \stackrel{\text{dist}}{=} f^\star(\Z) \stackrel{\text{dist}}{=} \hat{f}(\tildeZ)$ for $\tildeZ|\Y = y \sim \mathcal{N}\left(\hat{\mu}_y,\begin{pmatrix}\hat{D}_y & 0\\0&\hat{G} \end{pmatrix}\right)$. We now show that the set of all possible solutions for $\fhat$, denoted by $\mathcal{H}$, is restricted to maps of the form $f^\star \circ \begin{pmatrix}PD&0\\0 & G\end{pmatrix}$ for permutation and diagonal matrices of dimension $k_c^\star \times k_c^\star$ and general $k_s^\star \times k_s^\star$ matrix $G$. 

Remember in the noiseless case $\X:= f^\star(\Z)$. We then have the following equality:
\begin{equation*}
\begin{aligned}
\mathcal{H} &= \Bigg\{\text{continuous, one-to-one }f~{\big|}~ f^{-1}(\X)|\Y = y \sim \mathcal{N}\left(\mu_y,\begin{pmatrix}D_y & 0 \\ 0 & G \end{pmatrix}\right) ;  D_y \text{ diagonal}  \text{ for all } y\in\mathcal{Y} \Bigg\} \\
&{\stackrel{(a)}=} \Bigg\{ f^\star \circ g \text{ for a continuous, one-to-one }g~{\big|}~ g^{-1}(\Z)|\Y = y \sim \mathcal{N}\left(\mu_y,\begin{pmatrix}D_y & 0 \\ 0 & G \end{pmatrix}\right); D_y \text{ diagonal}  \text{ for all } y\in\mathcal{Y} \Bigg\}, \\
      \end{aligned}
    \label{eqn:chain}
\end{equation*}
where for every $y \in \mathcal{Y}$, $D_y \in \R^{k^\star_c \times k^\star_c}$ and $G \in \R^{k^\star_s \times k^\star_s}$. Here the relation ${\stackrel{(a)}=}$ follows from $f^\star$ being one-to-one and continuous from Assumption 1.1 as well as  $\X \stackrel{\text{dist}}{=} f^\star(\Z)$. We further have
 \begin{equation}
 \begin{aligned}
\mathcal{H}  &{\stackrel{(b)}=} \Bigg\{ f^\star \circ M \text{ for invertible matrix }M~{\big|}~ M^{-1}(\Z)|\Y = y \sim \mathcal{N}\left(\mu_y,\begin{pmatrix}D_y & 0 \\ 0 & G \end{pmatrix}\right); D_y \text{ diagonal}  \text{ for all } y\in\mathcal{Y} \Bigg\} \\
   &{\stackrel{(c)}=} \Bigg\{ f^\star \circ M~{\big|}~ M = \begin{pmatrix}D_y^{\star} & 0\\ 0 & G^\star \end{pmatrix}^{1/2} Q_{y} \begin{pmatrix} D_y & 0 \\ 0 & G \end{pmatrix}^{-1/2} ;  D_y \text{ diagonal}, Q_y \text{ orthogonal}  \text{ for all } y\in\mathcal{Y} \Bigg\}.
   \end{aligned}
\end{equation}
The relation ${\stackrel{(b)}=}$ follows from the fact that the set of one-to-one continuous operators that preserve Gaussianity are linear; the relation ${\stackrel{(c)}=}$ follows from $\Z|\Y = y \sim \mathcal{N}\left(\mu_{y}^{\star}, \begin{pmatrix}D_y^\star & 0\\ 0 & G^{\star}_s\end{pmatrix} \right)$ and the following calculations:
\begin{equation}
\begin{aligned}
&~~~~M^{-1}\begin{pmatrix}D_y^\star & 0\\0 & G^\star \end{pmatrix}M^{-T} = \begin{pmatrix}D_y & 0\\0& G \end{pmatrix}\\
&\Leftrightarrow  \begin{pmatrix}D_y & 0\\0& G \end{pmatrix}^{-1/2}M^{-1}\begin{pmatrix}D_y^\star & 0\\0 & G^\star \end{pmatrix}M^{-T}\begin{pmatrix}D_y & 0\\0& G \end{pmatrix}^{-1/2} = \mathrm{Id}\\
&\Leftrightarrow \begin{pmatrix}D_y & 0\\0& G \end{pmatrix}^{-1/2}M^{-1}\begin{pmatrix}D_y^\star & 0\\0 & G^\star \end{pmatrix}^{1/2} \text{ is an orthogonal matrix}.
\end{aligned}
\end{equation}
Consider the pair $y,\tilde{y}$ satisfying Assumption 1.2. Then, {since $M$ doesn't depend on $y$,} we have that:
\begin{eqnarray}
\begin{pmatrix}D_y^\star & 0\\ 0 & G^{\star} \end{pmatrix}^{1/2} Q_y \begin{pmatrix} D_y & 0 \\ 0 & G \end{pmatrix}^{-1/2} = \begin{pmatrix}D^{\star}_{\tilde{y}} & 0\\ 0 & G^{\star} \end{pmatrix}^{1/2} {Q}_{\tilde{y}} \begin{pmatrix} D_{\tilde{y}} & 0 \\ 0 & G \end{pmatrix}^{-1/2} \, .
\label{eqn:relation}
\end{eqnarray}
Define the quantities:
\begin{eqnarray*}
A^{\star} := \begin{pmatrix}(D_y^\star)^{-1}D_{\tilde{y}}^\star  & 0\\ 0 & \mathrm{Id} \end{pmatrix}~~~;~~~ A := \begin{pmatrix}(D_{y})^{-1}D_{\tilde{y}}  & 0\\ 0 & \mathrm{Id} \end{pmatrix} \, .
\end{eqnarray*}
Then the relation \eqref{eqn:relation} reduces to the following condition:
\begin{equation}
    (A^\star)^{-1/2}Q_y A^{1/2} \text{ is an orthogonal matrix}.
    \label{eqn:relation1}
\end{equation}
The relation \eqref{eqn:relation1} leads to the conclusion:
\begin{eqnarray}
Q_y A = A^\star{Q}_{y}\, .
\label{eqn:relation2}
\end{eqnarray}
Since by Assumption 1.2, the first $k_c^\star$ diagonal elements of $A^{\star}$ are distinct, $A, A^\star$ satisfy the assumptions of Lemma~\ref{lemma:1}. This implies that 
$Q_y = \begin{pmatrix}P_y & 0 \\ 0 & \bar{Q}_y\end{pmatrix}$ for some permutation {matrices} $P_y$ and orthogonal matrices {$\bar{Q}_y$} and hence
\begin{equation*}
\mathcal{H} = \Bigg\{ f^\star \circ M~{\big|}~ M = \begin{pmatrix} PD  & 0\\ 0 & G \end{pmatrix};  D\in \R^{k^\star_c\times k^\star_c} \text{ diagonal}, P\in \R^{k^\star_c\times k^\star_c} \text{ permutation matrix}, G \in \R^{k^\star_s \times k^\star_s} \Bigg\},
\end{equation*}
which concludes the proof.
\end{proof}


\begin{proof}[Proof of Lemma~\ref{lemma:1}]
For convenience, we first decompose the matrices $Q= \begin{pmatrix} Q_1 & Q_{12} \\ Q_{21} & Q_2 \end{pmatrix}$ where $Q_1 \in \R^{a \times a}$ and similarly for $D, \tildeD$.
We first show that $Q_{12}=0$.
The relation $Q\tildeD = {D}Q$ implies that for any pair of indices $(i, j)$ either $[Q]_{ij} = 0$ or $[Q]_{ij} \neq 0 ~~\&~~ [D]_{ii} = [\tildeD]_{jj}$.
Consider $i \in \{1,2,\dots,a\}$ and $j \in \{a+1,a+2,\dots,a+b\}$. By the above relation and the assumption that the first {$a$} entries of $D$ are not equal to one and the last {$b$} entries of $\tildeD$ are equal to one, we conclude that $[Q]_{ij} = 0$. Thus, we have established that $Q_{12} =0$. Further, because $QQ^\top = I$, we have that $Q_{21}=0$.

It remains to show that $Q_1$ is diagonal. 
First note that
since $Q$ is orthogonal, the matrix $Q_1$ must be orthogonal. 
The equality $Q\tildeD = D{Q}$ now implies that $Q_1D_1 = \tildeD_1 Q_1$. In particular, $Q_1 D_1 Q_1^T = \tildeD_1$,
that is $Q_1 D_1 Q_1^T$ is an eigen-decomposition of a diagonal matrix with distinct eigenvalues.  
By the uniqueness of eigen-decompositions, $Q_1$ must therefore be a permutation matrix. 
\end{proof}
\begin{proof}[Proof of Lemma~\ref{lemma:2}] 
Consider maximizing the reconstruction loss in \eqref{disVAE_pop}. In this setting, the VAE model searches for an approximation $\hat{f}(\tilde{\Z}) \stackrel{\text{dist}}{\approx} \X$ where the parameters of the VAE model (e.g. posterior $\phi^{cl}$, prior $\theta^{cl}$, $f$) are optimized to yield the best approximation of $\X$. In other words, VAE training approximates the following optimization:
\begin{equation}
    \begin{aligned}
            \argmax_{\substack{f,\theta^{cl}}}& \sum_{y\in\mathcal{Y}}\ProbY(\Y = y)\mathbb{E}_{\X\sim\mathbb{P}_{\X|\Y = y}}[\log{p}_{f,\theta^{cl}}(\X|\Y = y)],
    \end{aligned}
    \label{eqn:opt_disVAE_alg}
\end{equation}
where the likelihood $p_{f,\theta^{cl}}(\X)$ is defined with respect to the distribution $\tilde{\X} \stackrel{\text{dist}}{=} f(\tilde{\Z})$ with $\tilde{\Z}|\Y=y$ being a Gaussian distribution with parameters $\theta^{cl}$. 
Optimality for \eqref{eqn:opt_disVAE_alg} is achieved if $\tilde{\X} \distequal \X$, i.e. $p_{f,\theta^{cl}}(\X|\Y=y) = p^\star({\X|\Y=y})$. Since it is hard to maximize over $p_{f,\theta^{cl}}$ directly, 
the VAE training approach uses a surrogate for the density $p$  via the ELBO approximation. 
Specifically, recall that the ELBO is a lower bound for the log-likelihood: 
\begin{equation*}
\begin{aligned}
 \log{}(p^\star({\X|\Y=y})) & \geq \mathbb{E}_{q_{\phi^{cl}}(\tilde{\Z}|\X)}[\log{p}_f(\X|\tilde{\Z})]-\mathrm{KL}\left(q_{\phi^{cl}}(\tilde{\Z}|\X,\Y=y),p_{\theta^{cl}}(\tilde{\Z}|\Y=y)\right).
 \end{aligned}
 \label{eqn:ineq}
\end{equation*}

Equality holds if the approximate posterior matches the true posterior, that is in the noiseless case, for any $x \in \mathcal{X}$:
\begin{equation}
\begin{aligned}
q_{\phi^{cl}}(\tilde{\Z}|\X=x,\Y=y) = p_f(\tilde{\Z}|\tilde{\X}=x,\Y=y) = \delta_{f^{-1}(x)}.
    \end{aligned}
    \label{eqn:opt_post}
\end{equation} 
where we have appealed to $f$ being one-to-one from Assumption 1.1.. Hence, with the choice of the posterior in \eqref{eqn:opt_post}, and setting $\tilde{\X} \stackrel{\text{dist}}{=} \X$, the maximization over $f,\theta^{cl}$ of \eqref{eqn:opt_disVAE_alg} and the reconstruction loss in \eqref{disVAE_pop} are equivalent.\end{proof}

\subsubsection{Combining Lemma~\ref{lemma:pred} and \ref{lemma:cl}: concept learning and prediction guarantees}
Notice the following basic inequality:
\begin{equation*}
\begin{aligned}
    &\max_{\phi^p,\theta^p,f,\psi,\phi^{cl},\theta^{cl}} {\mathbb{E}}_{\X,\Y}[\mathcal{L}_\texttt{p}(\phi^p,{\theta}^p,f,\psi;\X,\Y)]+ {\mathbb{E}}_{\X,\Y}[\mathcal{L}_\texttt{cl}(\phi^{cl},{\theta}^{cl},f;\X,\Y)]  \\
    &\leq \max_{\phi^p,\theta^p,f,\psi} {\mathbb{E}}_{\X,\Y}[\mathcal{L}_\texttt{p}(\phi^p,{\theta}^p,f,\psi;\X,\Y)]  + \max_{\phi^{cl},\theta^{cl},f}  {\mathbb{E}}_{\X,\Y}[\mathcal{L}_\texttt{cl}(\phi^{cl},{\theta}^{cl},f;\X,\Y)],
\end{aligned}
\end{equation*}
where equality holds if there exists a decoder $f$ that is optimal for the optimization problems with respect to $\mathcal{L}_{p}$ and with respect to $\mathcal{L}_{cl}$. Lemmas~\ref{lemma:pred} and \ref{lemma:cl} guarantee this to be the case. We thus can conclude that the sum of the objectives $\mathcal{L}_p+\mathcal{L}_{cl}$ inherits the
predictive power of using the architecture corresponding to $\mathcal{L}_p$ and the concept learning capabilities
of using the architecture corresponding to $\mathcal{L}_{cl}$.
\subsection{Proof of Theorem 1: miss-specified number of latent variables}
\label{sec:proof_corollary}
In this section, we analyze the regularized estimator:
\begin{equation}
    \begin{aligned}
    &\max_{\substack{{\phi}^{p},{\theta}^p,{\phi}^{cl},{\theta}^{cl}},f,\psi}  {{\hat{\mathbb{E}}}_{\X,\Y}[\mathcal{L}_p(\phi^p,{\theta}^p,f,\psi;\X,\Y)]}+{\hat{\mathbb{E}}_{\X,\Y}\allowbreak[\mathcal{L}_{cl}({\phi}^{cl},{\theta}^{cl},f;\X,\Y)]}-\lambda_n\rho(f,\psi).
    \end{aligned}
    \label{eqn:CbPVAE_regularized}
\end{equation}

For our analysis, we define a few quantities. Let $f^\star_\text{extend}$ be one-to-one extension of $f^\star$ to a domain $\mathbb{R}^{k}$ where for any $v \in \mathbb{R}^{k}$:
\begin{equation*}
    f^\star_\text{extend} \circ B^\star(v) = f^\star\begin{pmatrix}v_{1:k_c^\star} \\ v_{k_c^\star+1:k_c+k_s^\star} \end{pmatrix}~~\text{ where } B^\star = \begin{pmatrix}\id_{k_c^\star} & 0 & 0 & 0 \\
0 & 0& 0 & 0 \\ 0 & 0& \id_{k_s^\star} & 0\\ 0 & 0 & 0 & 0\end{pmatrix} \in \R^{k\times k}.
\end{equation*}
Here, $\id_{(\cdot)}$ denotes an identify matrix with its size specified in the subscript. Finally, we let $\Z_\text{extend} \in \mathbb{R}^{k}$ be a random variable that is identical to $\Z$ (true core and style features) in certain coordinates, and a standard Gaussian in other coordinates. Specifically:
\begin{equation*}
    \begin{pmatrix}\id_{k_c^\star} & 0&0&0 \\0&0&\id_{k_s^\star} & 0\end{pmatrix}\Z_\text{extend} = \Z~~; \,~~~  \begin{pmatrix}0 & \id_{k_c - k_c^\star}&0&0 \\0&0&0 & \id_{k_s -k_s^\star}\end{pmatrix}\Z_\text{extend} \text{ is standard Gaussian}.
\end{equation*}

In the infinite data limit with the regularization parameter $\lambda_n$ tending to zero with larger sample size, the optimal parameters of \eqref{eqn:CbPVAE_regularized} are solutions to :
\begin{equation}
    \begin{aligned}
   \argmin_{f,\psi,\phi^p,\theta^p,\phi^{cl},\theta^{cl}}  
    \, &~~~{\rho(f,\psi)} \\
    \text{subject-to}\, &~~~f,\psi,\phi^p,\theta^p,\phi^{cl},\theta^{cl}\in \argmax_{f,\psi,\phi^p,\theta^p,\phi^{cl},\theta^{cl}}\mathbb{E}_{\X,\Y}[\mathcal{L}_\texttt{p}(\phi^p,{\theta}^p,f,\psi;\X,\Y)] \\&~~~~~~~~~~~~~~~~~~~~~~~~~~~~~~~~~~~~~~~~~~~~~+\mathbb{E}_{\X,\Y}[\mathcal{L}_\texttt{cl}(\phi^p,{\theta}^p,f,\psi;\X,\Y)].
    \end{aligned}
    \label{eqn:CbPVAE_regul}
\end{equation}
\\
\\
The proof of the corollary requires a few lemmas which we provide next and prove later.
\begin{customlemma}{5} We have the following equivalence for a set of parameters $f,\psi,\phi^p,\theta^p,\phi^{cl},\theta^{cl}$: 
\begin{equation*}
    \begin{aligned}
    f,\psi,\phi^p,\theta^p,\phi^{cl},\theta^{cl} &\in \argmax_{f,\psi,\phi^p,\theta^p,\phi^{cl},\theta^{cl}}\mathbb{E}_{\X,\Y \sim \ProbXY}[\mathcal{L}_\texttt{p}(\phi^p,{\theta}^p,f,\psi;\X,\Y)] + \mathbb{E}_{\X,\Y \sim \ProbXY}[\mathcal{L}_\texttt{cl}(\phi^p,{\theta}^p,f,\psi;\X,\Y)] \\
    &\Leftrightarrow\\
     f,\phi^p,\theta^p,\phi^{cl},\theta^{cl} &\in \argmax_{\substack{\phi^{p},\theta^{p},\phi^{cl},\theta^{cl},f}} \mathbb{E}_{\X  \sim\mathbb{P}_{\X}}\Bigg[
    \mathbb{E}_{q_{\phi^{p}}(\tilde{\Z}|\X)}[\log{p}_f(\X|\tilde{\Z})]-\mathrm{KL}\left(q_{\phi^p}(\tilde{\Z}|{\X}),p_{\theta^{p}}(\tilde{\Z})\right)\Bigg]\\&~~~~~~~~~~~~~~~+\mathbb{E}_{\X, \Y  \sim\mathbb{P}_{\X,\Y}}\Bigg[ \mathbb{E}_{q_{\phi^{cl}}(\tilde{\Z}|\X,\Y)}[\log{p}_f(\X|\tilde{\Z})]-\mathrm{KL}\left(q_{\phi^{cl}}(\tilde{\Z}|{\X},\Y),p_{\theta^{cl}}(\tilde{\Z}|\Y)\right)\Bigg]\\
    \psi,\phi^{p} &\in \argmax_{\psi,\phi^p}\mathbb{E}_{\X, \Y  \sim\mathbb{P}_{\X,\Y}}\Bigg[ \mathbb{E}_{q_{\phi^{p}}(\tilde{\Z}|\X)}[\log{p}_\psi(\Y|\tzc)]\Bigg]
    \end{aligned}
\end{equation*}
\label{lemma:prediction_same}
\end{customlemma}
\begin{customlemma}{6}We have the following implication for the parameters $\phi^{p},\theta^{p},\phi^{cl},\theta^{cl},f$: 
\begin{equation*}
\begin{aligned}
&\phi^{p},\theta^{p},\phi^{cl},\theta^{cl},f \in \argmax_{\substack{\phi^{p},\theta^{p},\phi^{cl},\theta^{cl},f}}\,~~~~ \mathbb{E}_{\X  \sim\mathbb{P}_{\X}}\Bigg[
    \mathbb{E}_{q_{\phi^{p}}(\tilde{\Z}|\X)}[\log{p}_f(\X|\tilde{\Z})]-\mathrm{KL}\left(q_{\phi^{p}}(\tilde{\Z}|{\X}),p_{\theta^{p}}(\tilde{\Z})\right)\Bigg]\\{\,}&~~~~~~~~~~~~~~~~~~~~~~~~~~~~~~~~~~~~~~~~+\mathbb{E}_{\X, \Y  \sim\mathbb{P}_{\X,\Y}}\Bigg[ \mathbb{E}_{q_{\phi^{cl}}(\tilde{\Z}|\X,\Y)}[\log{p}_f(\X|\tilde{\Z})]-\mathrm{KL}\left(q_{\phi^{cl}}(\tilde{\Z}|{\X},\Y),p_{\theta^{cl}}(\tilde{\Z}|\Y)\right)\Bigg] \\
    & \psi,\phi^{p} \in \argmax_{\psi,\phi^p}\mathbb{E}_{\X, \Y  \sim\mathbb{P}_{\X,\Y}}\Bigg[ \mathbb{E}_{q_{\phi^{p}}(\tilde{\Z}|\X)}[\log{p}_\psi(\Y|\tzc)]\Bigg]\\
    &\Rightarrow~~\text{for every }y \in \mathcal{Y}~~~f = f' \circ {B}, \X \stackrel{\text{dist}}{=} f(\tilde{\Z}) \text{ where } \tilde{\Z}|\Y=y \sim \mathcal{N}\left({\mu}_y,\begin{pmatrix} {D}_y & 0\\ 0 & {G} \end{pmatrix}\right) \text{ and }\\
    &~~~~~~{D}_y \in \R^{k_c \times k_c}, G \in \R^{k_s \times k_s}~~:~~q_{\phi^p}(\tilde{\Z}|\X) = p(\tilde{\Z}|\X), q_{\phi^\text{cl}}(\tilde{\Z}|\X,\Y=y) = p(\tilde{\Z}|\X,\Y=y). 
    \end{aligned}
\end{equation*}
\label{lemma:feasible}
\end{customlemma}
\begin{customlemma}{7} Consider the following optimization problem: 
\begin{equation*}
\begin{aligned}
(f'_\text{opt},B_\text{opt}) = \argmin_{f' \text{one-to-one},B} \, &~~~ \sum_{i = 1}^{k} \mathbb{I}(\|B_{:,i}\|_2>0) \\
\text{subject-to} \, &~~~ \text{ there exists a random vector }\tilde{\Z} \in \mathbb{R}^{k} \text{ with } \X \stackrel{\text{dist}}{=} f' \circ B(\tilde{\Z}) \\\,&~~~~\tilde{\Z}|\Y = y \sim \mathcal{N}\left({\mu}_y,\begin{pmatrix} {D}_y & 0\\ 0 & {G} \end{pmatrix}\right) \text{for all }y \in \mathcal{Y}
\end{aligned}
\end{equation*}
Then, the following statements hold:
\begin{enumerate}
    \item Any feasible $B$ has $k^\star$ (nonzero) linearly independent columns.
    \item Any optimal $B_\text{opt}$ has exactly $k^\star$ columns that are nonzero and linearly independent.
    \item Letting $B_\text{opt.reduced}$ be those nonzero columns, then,
   $f_\text{opt}'\circ B_\text{opt.reduced} = f^\star_\text{extend} \circ \begin{pmatrix}PD & 0 &0&0\\ 0 & 0 & H & 0\end{pmatrix}^T$ for some diagonal matrix $D \in \mathbb{R}^{k_c^\star \times k_c^\star}$, permutation matrix $P \in \mathbb{R}^{k_c^\star\times{k}_c^\star}$ and non-singular matrix $H \in \mathbb{R}^{k_s^\star \times k_s^\star}$.
    \item Letting $\tilde{\Z}_\text{reduced}$ be the latent features corresponding to nonzero columns in $B$, \\$\tilde{\Z}_\text{reduced}|\Y=y \sim \mathcal{N}\left(\mu_y,\begin{pmatrix} D^{-1}P^{-1}D^\star_yP^{-T}D^{-1} & 0 \\ 0 & H^{-1}G^\star{H}^{-T}\end{pmatrix}\right)$ for all $y \in \mathcal{Y}$. 
\end{enumerate}
\label{lemma:sparsest}
\end{customlemma}
\begin{proof}[Proof of Theorem~\ref{thm:PDVAE}] 
Throughout the proof, we take $f = f' \circ B$ and $\psi = \psi' \circ C$. We consider the following reformulation of the minimal value of the optimization \eqref{eqn:CbPVAE_regul}: 
\begin{equation*}
\begin{aligned}
    l^{(1)} := \min \,&~~~t\\\text{subject-to }\,&~~~ 
    \rho(f,\psi)\leq{t}\\
    \,&~~~ f,\psi,\phi^p,\theta^p,\phi^{cl},\theta^{cl}\in \argmax_{f,\psi,\phi^p,\theta^p,\phi^{cl},\theta^{cl}}\mathbb{E}_{\X,\Y \sim \ProbXY}[\mathcal{L}_\texttt{p}(\phi^p,{\theta}^p,f,\psi;\X,\Y)]\\\,&~~~~~~~~~~~~~~~~~~~~~~~~~~~~~~~~~~~~~~~~~~~~~~ + \mathbb{E}_{\X,\Y \sim \ProbXY}[\mathcal{L}_\texttt{cl}(\phi^p,{\theta}^p,f,\psi;\X,\Y)],
\end{aligned}
\label{eqn:first_constraint}
\end{equation*}
where we denote the set of feasible parameters $(t,f,\psi,\phi^p,\theta^p,\phi^{cl},\theta^{cl})$ of the optimization problem above by $\mathcal{S}^{(1)}$. Our objective is to show that any optimal encoder produces core features that are permutation and scaling of the true core features.  To that end, we  consider the following optimization problem:
\begin{equation*}
\begin{aligned}
    l^{(2)} := \min \,&~~~t\\\text{subject-to }\,&~~~ 
    \rho(f,\psi)\leq{t}\\
    \,&~~~ f,\phi^p,\theta^p,\phi^{cl},\theta^{cl} \in \argmax_{f,\phi^p,\theta^p,\phi^{cl},\theta^{cl}} \mathbb{E}_{\X  \sim\mathbb{P}_{\X}}\Bigg[
    \mathbb{E}_{q_{\phi^{p}}(\tilde{\Z}|\X)}[\log{p}_f(\X|\tilde{\Z})]-\mathrm{KL}\left(q_{\phi^p}(\tilde{\Z}|{\X}),p_{\theta^{p}}(\tilde{\Z})\right)\Bigg]\\{\,}&~~~~~~~~~~~~~~~~~~~~~~~~+\mathbb{E}_{\X,\Y  \sim\mathbb{P}_{\X,\Y}}\Bigg[ \mathbb{E}_{q_{\phi^{cl}}(\tilde{\Z}|\X,\Y)}[\log{p}_f(\X|\tilde{\Z})]-\mathrm{KL}\left(q_{\phi^{cl}}(\tilde{\Z}|{\X},\Y),p_{\theta^{cl}}(\tilde{\Z}|\Y)\right)\Bigg]\\
    {\,}&~~~ \psi,\phi^{p} \in \argmax_{\psi,\phi^p}\mathbb{E}_{\X,\Y  \sim\mathbb{P}_{\X,\Y}}\Bigg[ \mathbb{E}_{q_{\phi^{p}}(\tilde{\Z}|\X)}[\log{p}_\psi(\Y|\tilde{\Z})]\Bigg],
\end{aligned}
\end{equation*}
where the set of feasible parameters $(t,f,\psi,\phi^p,\theta^p,\phi^{cl},\theta^{cl})$ are denoted by $\mathcal{S}^{(2)}$. By Lemma~\ref{lemma:prediction_same}, we have that $\mathcal{S}^{(2)} =\mathcal{S}^{(1)}$ so that $l^{(2)}=l^{(1)}$. We then relax the constraint set as follows:
\begin{equation*}
\begin{aligned}
    l^{(3)} := \min \,&~~~t\\\text{subject-to }\,&~~~ 
    \sum_{i = 1}^{k}\mathbb{I}(\|B_{:,i}\|_2>0)\leq{t}\\
    \,&~~~ f,\phi^p,\theta^p,\phi^{cl},\theta^{cl} \in \argmax_{f,\phi^p,\theta^p,\phi^{cl},\theta^{cl}} \mathbb{E}_{\X  \sim\mathbb{P}_{\X}}\Bigg[
    \mathbb{E}_{q_{\phi^{p}}(\tilde{\Z}|\X)}[\log{p}_f(\X|\tilde{\Z})]-\mathrm{KL}\left(q_{\phi^p}(\tilde{\Z}|{\X}),p_{\theta^{p}}(\tilde{\Z})\right)\Bigg]\\{\,}&~~~~~~~~~~~~~~~~~~~~~~~~+\mathbb{E}_{\X,\Y  \sim\mathbb{P}_{\X,\Y}}\Bigg[ \mathbb{E}_{q_{\phi^{cl}}(\tilde{\Z}|\X,\Y)}[\log{p}_f(\X|\tilde{\Z})]-\mathrm{KL}\left(q_{\phi^{cl}}(\tilde{\Z}|{\X},\Y),p_{\theta^{cl}}(\tilde{\Z}|\Y)\right)\Bigg]\\
     {\,}&~~~ \psi,\phi^{p} \in \argmax_{\psi,\phi^p}\mathbb{E}_{\X,\Y  \sim\mathbb{P}_{\X,\Y}}\Bigg[ \mathbb{E}_{q_{\phi^{p}}(\tilde{\Z}|\X)}[\log{p}_\psi(\Y|\tilde{\Z})]\Bigg],
\end{aligned}
\end{equation*}
where the set of feasible parameters $(t,f,\psi,\phi^p,\theta^p,\phi^{cl},\theta^{cl})$ are denoted by $\mathcal{S}^{(3)}$. Evidently, $\mathcal{S}^{(3)} \supseteq \mathcal{S}^{(2)} \supseteq \mathcal{S}^{(1)}$ and thus $l^{(1)} \geq l^{(2)} \geq l^{(3)}$. Let $\mathcal{S}^{(3)}_{\text{opt}}$ be the optimal set of parameters in $\mathcal{S}^{(3)}$. Appealing to Lemma~\ref{lemma:feasible} and Lemma~\ref{lemma:sparsest}, the set $\mathcal{S}^{(3)}_\text{opt}$ given by:
\begin{equation}
\begin{aligned}
    \mathcal{S}_\text{opt}^{(3)} = \Bigg\{&(t,f,\psi,\phi^p,\theta^p,\phi^{cl},\theta^{cl})~|~ t = k, \text{there exists a random vector }\tilde{\Z} \in \R^{k} ~ \text{ and matrices } P,D,H \text{ s.t. }\\&\X \stackrel{\text{dist}}{=} f'\circ{B}\tilde{\Z}~~,~~ f'\circ{B}_\text{reduced} = f^\star_\text{extend}\begin{pmatrix}PD & 0 &0&0\\ 0 & 0 & H & 0\end{pmatrix}^T \text{ where } B_\text{reduced} \in \R^{k\times{k}^\star}\\
    &\theta^{cl} \text{ parameters of the Gaussian random vector }\tilde{\Z}|\Y = y \text{ where: }\\
    &~~~~\tilde{\Z}_\text{reduced}|\Y=y \sim \mathcal{N}\left(\mu_y,\begin{pmatrix} D^{-1}P^{-1}D^\star_yP^{-T}D^{-1} & 0 \\ 0 & H^{-1}G^\star{H}^{-T}\end{pmatrix}\right)\\
    &\theta^{p} \text{ parameters of the distribution of }\tilde{\Z}~~,~~q_{\phi^p}(\tilde{\Z}|\X) = q_{\phi^{cl}}(\tilde{\Z}|\X,\Y) = p(\tilde{\Z}|\X) \\
    &\psi \in \argmax \mathbb{E}_{\X,\Y  \sim\mathbb{P}_{\X,\Y}}\Bigg[ \mathbb{E}_{q_{\phi^{p}}(\tilde{\Z}|\X)}[\log{p}_\psi(\Y|\tzc)]\Bigg]\Bigg\}.
\end{aligned}
\label{eqn:choice_params}
\end{equation}
Here, $P$ is a $k_c^\star \times k_c^\star$ permutation matrix, $D$ is a $k_c^\star 
\times k_c^\star$ diagonal matrix, and $H$ is a $k_s^\star \times k_s^\star$ non-singular matrix. Furthermore, $B_\text{reduced}$ is the $k^\star$ nonzero columns of $B$ and $\tilde{\Z}_\text{reduced}$ are the components of $\tilde{\Z}$ corresponding to the nonzero columns of $B$. Take any optimal set of parameters in \eqref{eqn:choice_params}. Noting that $\X \distequal f^\star_\text{extend}B^\star\Z_\text{extend}$, it is straightforward to check that the first $k_c^\star$ components $\tilde{\Z}_\text{reduced}|\X$ are a permutation and linear scaling of $\zc$, and thus the posterior samples are optimally predictive. As such, one possibility for an optimal predictor is $\psi = \psi'\circ C$ where $C$ has the same nonzero columns as $B$ in \eqref{eqn:choice_params}. Notice that the resulting set of parameters {is} feasible in the set $\mathcal{S}^{(1)}$ and yield the objective value $t = k$. In other words, we have shown that $l^{(1)} = l^{(2)} = l^{(3)} = k$. 

Now let $\mathcal{S}^{(1)}_\text{opt}$ be the optimal set of parameters associated with $l^{(1)}$. By Lemmas~\ref{lemma:prediction_same}, \ref{lemma:feasible} and \ref{lemma:sparsest}, any optimal $B$ should have at least $k$ nonzero linearly independent columns. Thus, to attain the lower bound $l^{(1)} = k^\star$, the optimal $B$ should indeed only have $k^\star$ nonzero columns. This observation implies that the constraint $\sum_{i = 1}^{k}\mathbb{I}(\|B_{:,i}\|_2>0)\leq{t}$ can be added to feasibility set $\mathcal{S}^{(1)}$ without changing the optimal value. Thus, we have concluded that $\mathcal{S}^{(1)}_\text{opt} \subseteq \mathcal{S}^{(4)}_\text{opt}$. Since any parameters in \eqref{eqn:choice_params} lead to the posterior samples $\begin{pmatrix}\id_{k_c}&0\end{pmatrix}\tilde{\Z}_\text{reduced}|\X$ that are a permutation and linear scaling of the samples of $\zc$, we have the desired result. \end{proof}
We now prove the Lemmas~\ref{lemma:prediction_same},\ref{lemma:feasible}, and \ref{lemma:sparsest} that were used in the proof of the corollary. 
\begin{proof}[Proof of Lemma~\ref{lemma:prediction_same}]
The direction $\leftarrow$ follows in a straightforward manner. For the direction $\rightarrow$, we introduce some notation. Define:
\begin{equation*}
\begin{aligned}
    h(f,\phi^p,\theta^p,\phi^{cl},\theta^{cl}) &:= \mathbb{E}_{\X  \sim\mathbb{P}_{\X}}\Bigg[
    \mathbb{E}_{q_{\phi^{p}}(\tilde{\Z}|\X)}[\log{p}_f(\X|\tilde{\Z})]-\mathrm{KL}\left(q_{\phi^p}(\tilde{\Z}|{\X}),p_{\theta^{p}}(\tilde{\Z})\right)\Bigg]\\&~~~~~~~~~~~~~~~+\mathbb{E}_{\X, \Y  \sim\mathbb{P}_{\X,\Y}}\Bigg[ \mathbb{E}_{q_{\phi^{cl}}(\tilde{\Z}|\X,\Y)}[\log{p}_f(\X|\tilde{\Z})]-\mathrm{KL}\left(q_{\phi^{cl}}(\tilde{\Z}|{\X},\Y),p_{\theta^{cl}}(\tilde{\Z}|\Y)\right)\Bigg]\\
    g(\phi^p,\psi) &:= \mathbb{E}_{\X, \Y  \sim\mathbb{P}_{\X,\Y}}\Bigg[ \mathbb{E}_{q_{\phi^{p}}(\tilde{\Z}|\X)}[\log{p}_\psi(\Y|\tilde{\Z})]\Bigg]
\end{aligned}
\end{equation*}
Note that $\mathbb{E}_{\X,\Y \sim \ProbXY}[\mathcal{L}_\texttt{p}(\phi^p,{\theta}^p,f,\psi;\X,\Y)] + \mathbb{E}_{\X,\Y \sim \ProbXY}[\mathcal{L}_\texttt{cl}(\phi^p,{\theta}^p,f,\psi;\X,\Y)] = h(f,\phi^p,\theta^p,\phi^{cl},\theta^{cl})+ g(\phi^p,\psi)$. Thus, we have the following inequality:
\begin{equation}
\begin{aligned}
    &\max_{f,\psi,\phi^p,\theta^p,\phi^{cl},\theta^{cl}}\mathbb{E}_{\X,\Y \sim \ProbXY}[\mathcal{L}_\texttt{p}(\phi^p,{\theta}^p,f,\psi;\X,\Y)] + \mathbb{E}_{\X,\Y \sim \ProbXY}[\mathcal{L}_\texttt{cl}(\phi^p,{\theta}^p,f,\psi;\X,\Y)]\\
    &\leq \max_{\substack{\phi^{p},\theta^{p},\phi^{cl},\theta^{cl},f}}  h(f,\phi^p,\theta^p,\phi^{cl},\theta^{cl})+\max_{\psi,\phi^p} g(\phi^p,\psi).
\end{aligned}
\label{eqn:inequality_final}
\end{equation}
Consider the parameters in \eqref{eqn:choice_params}. Notice that they optimize each term in the right hand side of the inequality above and are feasible in the optimization problem in left hand side of the inequality. Thus, the inequality in \eqref{eqn:inequality_final} is actually an equality. Let $t_\text{opt}$ be the optimal value of either side of the equality.

Suppose for a proof of contradiction that the direction $\rightarrow$ is not valid. In other words, consider a set of maximizers  $(f_\text{opt},\psi_\text{opt},\phi^p_\text{opt},\theta^p_\text{opt},\phi^{cl}_\text{opt},\theta^{cl}_\text{opt})$ for the left hand side of the equation above that are not maximal in either term in the right hand side. Then:
\begin{equation*}
\begin{aligned}
    t_\text{opt} &= h(f_\text{opt},\phi^p_\text{opt},\theta^p_\text{opt},\phi^{cl}_\text{opt},\theta^{cl}_\text{opt})+ g(\phi^p_\text{opt},\psi_\text{opt})\\
    &< \max_{\substack{\phi^{p},\theta^{p},\phi^{cl},\theta^{cl},f}}  h(f,\phi^p,\theta^p,\phi^{cl},\theta^{cl})+\max_{\psi,\phi^p} g(\phi^p,\psi).
\end{aligned}
\end{equation*}
This however contradicts the fact that the inequality \eqref{eqn:inequality_final} is an equality.
\end{proof}
\begin{proof}[Proof of Lemma~\ref{lemma:feasible}] We have the following inequality:
\begin{equation*}
\begin{aligned}
    &\argmax_{f,\phi^p,\theta^p,\phi^{cl},\theta^{cl}}\mathbb{E}_{\X  \sim\mathbb{P}_{\X}}\Bigg[
    \mathbb{E}_{q_{\phi^{p}}(\tilde{\Z}|\X)}[\log{p}_f(\X|\tilde{\Z})]-\mathrm{KL}\left(q_{\phi^p}(\tilde{\Z}|{\X}),p_{\theta^{p}}(\tilde{\Z})\right)\Bigg]\\{\,}&~~~+\mathbb{E}_{\X,\Y \sim\mathbb{P}_{\X,\Y}}\Bigg[ \mathbb{E}_{q_{\phi^{cl}}(\tilde{\Z}|\X,\Y)}[\log{p}_f(\X|\tilde{\Z})]-\mathrm{KL}\left(q_{\phi^{cl}}(\tilde{\Z}|{\X},\Y),p_{\theta^{cl}}(\tilde{\Z}|\Y)\right)\Bigg] \\
    &\leq \underbrace{\argmax_{\substack{\phi^{p},\theta^{p},f}} \mathbb{E}_{\X  \sim\mathbb{P}_{\X}}\Bigg[
    \mathbb{E}_{q_{\phi^{p}}(\tilde{\Z}|\X)}[\log{p}_f(\X|\tilde{\Z})]-\mathrm{KL}\left(q_{\phi^p}(\tilde{\Z}|{\X}),p_{\theta^{p}}(\tilde{\Z})\right)\Bigg]}_\text{Term 1}\\
    &+\underbrace{\argmax_{\substack{\phi^{cl},\theta^{cl},f}}  \mathbb{E}_{\X,\Y  \sim\mathbb{P}_{\X,\Y}}\Bigg[ \mathbb{E}_{q_{\phi^{cl}}(\tilde{\Z}|\X,\Y)}[\log{p}_f(\X|\tilde{\Z})]-\mathrm{KL}\left(q_{\phi^{cl}}(\tilde{\Z}|{\X},\Y),p_{\theta^{cl}}(\tilde{\Z}|\Y)\right)\Bigg]}_\text{Term 2}.\\
\end{aligned}
\end{equation*}
 Notice that the different terms in the equation above have the common parameters $f$. Thus, the inequality in the equation above is an equality if there exists an $f$ that is optimal for each of the terms in the relation above. Furthermore, from proof of Lemma \ref{lemma:pred}, Term 1 is optimized when $\X \stackrel{\text{dist}}{=} f(\tilde{\Z})$ for $\tilde{\Z}$ being a mixture Gaussian and $q_{\phi^p}(\tilde{\Z}|\X) = p(\tilde{\Z}|\X)$. From Lemma \ref{lemma:2}, Term 2 is maximized when $\X \stackrel{\text{dist}}{=} f(\tilde{\Z})$ for $\tilde{\Z}|\Y=y$ being a Gaussian distribution with appropriate covariance matrix. Consider the parameters in \eqref{eqn:choice_params}; they satisfy the properties above for the same decoder. 
 
It remains to check that there exists a set of parameters in the right hand implication of Lemma~\ref{lemma:feasible} that are optimally predictive. This follows from taking the parameters \eqref{eqn:choice_params} and noting that that the posterior samples $\tilde{\Z}|\X$ are optimally predictive. 
\end{proof}
\begin{proof}[Proof of Lemma~\ref{lemma:sparsest}]
By definition, any feasible $f = f'\circ{B}$ satisfies:
\begin{equation*}
    \X  \stackrel{\text{dist}}{=} {f}' \circ {B} (\tilde{\Z})~~\text{and}\, \tilde{\Z}|\Y = y \sim \mathcal{N}\left({\mu}_y,\begin{pmatrix} {D}_y & 0\\ 0 & {G} \end{pmatrix}\right) \text{ where }D_y \in \R^{k_c \times k_c} \text{ and }G \in \R^{k_s \times k_s}.
\end{equation*}
\paragraph{Proof of 1.}
Applying $(f^\star_\text{extend})^{-1}$ to both sides of the relation $\X  \stackrel{\text{dist}}{=} {f}' \circ {B}\tilde{\Z}$ and noting that $f'$ by Assumption 1.1 can be expressed as $f' = f^\star_\text{extend} \circ {g}$ for a continuous and one-to-one function ${g}$, we have:
\begin{equation*}
    B^\star\Z_\text{extend} \stackrel{\text{dist}}{=} {g} \circ {B}\tilde{\Z}.
\end{equation*}
Since $\Z|\Y = y$ and $\tilde{\Z}|\Y=y$ are Gaussian and ${g}$ is a continuous and one-to-one function, ${g}$ must be a linear map; we denote $g$ by the matrix $N \in \R^{k \times k}$. By the definition of $B^\star$, it is straightforward then to argue that ${B}$ must have $k^\star$ total number of linearly independent columns; these linearly independent columns are nonzero by definition. We have proven the first item in Lemma~\ref{lemma:sparsest}.

\paragraph{Proof of 2.} Gathering all the facts so far, we have that for any feasible $f = f' \circ B$:
\begin{equation}
    f' = f^\star_\text{extend} \circ N ~~~;~~~ B^\star\Z_\text{extend} \stackrel{\text{dist}}{=} NB\tilde{\Z}~~;~~ \text{rank}(B) = k_c+k_s.
    \label{eqn:relations_exact}
\end{equation}
Furthermore, by the objective of \eqref{eqn:CbPVAE_regul}, we have that an optimal $B$ must have exactly $k = k_c+k_s$ total number of linearly independent nonzero columns (i.e. $k$ of the $k$ latent features have some visualization power). We have thus concluded the second item in the lemma.

\paragraph{Proof of 3.} Let $B_\text{reduced} \in \mathbb{R}^{k \times k^\star}$ be the non-zero columns of $B$, so that $B\tilde{\Z} = B_\text{reduced}\tilde{\Z}_\text{reduced}$ where $\tilde{\Z}_\text{reduced} \in \mathbb{R}^{k}$ and for every $y \in \mathcal{Y}$:
\begin{equation*}
   \tilde{\Z}_\text{reduced}|\Y = y \sim \mathcal{N}\left({\mu}_y,\begin{pmatrix}{D}_y & 0 \\ 0 & {G} \end{pmatrix}\right). 
\end{equation*}
Here, ${D}_y$ is a diagonal matrix. Without loss of generality, we assume that every diagonal entry in ${D}_y$ has some variation across $y \in \mathcal{Y}$; otherwise, we can concatenate the components that do not vary to the general matrix $G$. The dimension of ${D}_y$ and $G$ corresponds to the number of core and style features that are selected by the nonzero columns of $B$. Note that so far, we have only established that the dimensions of $D_y$ and $G$ sum up to $k$. In what follows we show that the numbers of estimated core and style features equal to $k_c^\star, k_s^\star$.

Let $M$ be the $k^\star \times k^\star$ matrix $M:= \begin{pmatrix}\id_{k_c} & 0 & 0&0 \\ 0 & 0 & \id_{k_s} & 0\end{pmatrix}NB_\text{reduced}$. By relation \eqref{eqn:relations_exact}, $\Z \stackrel{\text{dist}}{=}M\tilde{\Z}_\text{reduced}$. Since the distribution of $\Z$ is non-degenerate, $M$ is a non-singular matrix. Therefore, for every $y\in\mathcal{Y}$ we have 
\begin{equation*}
\begin{aligned}
    M^{-1}\Z|\Y = y \sim \mathcal{N}\left(\mu_y,\begin{pmatrix} {D}_y & 0 \\ 0 & {G}\end{pmatrix}\right). 
\end{aligned}
\end{equation*}
Following a similar analysis as in the proof of Lemma~\ref{lemma:cl}, we conclude that  for every $y \in \mathcal{Y}$, there exists an orthogonal matrix $Q_y$ 
such that
\begin{equation}
\begin{aligned}
 &   M = \begin{pmatrix} D_y^\star & 0 \\ 0 & G^\star \end{pmatrix}^{1/2} Q_y \begin{pmatrix} {D}_y & 0 \\ 0 & {G}\end{pmatrix}^{-1/2}.
    \end{aligned}
    \label{rel_1}
\end{equation}
Choosing $y,\tilde{y} \in \mathcal{Y}$ 
that satisfy Assumption 1.2, relation \eqref{rel_1} implies that
\begin{equation}
    {A^\star}^{-1/2}{Q}_yA^{1/2} \text{ is an orthogonal matrix}.
    \label{rel_3}
\end{equation}
Here, $A^\star:=\begin{pmatrix} A^\star_1 & 0 \\ 0 &A^\star_2\end{pmatrix}$ where $A^\star_1 = D_y^\star[D_{\tilde{y}}^\star]^{-1}$ and $A^\star_2 = \id$ and $A = \begin{pmatrix}A_1 & 0 \\ 0 & A_2 \end{pmatrix}$, where $A_1 = {D}_y{D}_{\tilde{y}}^{-1}$ and $A_2 = \id$. Relation \eqref{rel_3} implies that:
\begin{equation*}
    Q_yA = A^\star{Q}_y.
\end{equation*}
Notice that the singular values of $Q_yA$ are equal to the singular values of $A$, since product by orthogonal matrices preserves the singular values. Similarly, singular values of $A^\star{Q}_y$ are equal to the singular values of $A^\star$. Thus, $A^\star$ and $A$ have the same singular values. In other words, since $A^\star$ has $k_s^\star$ singular values equal to one and $k_c^\star$ singular values not equal to one (by Assumption 1.2), $A_1$ must have exactly $k_c^\star$ diagonal elements not equal to one. This allows us to conclude that the dimensions of $D_y$ are greater than or equal to $k_c^\star$.  

From the analysis above, we can partition the matrix $A$ as follows: $\begin{pmatrix}\tilde{A}_1 & 0 \\ 0 & \id \end{pmatrix}$ where $\tilde{A}_1$ is a diagonal matrix of dimension $k_c^\star \times k_c^\star$ with all distinct entries. Appealing to Lemma~\ref{lemma:1}, we then conclude that $Q_y = \begin{pmatrix}P & 0 \\ 0 & H\end{pmatrix}$ for a permutation matrix $P \in \R^{k_c^\star\times{k}_c^\star}$, diagonal matrix $D \in \R^{k_c^\star \times k_c^\star}$ and a non-singular matrix $H \in \R^{k_s^\star \times k_s^\star}$. Combining this with the expression of $M$ \eqref{rel_1} and the fact that the dimensions of $D_y$ are greater than or equal to $k_c^\star$, we conclude that $M$ takes the form:
\begin{equation*}
    M = \begin{pmatrix}PD & 0 \\0&H \end{pmatrix}.
\end{equation*}
Finally, combining the relation above with \eqref{eqn:relations_exact} and the fact that $\begin{pmatrix} 0 & \id_{k_c-k_c^\star} & 0 & 0\\0 & 0&0&\id_{k_s-k_s^\star}\end{pmatrix}NB = 0$ yields the third item of the lemma. 
\paragraph{Proof of 4.} The final component of the lemma also follows from the relation $\Z \stackrel{\text{dist}}{=}M\tilde{\Z}_\text{reduced}$.
\end{proof}

\subsection{Analysis of $\mathcal{L}_{cl}$ in the noisy setting}
\label{thm:noisy}
In this section, our objective is to show that the density $q_{\phi^{cl}}(\tilde{\Z}|\X,\Y)$ matches the density $p^\star({PD\zc\allowbreak|\X,\Y})$ for some $k_c^\star \times k_c^\star$ permutation matrix and $k_c^\star \times k_c^\star$ diagonal matrix. Throughout the following discussion, $p^\star(\cdot)$ represents the ground truth density corresponding to a specified random variable.

In addition to Assumptions 1, analysis of the noisy setting requires the following mild assumption:

\begin{equation*}
    \begin{aligned}
         \text{Assumption 3 : the Fourier transform of the density of } \epsilon  \text{ is non-negative everywhere}.
    \end{aligned}
\end{equation*}

It is straightforward to extend the characterization in Lemma~\ref{lemma:2} to the noisy case and conclude that $\X \stackrel{\text{dist}}{=} f^\star(\Z)+\epsilon \stackrel{\text{dist}}{=} {f}(\tilde{\Z})+\epsilon$ where $\tilde{\Z}|\Y = y \sim \mathcal{N}\left({\mu}_y,\begin{pmatrix}{D}_y & 0\\0&{G} \end{pmatrix}\right)$. Here, $D_y$ is a $k_c^\star \times k_c^\star$ diagonal matrix and $G$ is a $k_s^\star \times k_s^\star$ matrix. Since $\tilde{\Z} \independent \epsilon$, $\Z \independent \epsilon$, we have that:
\begin{equation*}
\begin{aligned}
    \mathcal{F}[p({{f}^\star(\Z)+\epsilon})] &= \mathcal{F}[p(\epsilon)]\mathcal{F}[p({f^\star(\Z)})] \\
    \mathcal{F}[p({{f}(\tilde{\Z})+\epsilon})] &= \mathcal{F}[p(\epsilon)]\mathcal{F}[p({{f}(\tilde{\Z}))}], 
\end{aligned}
\end{equation*}
where $\mathcal{F}[\cdot]$ represents the Fourier transform and $p({\cdot})$ represents the density function with respect to a random variable. Since $p({{f}^\star(\Z)+\epsilon}) = p({{f}(\tilde{\Z})+\epsilon})$, appealing to Assumption 3, we have that $\mathcal{F}[p({{f}^\star({\Z})})] = \mathcal{F}[p({{f}(\tilde{\Z})})]$, or equivalently, ${f}(\tilde{\Z}) \stackrel{\text{dist}}{=}f^\star(\Z)$. We then have from the chain of equalities in \eqref{eqn:chain} that ${f} = f^\star \circ \begin{pmatrix}PD & 0 \\ 0 & N \end{pmatrix}^{-1}$. Combining this with the fact that ${f}(\tilde{\Z}) \stackrel{\text{dist}}{=} f^\star(\Z)$, we conclude that $\tilde{\Z} \stackrel{\text{dist}}{=} \begin{pmatrix}PD & 0 \\ 0 & N \end{pmatrix}\Z$. Notice that:
\begin{equation}
\begin{aligned}
    {p}(\tilde{z}_c|x,y) &= \int {p}(\tilde{z}_c,\tilde{z}_s|x,y)\partial{z_s}
    {\stackrel{(a)}=}\frac{\int {p}_f(x|\tilde{z}_c,\tilde{z}_s){p}(\tilde{z}_c|y){p}(\tilde{z}_s)\partial{\tilde{z}_s}}{p^\star(x|y)} 
    {\stackrel{(b)}=} \frac{\int {p}_f\left(x|\tilde{z}_c;N{z}_s\right){p}(\tilde{z}_c|y)p^\star({z}_s)\partial{{z}_s}}{p^\star(x|y)},
    \end{aligned}
    \label{posterior_approx}
\end{equation}
The equality ${\stackrel{(a)}=}$ follows from $\zc \independent \zs$ and the equality ${\stackrel{(b)}=}$ follows from the change of variables $N{z}_s \leftarrow \tilde{z}_s$.

Now we examine the posterior density of $PD\zc|\X,\Y$. Appealing to the same line of reasoning as \eqref{posterior_approx}, we have that:
\begin{equation}
\begin{aligned}
    {p}^\star(z_c|x,y) &= \int p^\star(z_c,z_s|x,y)\partial{z_s} {\stackrel{(a)}=}\frac{\int {p}^\star(x|z_c,z_s){p}^\star(z_c|y)p^\star(z_s)\partial{z_s}}{p^\star(x|y)} \\&{\stackrel{(b)}=} \frac{\int p^\star(x|D^{-1}P^{-1}z_c;z_s){p}^\star(z_c|y){p}^\star(z_s)\partial{z_s}}{p^\star(x|y)}\\
     &{\stackrel{(c)}=} \frac{\int {p}_f\left(x|z_c;N{z}_s\right){p}^\star(z_c|y)p^\star({z}_s)\partial{{z}_s}}{p^\star(x|y)}{\stackrel{(d)}=} \frac{\int {p}_f\left(x|z_c;N{z}_s\right){p}(\tilde{z}_c|y)p^\star({z}_s)\partial{{z}_s}}{p^\star(x|y)},
    \end{aligned}
    \label{posterior_true}
\end{equation}
where ${p}^\star(x|z_c,z_s)$ is the density of the distribution $\X|PD\zc,\zs$ and ${p}^\star(z_c|y)$ is the density of the distribution $PD\zc|\Y$. Here, the equality ${\stackrel{(a)}=}$ follows from $PD\zc \independent \zs$, ${\stackrel{(b)}=}$ follows from the density $\X|(PD\zc,\zs) = p^\star(\X|(D^{-1}P^{-1}\zc;\zs))$, and ${\stackrel{(c)}=}$ follows from the relationship between ${f}$ and $f^\star$. Finally,
${\stackrel{(d)}=}$ follows from the equality $p^\star(z_c|y) = {p}(\tilde{z}_c|y)$ since $PD\zc \stackrel{\text{dist}}{=} \tzc$ (due to the  relation between ${\tilde{\Z},\Z}$ and that they are both Gaussian random variables after conditioning on a label). Comparing \eqref{posterior_approx} and \eqref{posterior_true}, we have the desired result.

\section{Comparisons with \cite{VAE-ICA}}
\label{sec:comparison_with_khemakhem}
Previously, \cite{VAE-ICA} proved that supervision enables identifiability of the latent features. The only similarity of our results with this work can be found in Lemma~\ref{lemma:cl}, although our guarantees distinguish core and style features and allow for more relaxed assumptions. In particular, while the target label can take two distinct values for Assumption 1.2 to be satisfied, the assumption in \cite{VAE-ICA} -- when specialized to the Gaussian prior --  requires that the target label takes at least $2k$ distinct values where $k$ is the number of the latent features. On all the remaining theoretical and methodological aspects, our setting differs substantially from \cite{VAE-ICA}. Importantly, while our method can be employed for interpretable predictions, their proposed I-VAE is simply not applicable in our setting, as it requires labels as inputs and thus cannot perform prediction. Furthermore, our methodology and theoretical guarantees (see Theorem~\ref{thm:PDVAE}) covers the case of overparameterized latent spaces, i.e. the case in which the model allows for mode latent features than the ground truth ones. This is another practically relevant novelty, as in general the number of ground truth features is unknown.

\section{Implementation details}
\label{sec:impdetails}

\subsection{Datasets and pre-processing}
\label{sec:datadetails}
For all the datasets, pixels are transformed to have values between 0 and 1 and the image size of 64x64 is kept. We randomly fix train and test set with sizes respectively of $60\%$ and $40\%$ for MPI3D, $90\%$ and $10\%$ for shapes3D, SmallNORB and Plantvillage datasets. For the \chest{} dataset, we use the pre-defined train test splits.

\subsection{Hyperparameter selection}
Hyperparameter selection has been performed via visual inspection of the traversals \emph{on the training set}. 
This is a correct validation of the algorithm, since no test data has been utilized for model selection. 
Furthermore, it reflects the procedure that we propose in the paper, where model selection is carried out with a human expert. 
All the traversals and results we report are then obtained on the test set as usual.

\subsection{Hyper-parameters and training configuration}
\label{sec: hyperparams and training}
Most hyper-parameters and training configurations are kept fixed across datasets. An overview of hyper-parameter settings is shown in Table \ref{table:hyperparameters}. We set a fixed dimension of $(\zc, \zs) = (10, 20)$. Notice that, similar to the real-life setting where the exact dimension of the true underlying latent features is unknown we allow the latent dimension of core and style latent features to be higher than theoretically needed for all the synthetic datasets. \\
We set the following values of (prediction term weight, group sparsity regularization) for the experiments: (50, 0.05) for MPI3D; (10, 0.0001) for Shapes3D; (50, 0.01) for SmallNORB; (200, 0.01) for PlantVillage; (200, 0.05) for ChestXRay.

\begin{table}
\caption{Hyperparameter settings for all datasets.}
\label{table:hyperparameters}
\centering
 \begin{tabular}{ll} 
     \toprule
     \textbf{Hyperparameter} & \textbf{Value} \\ 
     \midrule 
     Core latent space dimension & 10  \\
     Style latent space dimension & 20  \\
     Batch size & 132  \\ 
     Optimizer & ADAM \\
     Learning rate & 5e-4  \\
     Decoder type & Bernoulli \\
     Prediction loss & binary cross entropy \\
     Training steps & $150,000$ \\
     \bottomrule
\end{tabular}

\end{table}

\subsection{\methName{} model architecture}
\label{sec: model architecture}
The architecture of {\methName} consists of five main modules that share weights where appropriate. The five modules are: the encoder, decoder and predictor in the prediction VAE, and the encoder and decoder in the concept-learning VAE.

{\bf Predictor} The predictor in the prediction VAE is a simple linear predictor mapping from core latent features to the labels where each label has a separate linear predictor.

{\bf Decoders} The decoders in both the prediction VAE and the concept-learning VAE share weights with an architecture as depicted on the right in Table \ref{table:encoderdecoder}.

{\bf Encoders} The encoder in the prediction VAE consists of a backbone with architecture as depicted on the left in Table \ref{table:encoderdecoder} and two parallel fully connected layers. The output of the backbone is fed into the two separate fully connected layers where one is learning the posterior distribution of core latent features and one is learning the posterior distribution of style latent features.

The encoder in the concept-learning VAE shares all weights that are associated with style latent features with the encoder of the prediction VAE. In particular, the weights of the backbone and fully connected layer that output the posterior distribution of style latent features are shared across both. 
No weights are shared for that part of the concept-learning VAE associated with core latent features. Thus, a separate backbone and fully connected layer are used to learn the posterior distribution of core latent features in the concept-learning VAE. The label $y$ is incorporated in the concept-learning VAE by feeding it jointly with the output of the backbone to the fully connected layer that outputs the posterior distribution of the core latent features.

\begin{table}[!h]
\caption{Encoder backbone and decoder architecture. \textbf{Abbreviations:} $c$ denotes the channel size of the input image, $\tilde{k}_c$ the dimension of core latent factors, and $\tilde{k}_s$ the dimension of style latent factors. \textbf{Layer parameters:} For fully connected layers the first parameter denotes input dimension, and the second parameter output dimension. For LeakyReLU the parameter denotes its negative slope. For dropout layers the parameter denotes the probability that a whole channel is dropped out (2D dropout). For convolutional and transposed convolutional layers the parameters can be interpreted as follows: output channel size, kernel size, stride, padding.}
\label{table:encoderdecoder}
\centering
 \begin{tabular}[t]{cccc}
 \toprule
 \multicolumn{2}{c}{\textbf{Encoder backbone}} & \multicolumn{2}{c}{\textbf{Decoder}} \\
 \cmidrule(r){1-2} \cmidrule(r){3-4}
 Input size: & $64$ x $64$ x $c$  & Input size: & $30 = \tilde{k}_c + \tilde{k}_s$ \\
 Output size: & 256 & Output size: & $64$ x $64$ x $c$  \\
 \cmidrule(r){1-2} \cmidrule(r){3-4}
 \textbf{Layer} & \textbf{Parameters} & \textbf{Layer} & \textbf{Parameters} \\
 \cmidrule(r){1-2} \cmidrule(r){3-4}
 Conv & 64, 3, 2, 1  & FC & 30, 512   \\
 LeakyReLU & 0.01 &  ReLU  \\
 Dropout & 0.1 & FC & 512, 1024  \\
 Conv & 64, 3, 2, 1  & ConvTranspose & 64, 3, 2, 0\\
 LeakyReLU & 0.01 & ReLU \\
 Dropout & 0.1 & ConvTranspose & 64, 3, 2, 1 \\
 Conv & 64, 3, 2, 1  & ReLU \\
 LeakyReLU & 0.01 & ConvTranspose & 64, 3, 2, 1 \\
 Dropout & 0.1 & ReLU \\
 Conv & 64, 3, 2, 1  & ConvTranspose & $c$, 4, 2, 2 \\
 LeakyReLU & 0.01\\
 Dropout & 0.1\\
 Flatten & \\
 FC & 1024, 256  \\
 \bottomrule 
\end{tabular}
\end{table}

\subsection{SENN and CCVAE model architecture}
For SENN, we employ the same architecture as for \methName{}. In particular, we utilize the encoder backbone presented in Table~\ref{table:encoderdecoder}, and map the encoding obtained from the backbone to the core and style features via two parallel linear layers. The decoder utilized is the same as in Table. Furthermore, SENN employs an additional mapping from the input $\X$ to the prediction weights utilized on top of the core features. This mapping is given by a neural network with structure \texttt{Conv(32, 4, 2, 0), MaxPool(2, 2), ReLU, Conv(32, 4, 2, 0), MaxPool(2, 2), ReLU}. To map to the prediction weights, we flatten and then utilize two linear layers with output dimension $288$ and $128$ and activations ReLU and Tanh respectively.

For CCVAE, we employ the publicly available architecture from the authors at \\ \href{https://github.com/thwjoy/ccvae}{https://github.com/thwjoy/ccvae}. We note that the encoder-decoder pair is roughly equivalent to that utilized for \methName{}.

\newpage
\section{Details on synthetic datasets}
\label{sec: details synthetic}

\subsection{MPI3D traversals and details}
\label{sec:mpitraversals}
\begin{figure}[h!]
    \centering
    \begin{subfigure}[b]{0.3\linewidth}
    \includegraphics[scale = 0.4]{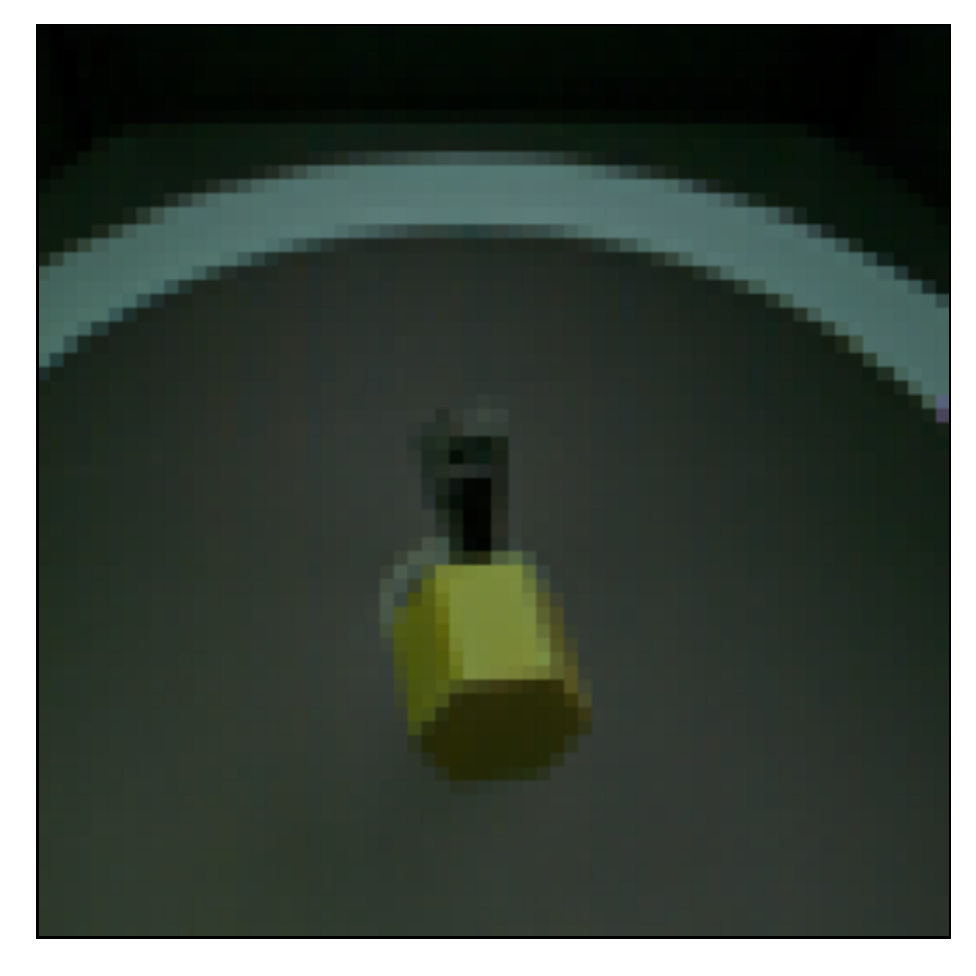}
    \end{subfigure}
    \begin{subfigure}[b]{0.3\linewidth}
    \includegraphics[scale = 0.4]{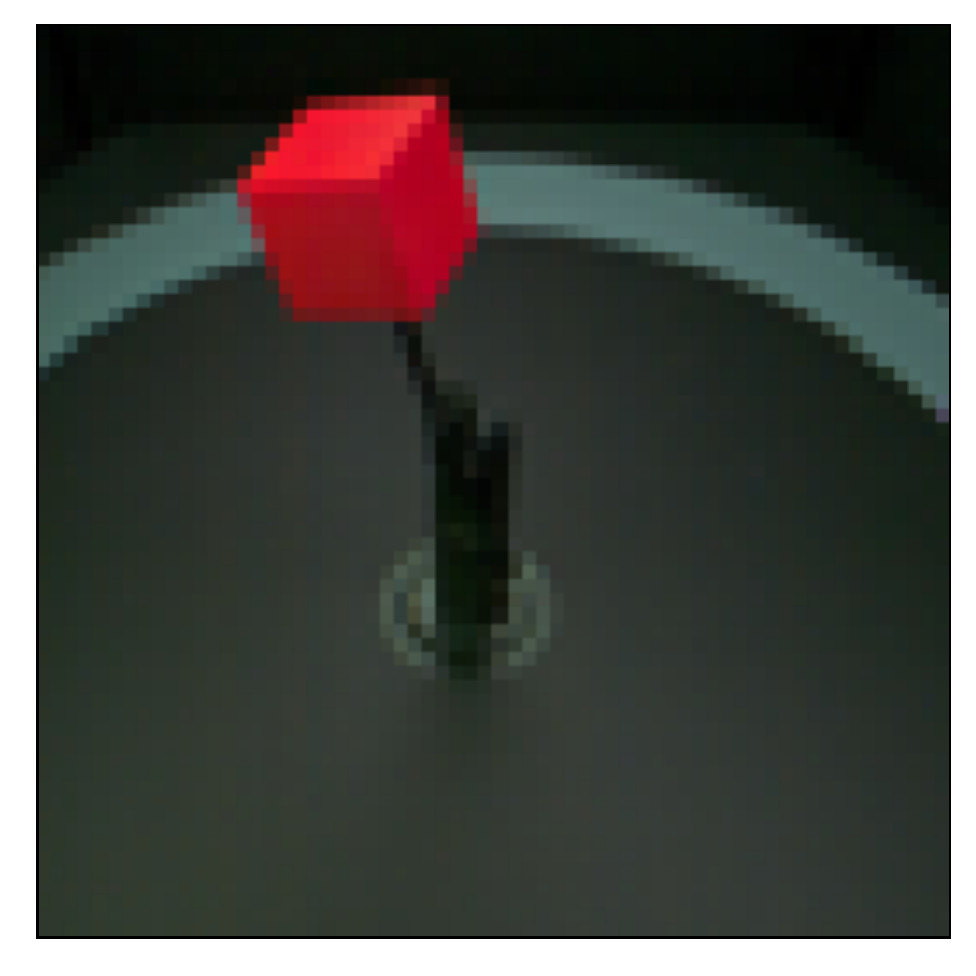}
        \end{subfigure}
    \begin{subfigure}[b]{0.3\linewidth}
    \includegraphics[scale = 0.4]{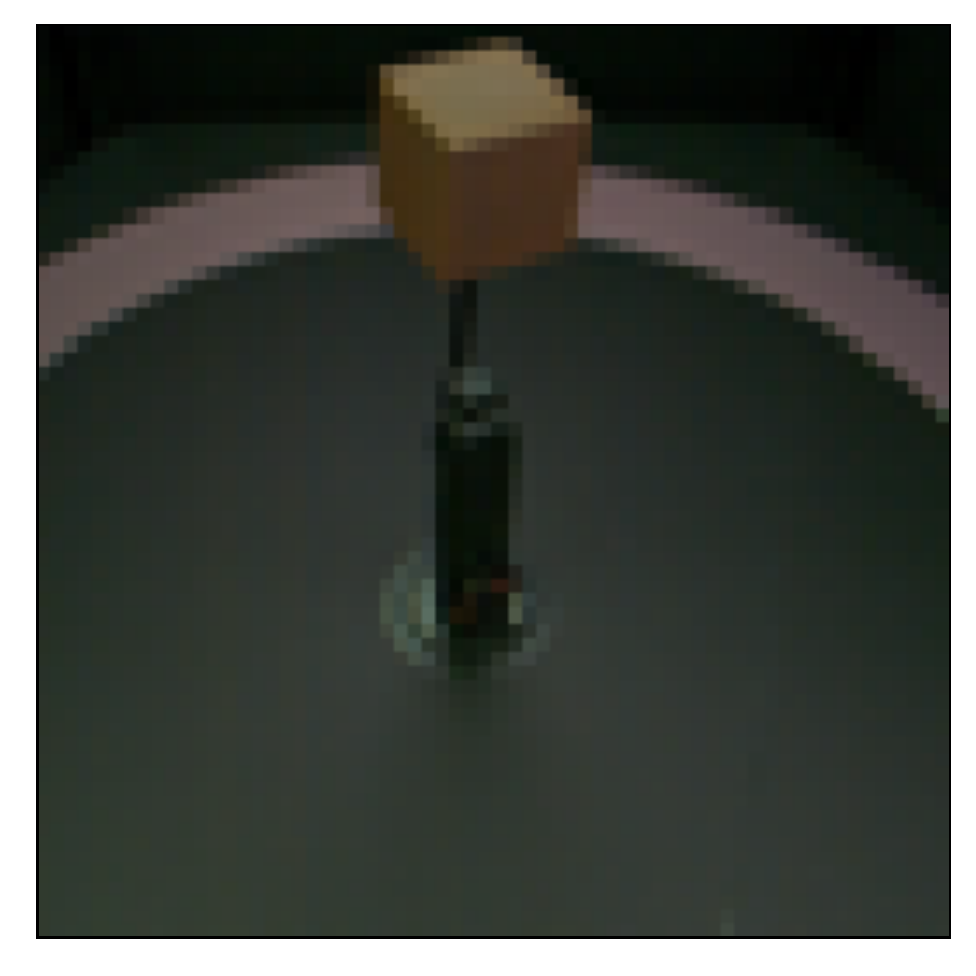}
        \end{subfigure}
    \caption{Some example images from the MPI3D dataset.}
    \label{fig:mpi_examples}
\end{figure}

\begin{figure}[h!]
\centering
\begin{subfigure}[b]{0.325\linewidth}
\centering
    \includegraphics[width=\textwidth]{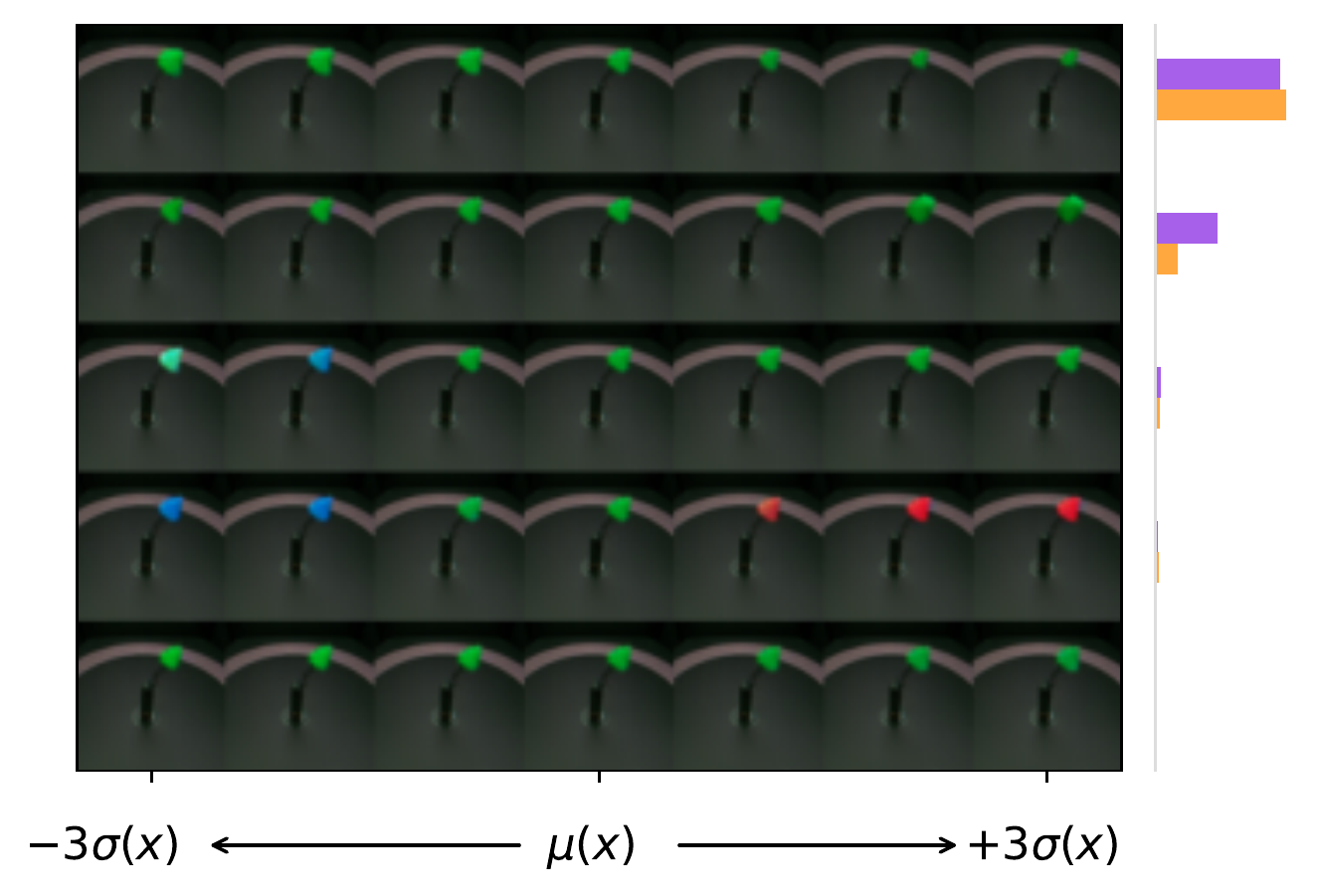}
        \end{subfigure}
\begin{subfigure}[b]{0.325\linewidth}
\centering
    \includegraphics[width=\textwidth]{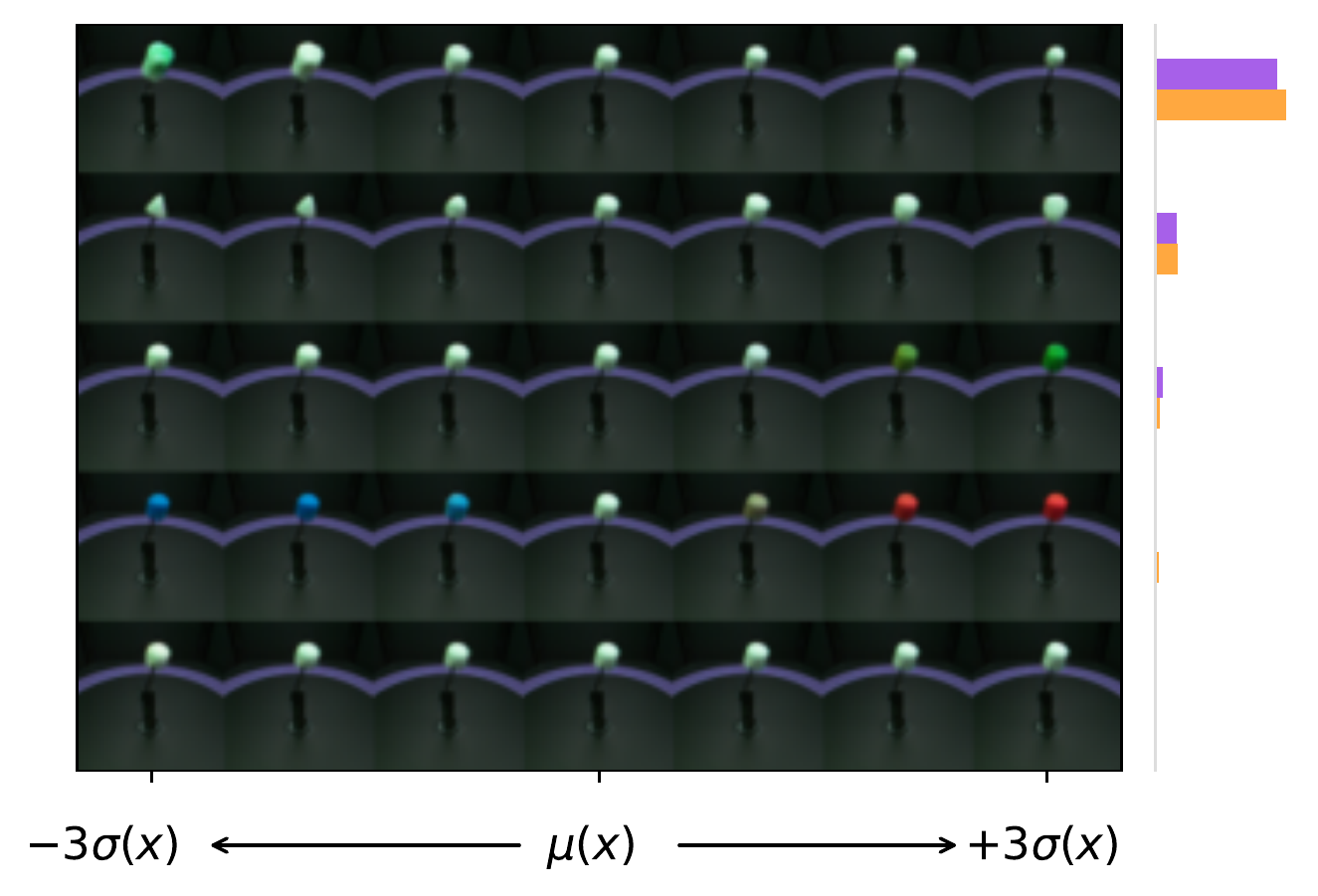}
        \end{subfigure}
\begin{subfigure}[b]{0.325\linewidth}
\centering
    \includegraphics[width=\textwidth]{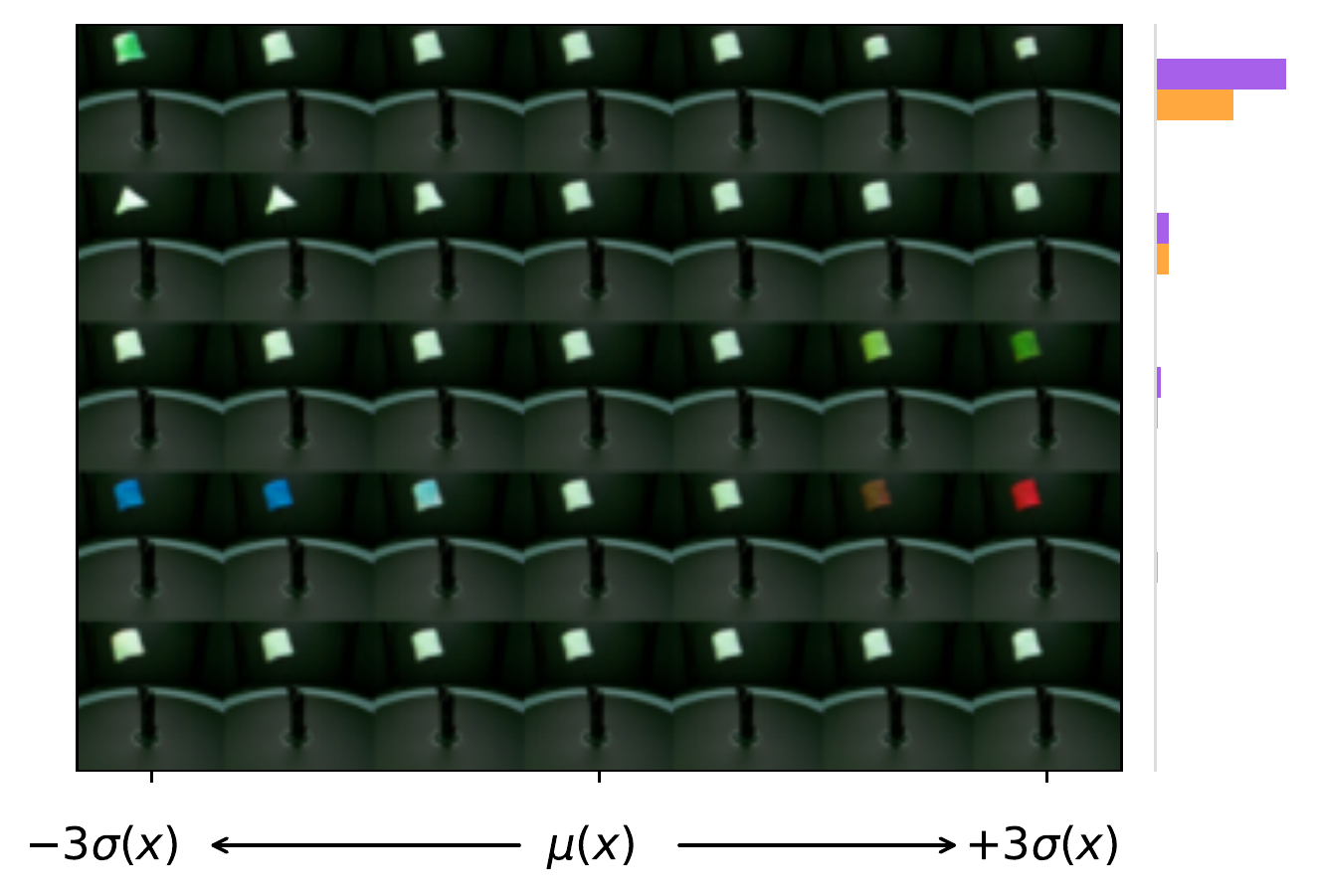}
        \end{subfigure}
\begin{subfigure}[b]{0.325\linewidth}
\centering
    \includegraphics[width=\textwidth]{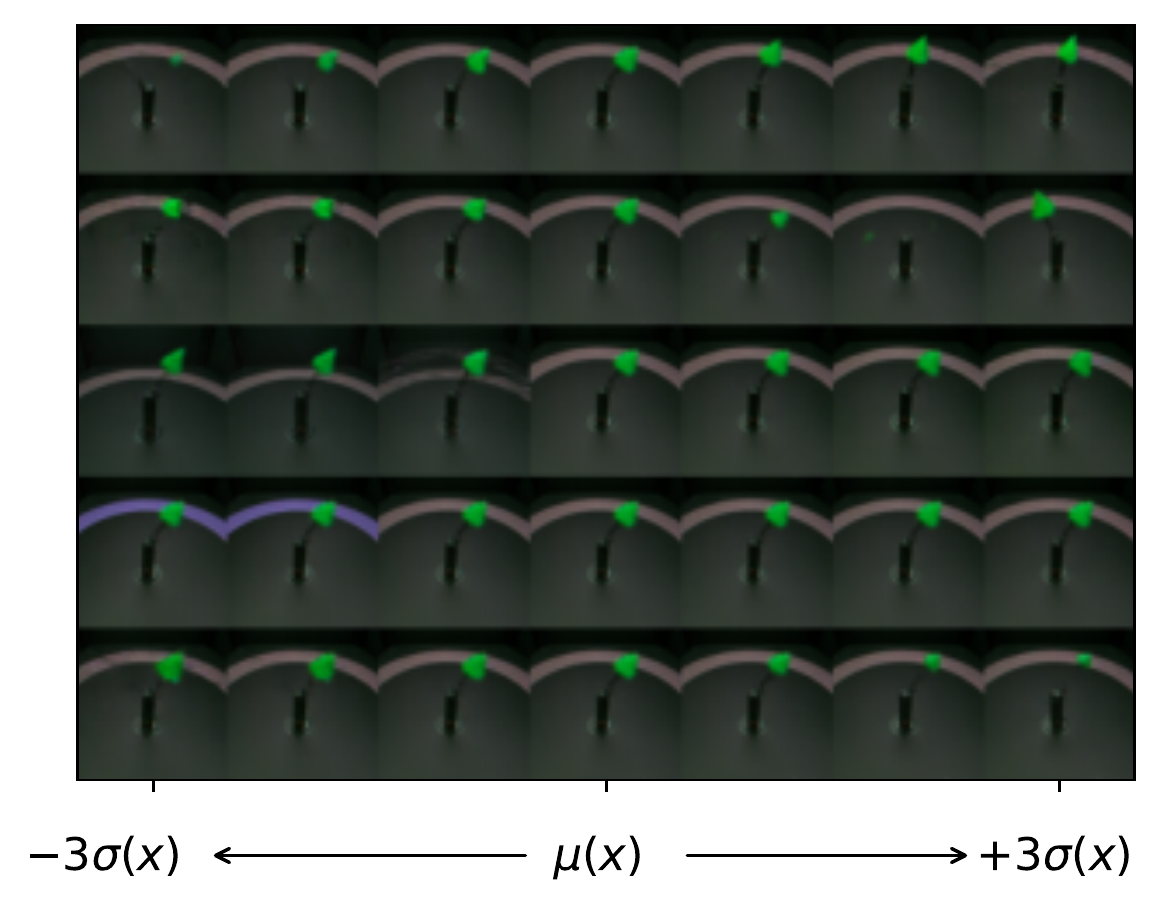}
        \end{subfigure}
\begin{subfigure}[b]{0.325\linewidth}
\centering
    \includegraphics[width=\textwidth]{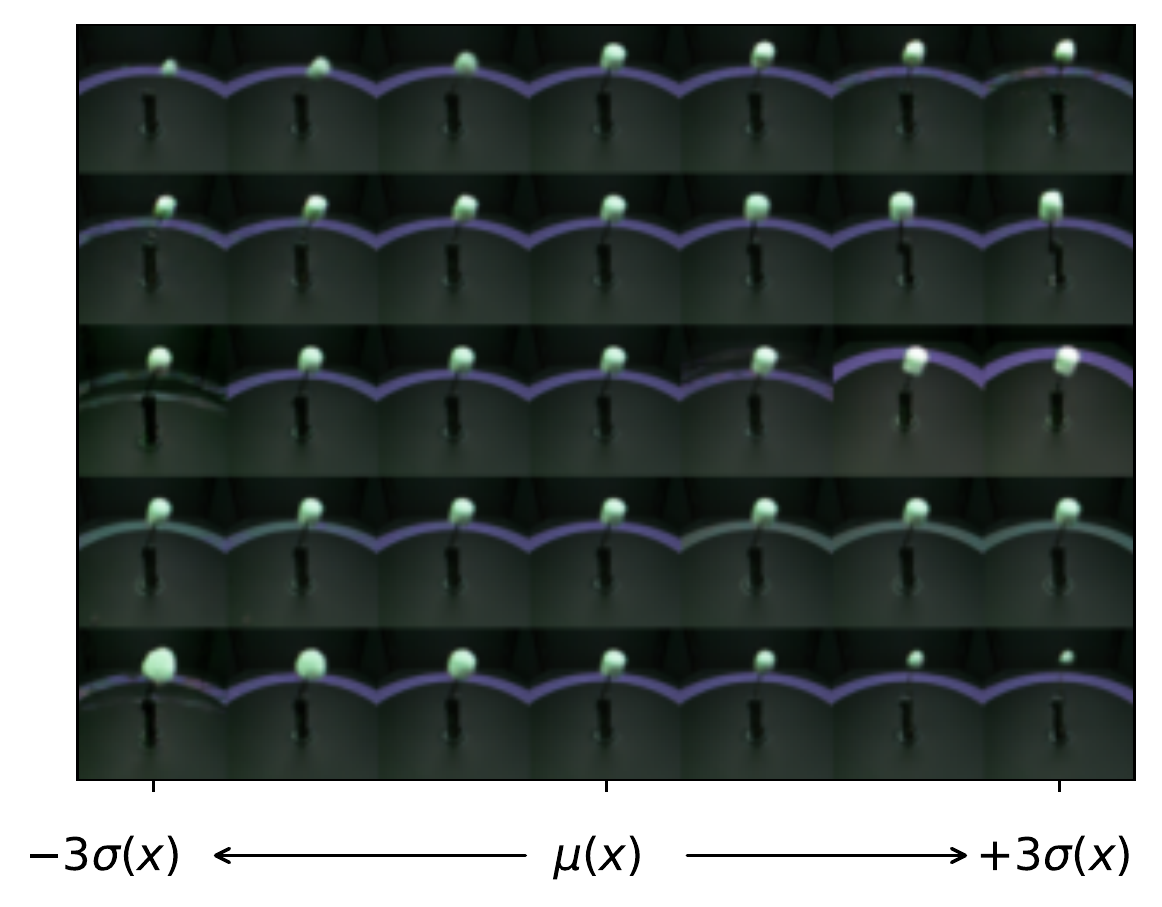}
        \end{subfigure}
\begin{subfigure}[b]{0.325\linewidth}
\centering
    \includegraphics[width=\textwidth]{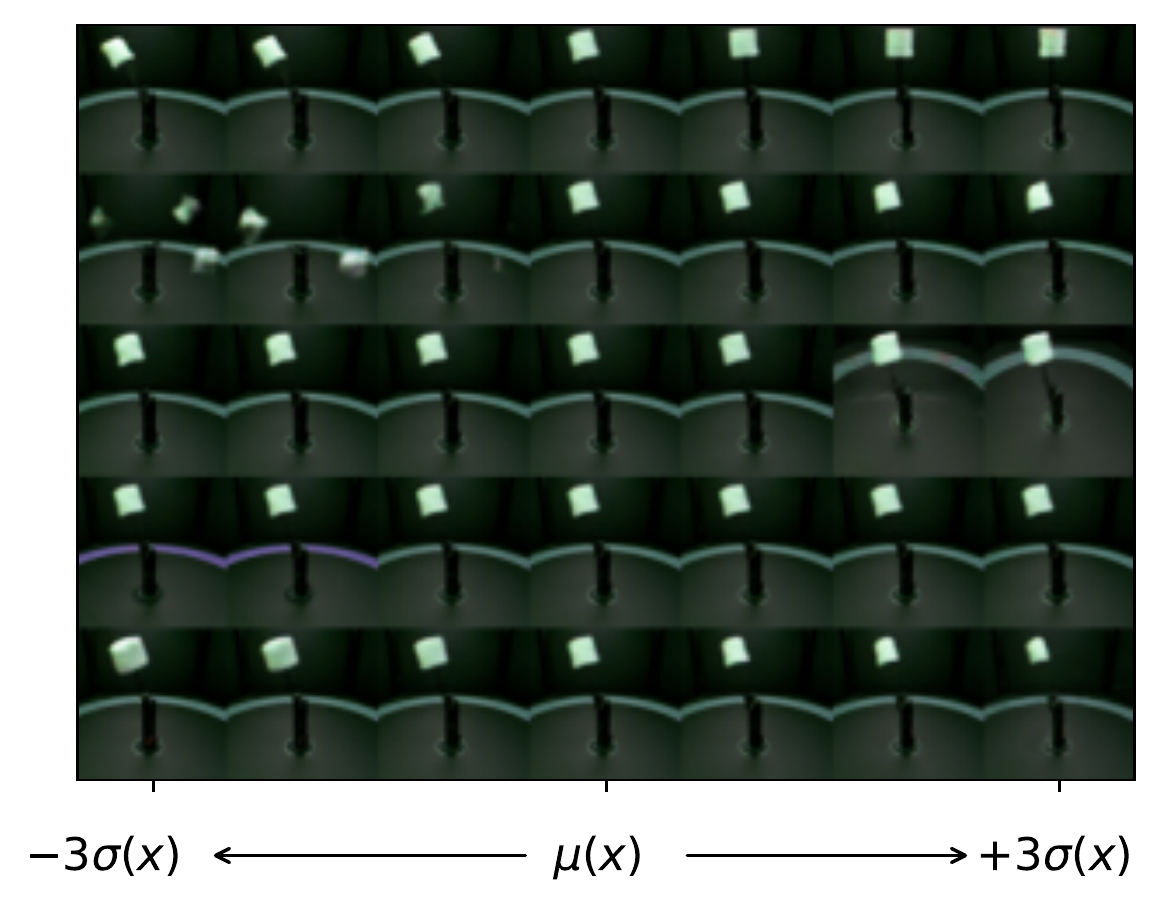}
        \end{subfigure}
\caption{\methName{} traversals of $\zc$ (first row) and $\zs$ on MPI3D dataset.}
\label{fig:traversals_mpi_clap}
\end{figure}

The MPI3D dataset is an artificial dataset of images where the ground truth factors of variation are object color, shape and size, as well height of the camera, background color, horizontal and vertical axes of the camera. In Figure~\ref{fig:mpi_examples}, we present some example images from the dataset. \\
We create synthetic labels according to the following rules: for the first label, $y = 1$ if color in \{white, green, brown, olive\} and shape in \{cone, cube, cylinder, sphere\} and size in \{small\}, and $y = 0$ otherwise; for the second label, $y = 1$ if color in \{green, red, blue\} and size in \{large\}, and $y = 0$ otherwise; for the third label, $y = 1$ if shape in \{cone, pyramid\}, and $y = 0$ otherwise; for the final label, $y = 1$ if shape in \{cylinder, hexagonal, pyramid\}, and $y = 0$ otherwise. \\
We present additional MPI3D traversals of \methName{} for $\zc$ (first row) and $\zs$ (second row) in Figure~\ref{fig:traversals_mpi_clap}.

\newpage
\subsection{Shapes3D}
\label{sec:shapes3dtraversals}
\begin{figure}[h!]
    \centering
    \begin{subfigure}[b]{0.325\linewidth}
    \includegraphics[scale = 0.4]{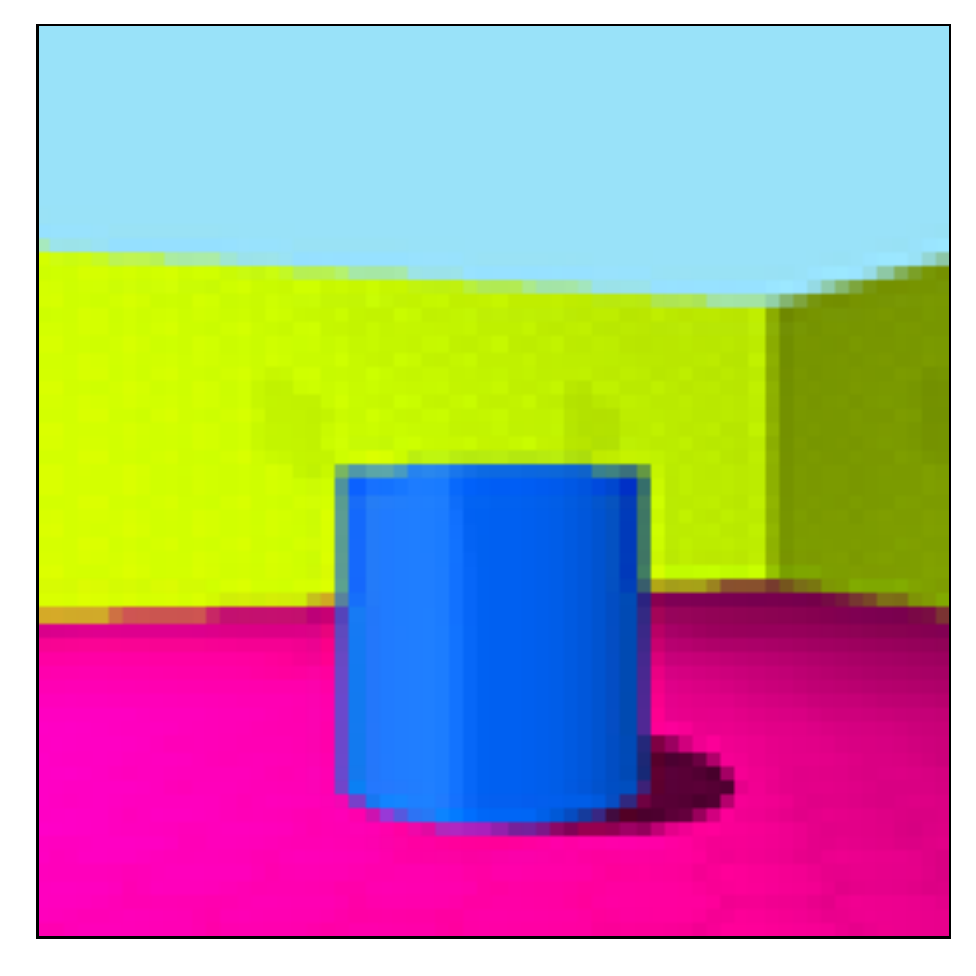}
        \end{subfigure}
    \begin{subfigure}[b]{0.325\linewidth}
        \centering
    \includegraphics[scale = 0.4]{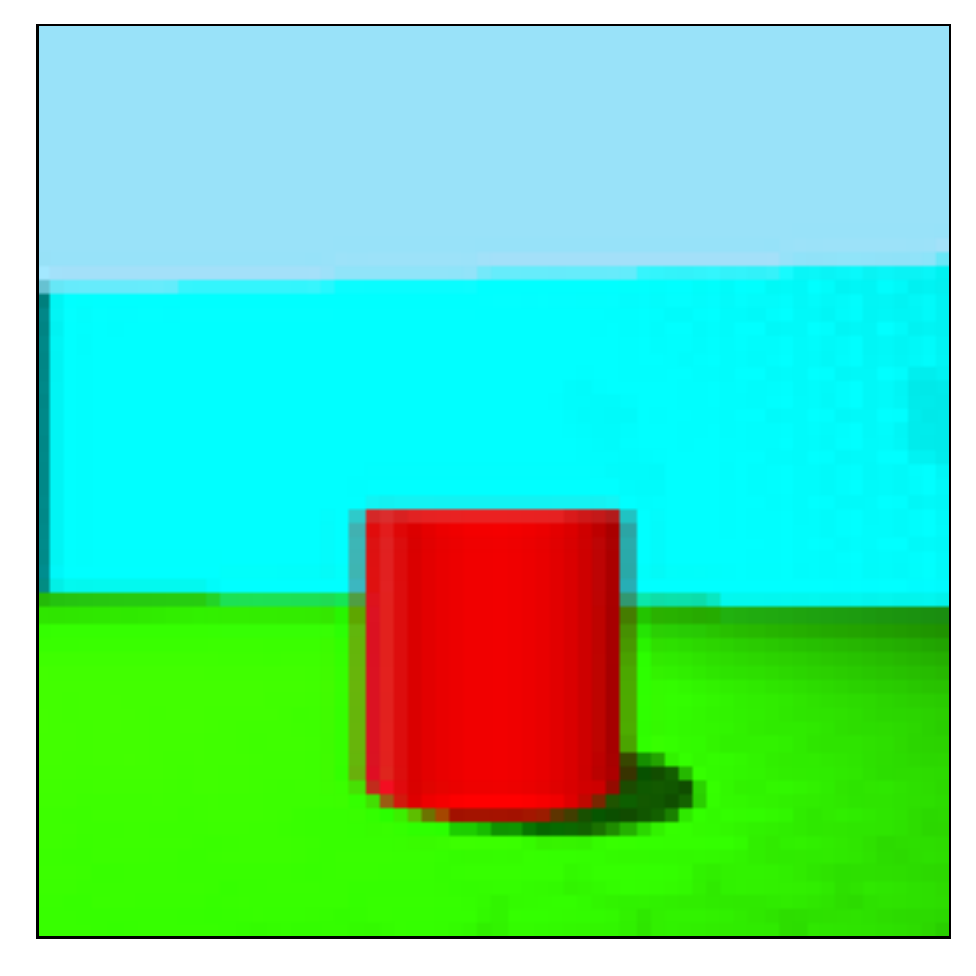}
        \end{subfigure}
    \begin{subfigure}[b]{0.325\linewidth}
        \centering
    \includegraphics[scale = 0.4]{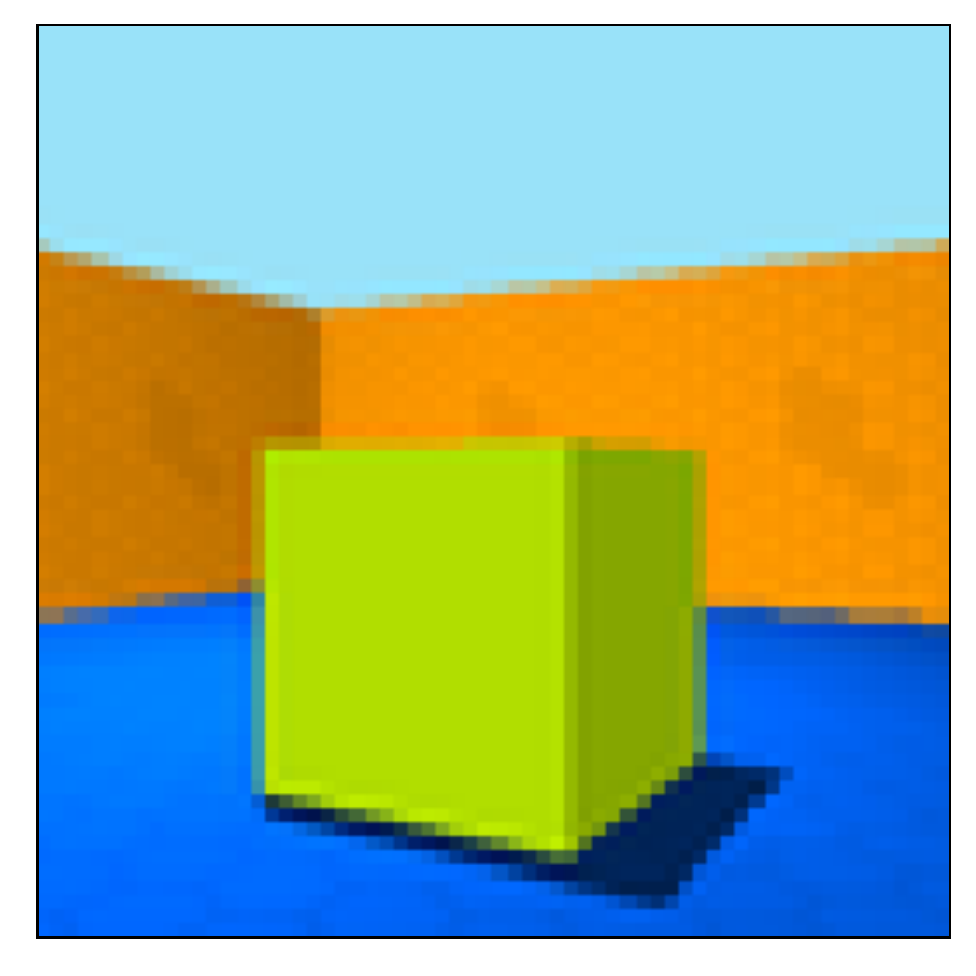}
        \end{subfigure}
    \caption{Some example images from the Shapes3D dataset.}
    \label{fig:shapes3d_examples}
\end{figure}

\begin{figure}[h!]
\centering
\begin{subfigure}[b]{0.325\linewidth}
    \includegraphics[width=\linewidth]{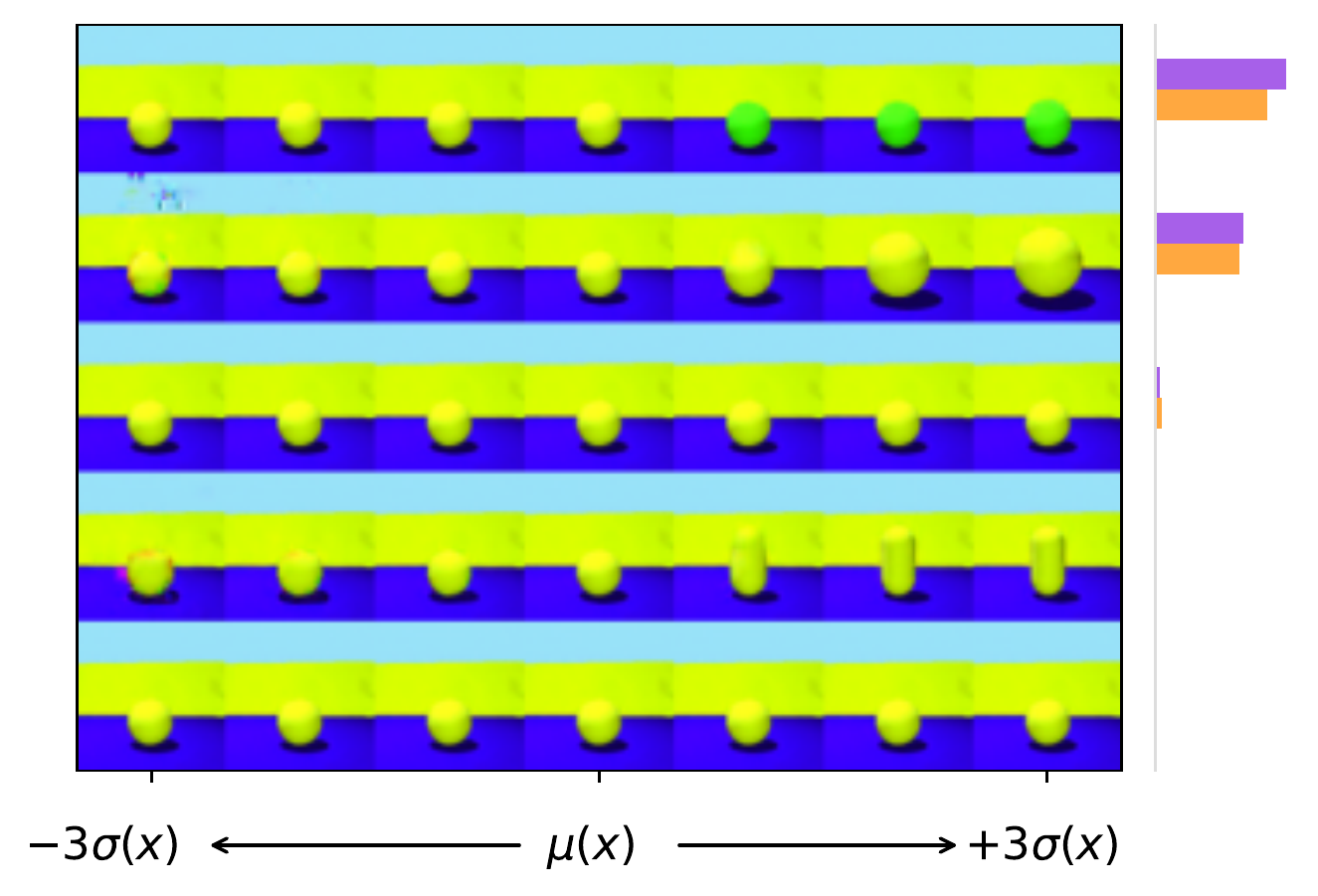}
        \end{subfigure}
\begin{subfigure}[b]{0.325\linewidth}
\centering
    \includegraphics[width=\linewidth]{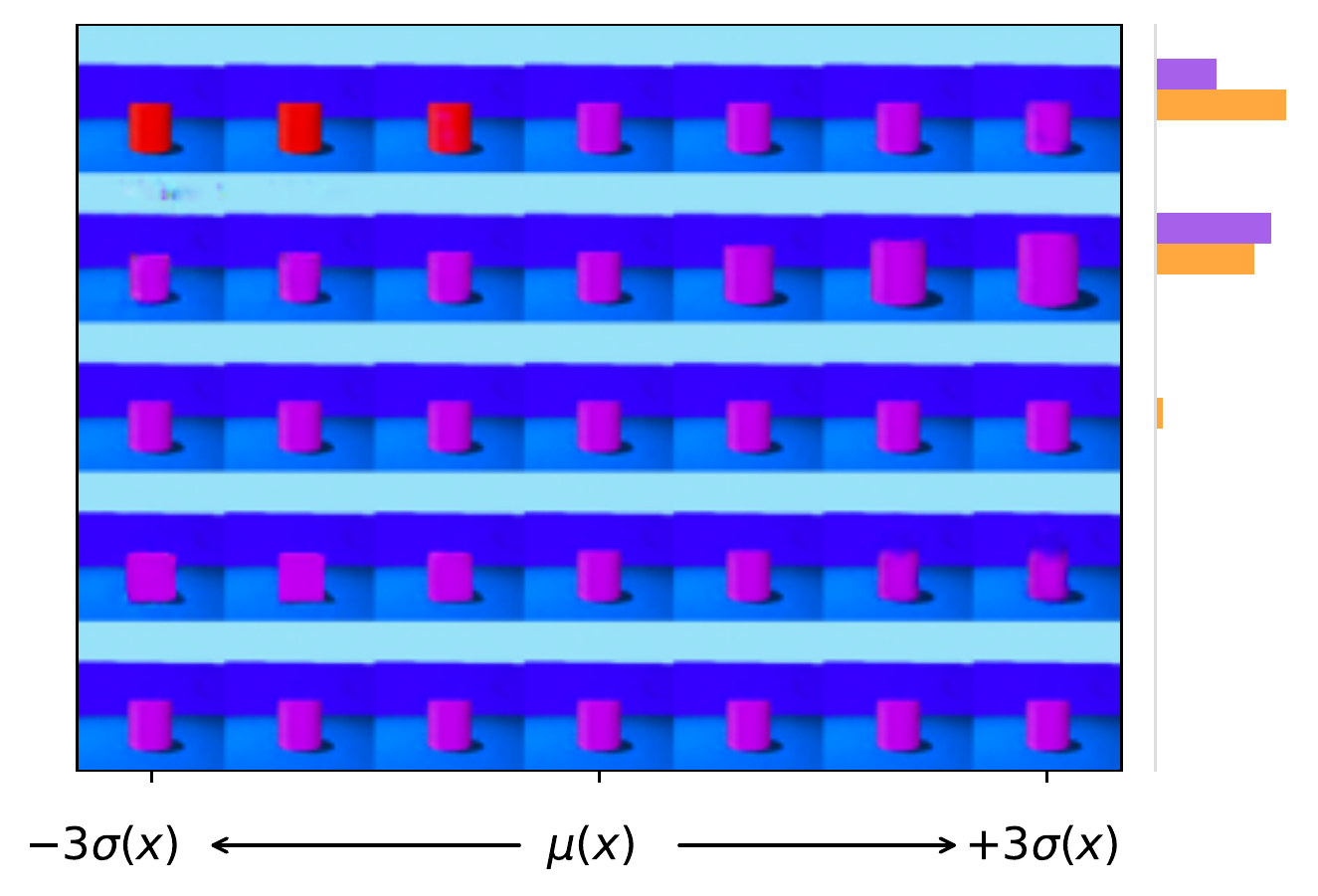}
        \end{subfigure}
\begin{subfigure}[b]{0.325\linewidth}
\centering
    \includegraphics[width=\linewidth]{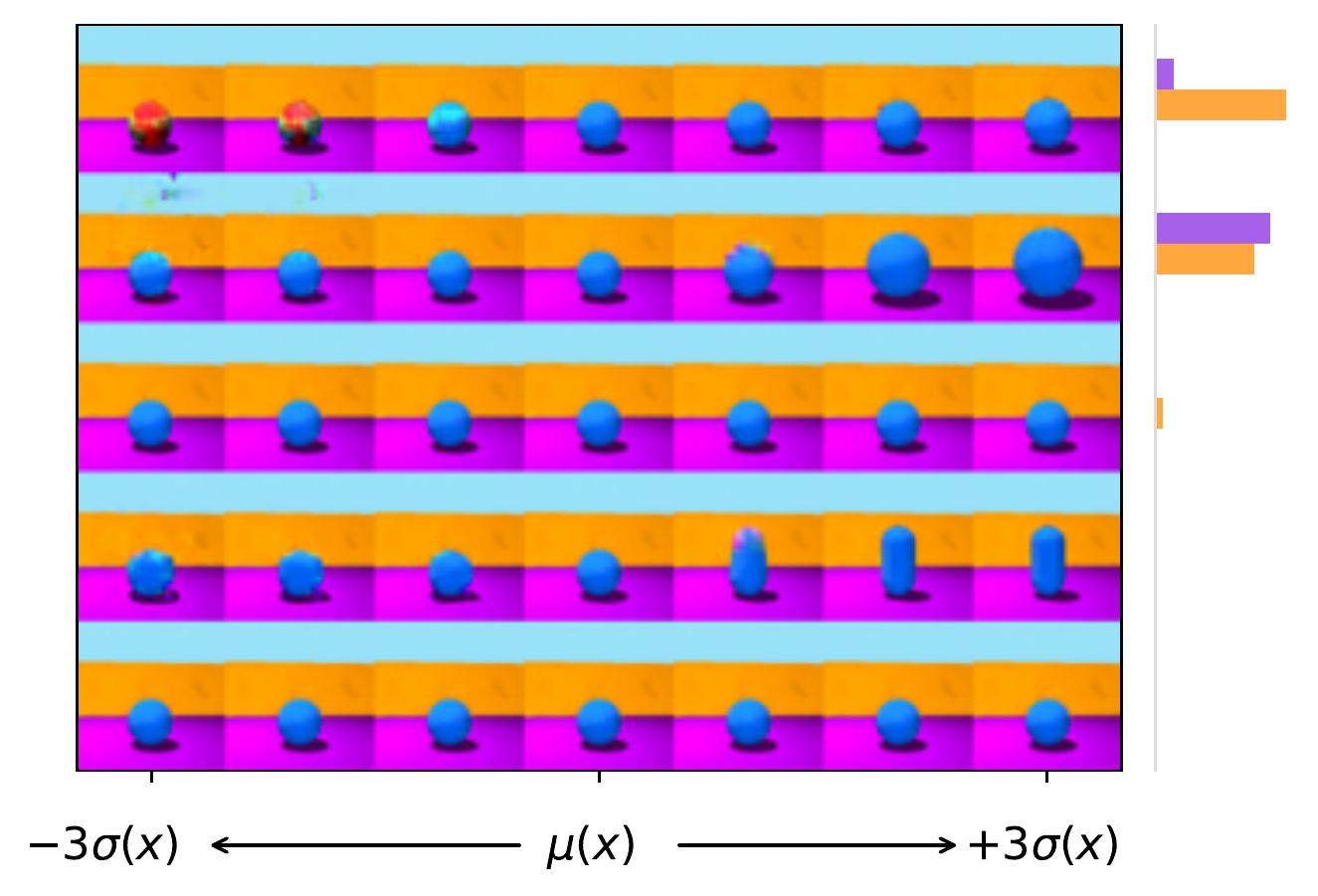}
        \end{subfigure}
\begin{subfigure}[b]{0.325\linewidth}
\centering
    \includegraphics[width=\linewidth]{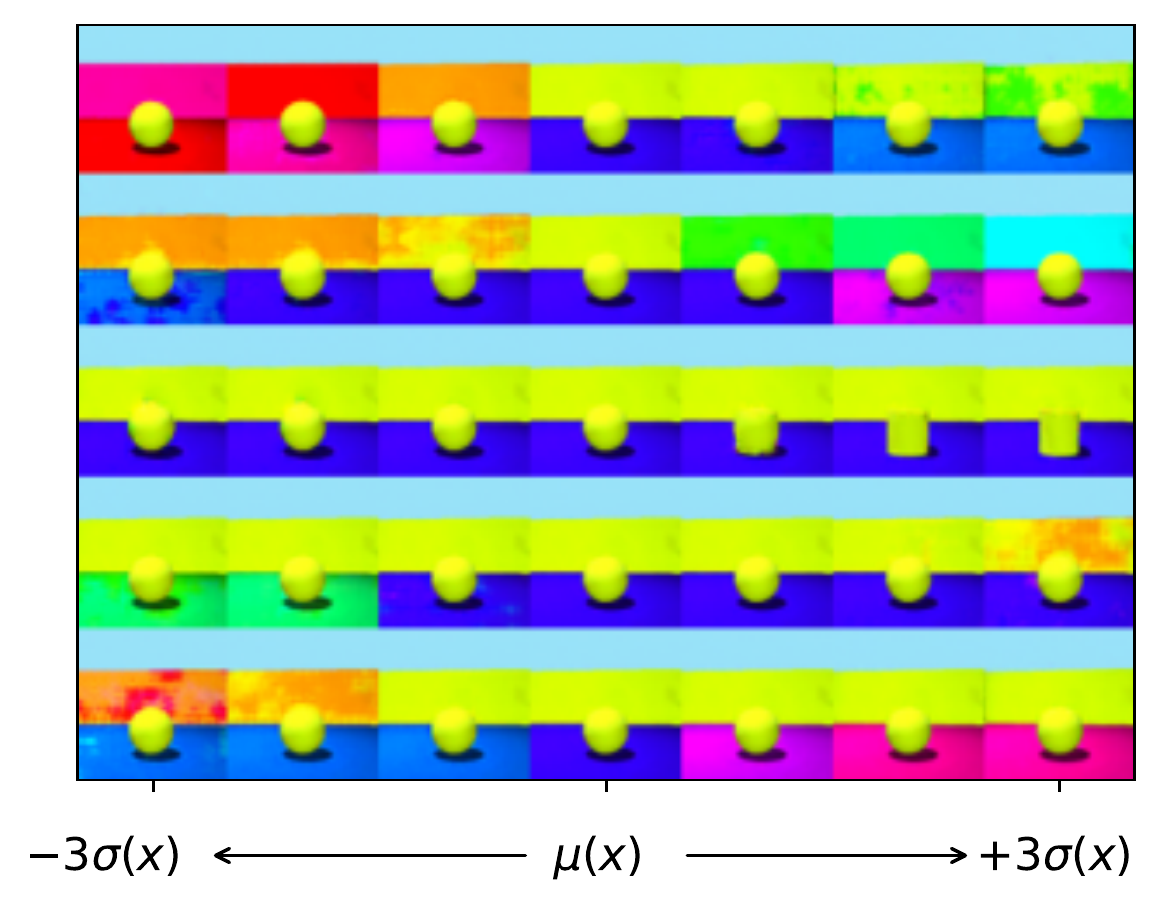}
        \end{subfigure}
\begin{subfigure}[b]{0.325\linewidth}
\centering
    \includegraphics[width=\linewidth]{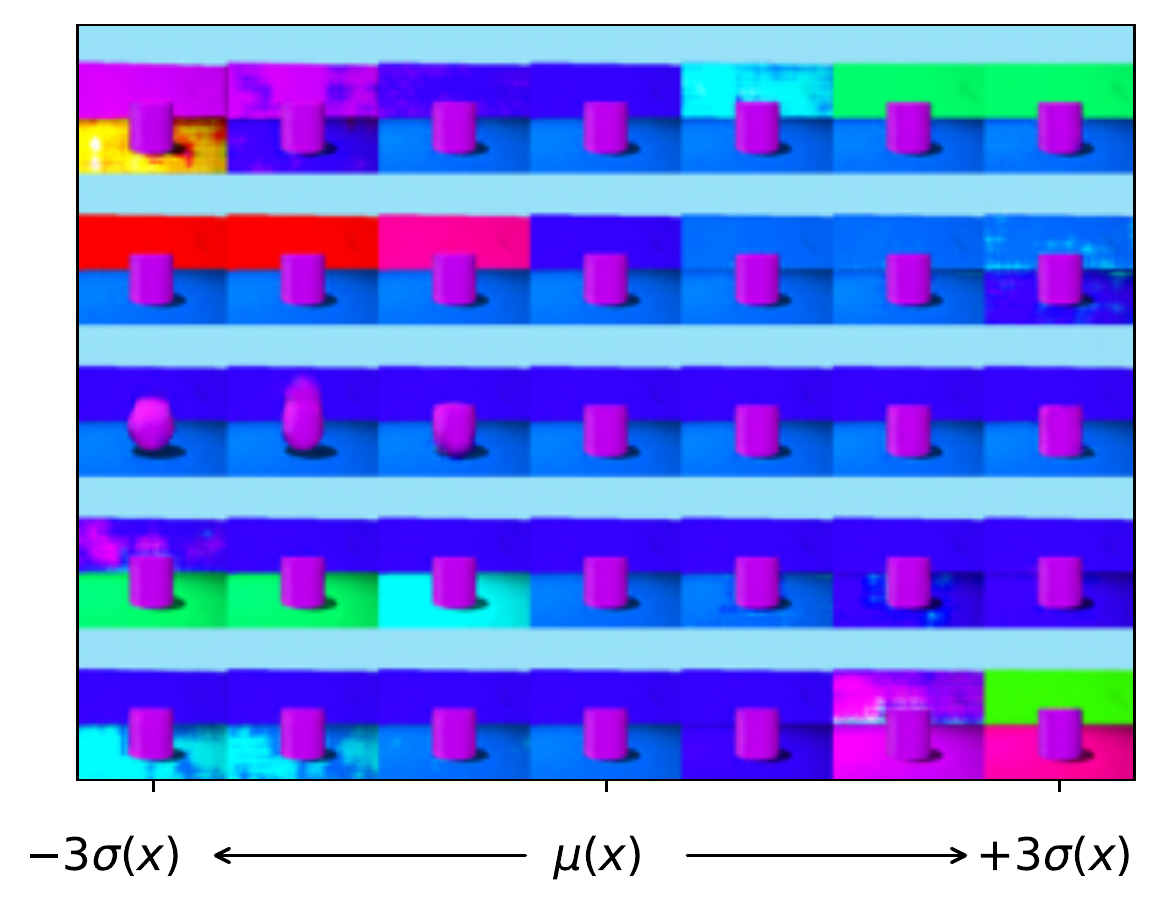}
        \end{subfigure}
\begin{subfigure}[b]{0.325\linewidth}
\centering
    \includegraphics[width=\linewidth]{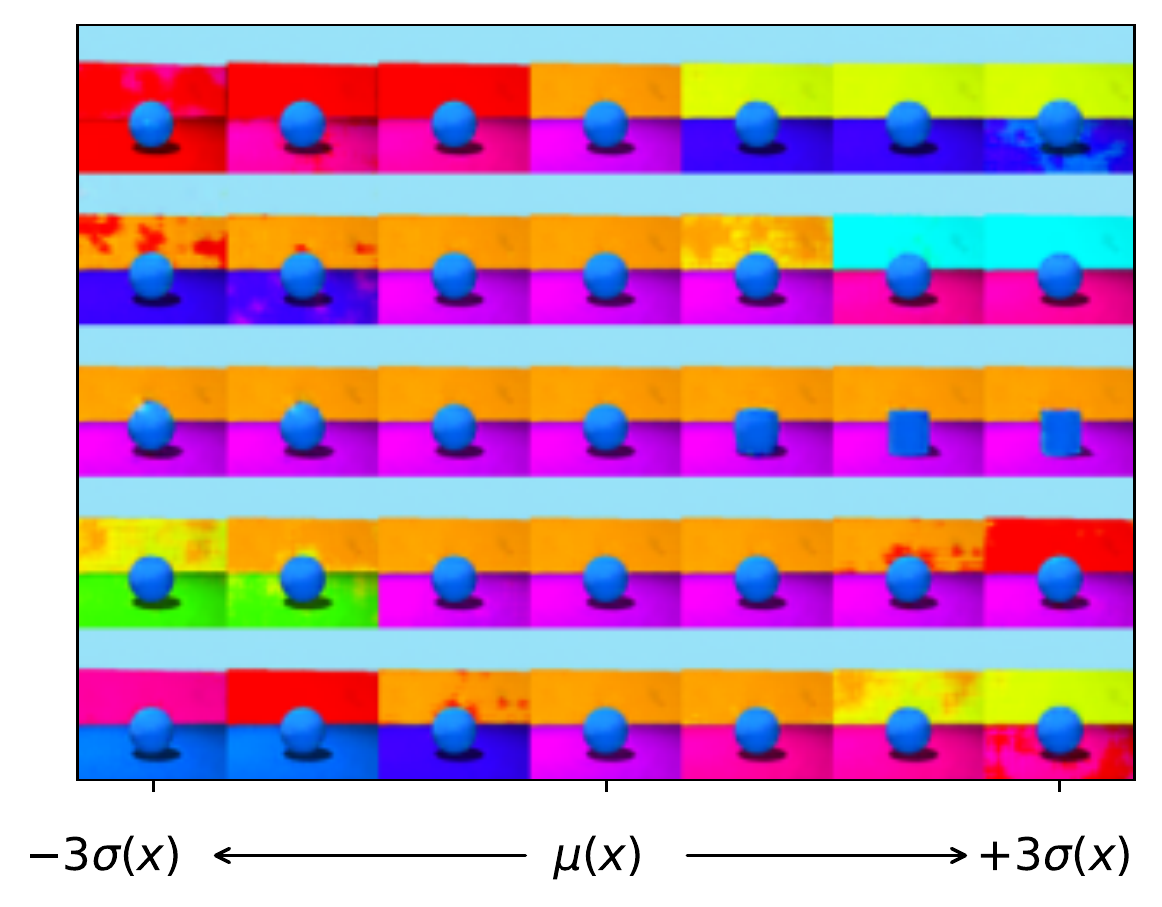}
        \end{subfigure}
\caption{\methName{} traversals of $\zc$ (first row) and $\zs$ on the Shapes3D dataset.}
\label{fig:traversals_shapes3d_clap}
\end{figure}

The Shapes3D dataset is composed of synthetic images of different 3D objects. The ground truth factors of variations are floor hue, background wall hue, orientation, and the object's hue, scale, and shape. In Figure~\ref{fig:shapes3d_examples}, we present some example images from the dataset.\\
We rescale all the factors of variation, which are already discrete, to take integer values starting from 0. In particular, hue and scale of the object have values in $[0, 9]$ and are used to create the synthetic labels. We create synthetic labels according to the following rules: for the first label, $y = 1$ if scale $\le 5$ and hue $\ge 3$, $y = 0$ otherwise; for the second label, $y = 1$ if scale $\ge 3$ and hue $\ge 3$, $y = 0$ otherwise; for the second label, $y = 1$ if scale $\le 4$ and hue $\ge 2$, $y = 0$ otherwise; for the final label, $y = 1$ if scale $\ge 5$, $y = 0$ otherwise. \\
We present \methName{} traversals on the Shapes3D dataset for $\zc$ (first row) and $\zs$ (second row) in Figure~\ref{fig:traversals_shapes3d_clap}.

\newpage
\subsection{SmallNORB}
\label{sec:smallnorbtraversals}
\begin{figure}[h!]
    \centering
    \begin{subfigure}[b]{0.3\linewidth}
    \includegraphics[scale = 0.4]{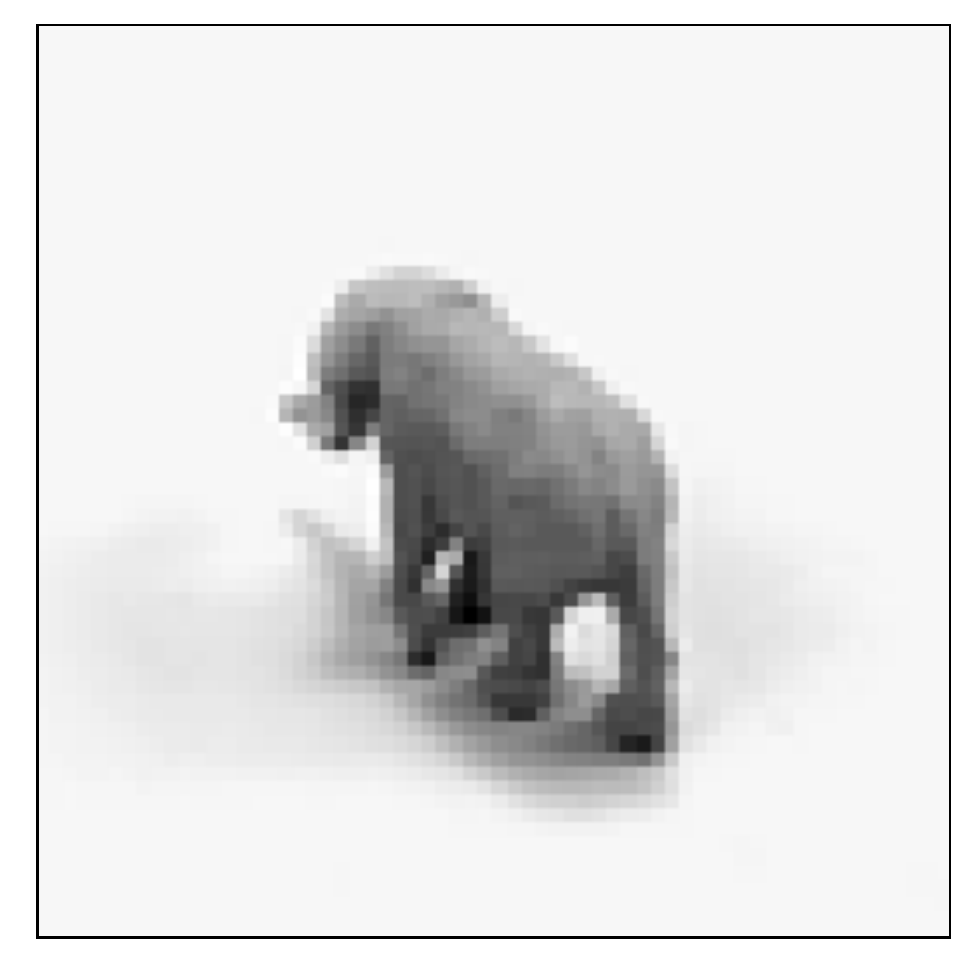}
        \end{subfigure}
    \begin{subfigure}[b]{0.3\linewidth}
    \includegraphics[scale=0.4]{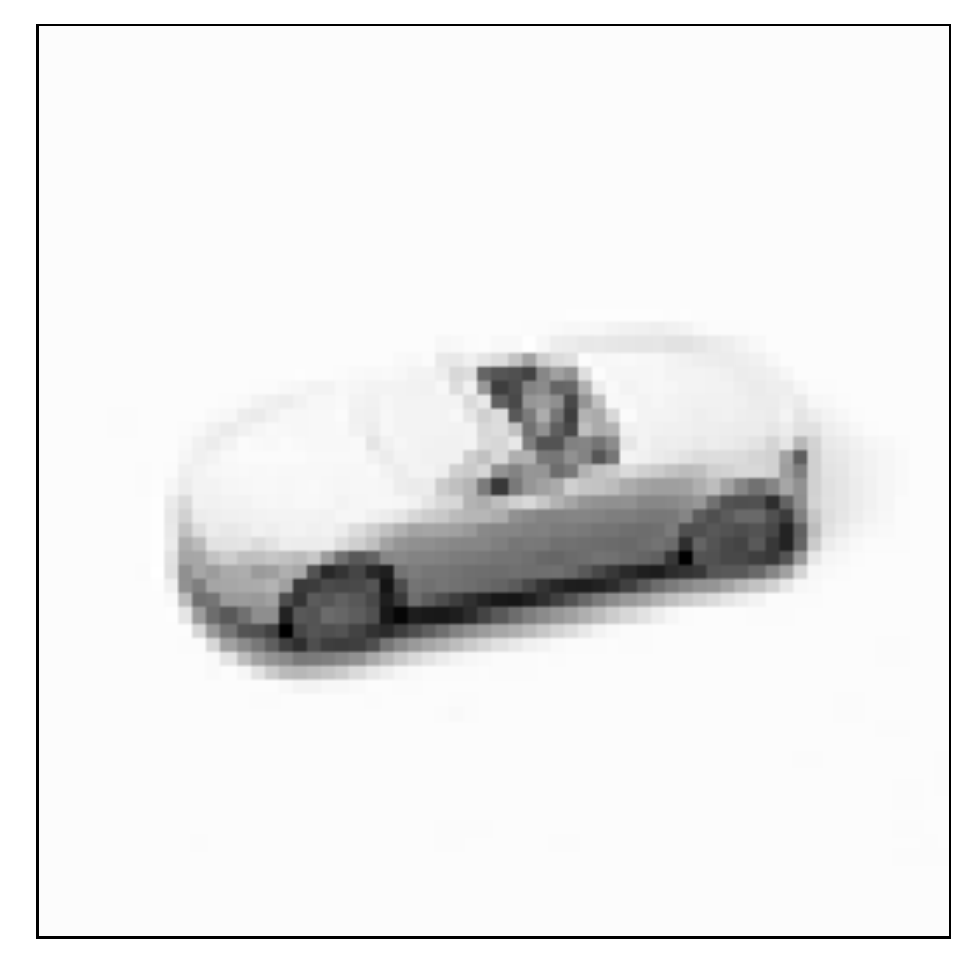}
        \end{subfigure}
    \begin{subfigure}[b]{0.3\linewidth}
    \includegraphics[scale = 0.4]{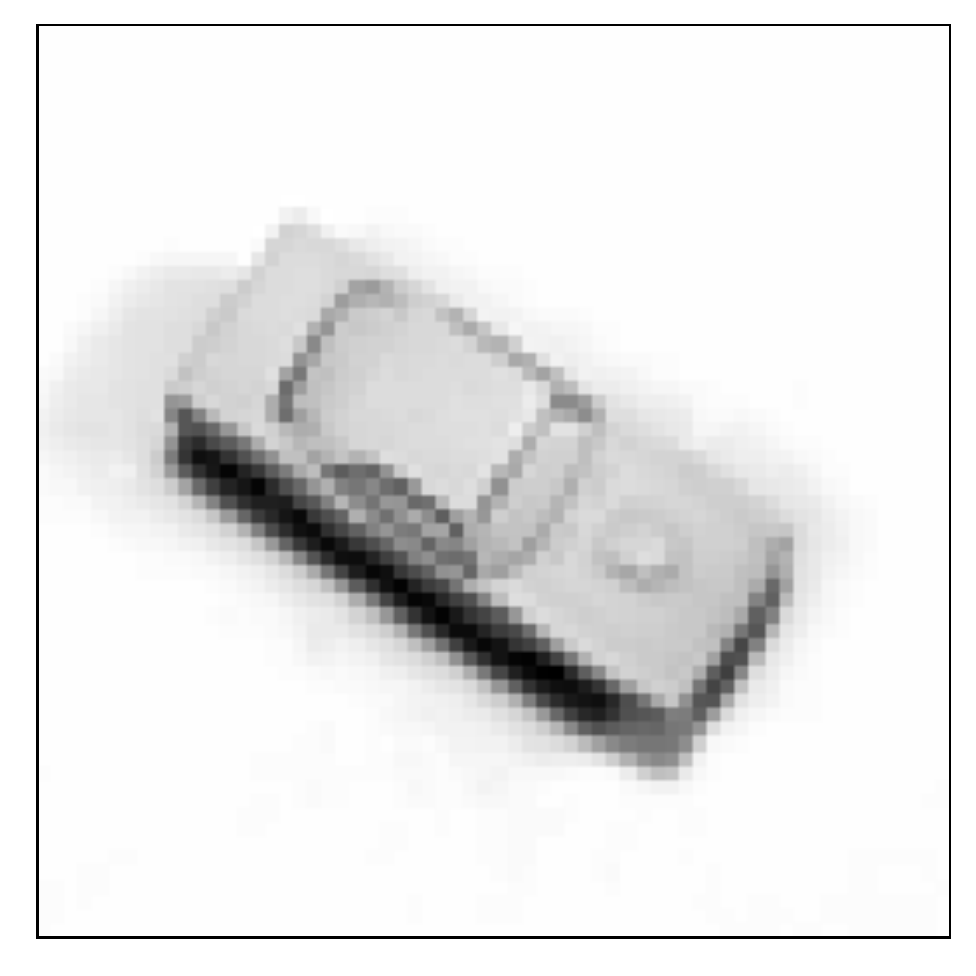}
        \end{subfigure}
    \caption{Some example images from the SmallNORB dataset.}
    \label{fig:smallnorb_examples}
\end{figure}

\begin{figure}[h!]
\centering
\begin{subfigure}[b]{0.325\linewidth}
    \includegraphics[width=\linewidth]{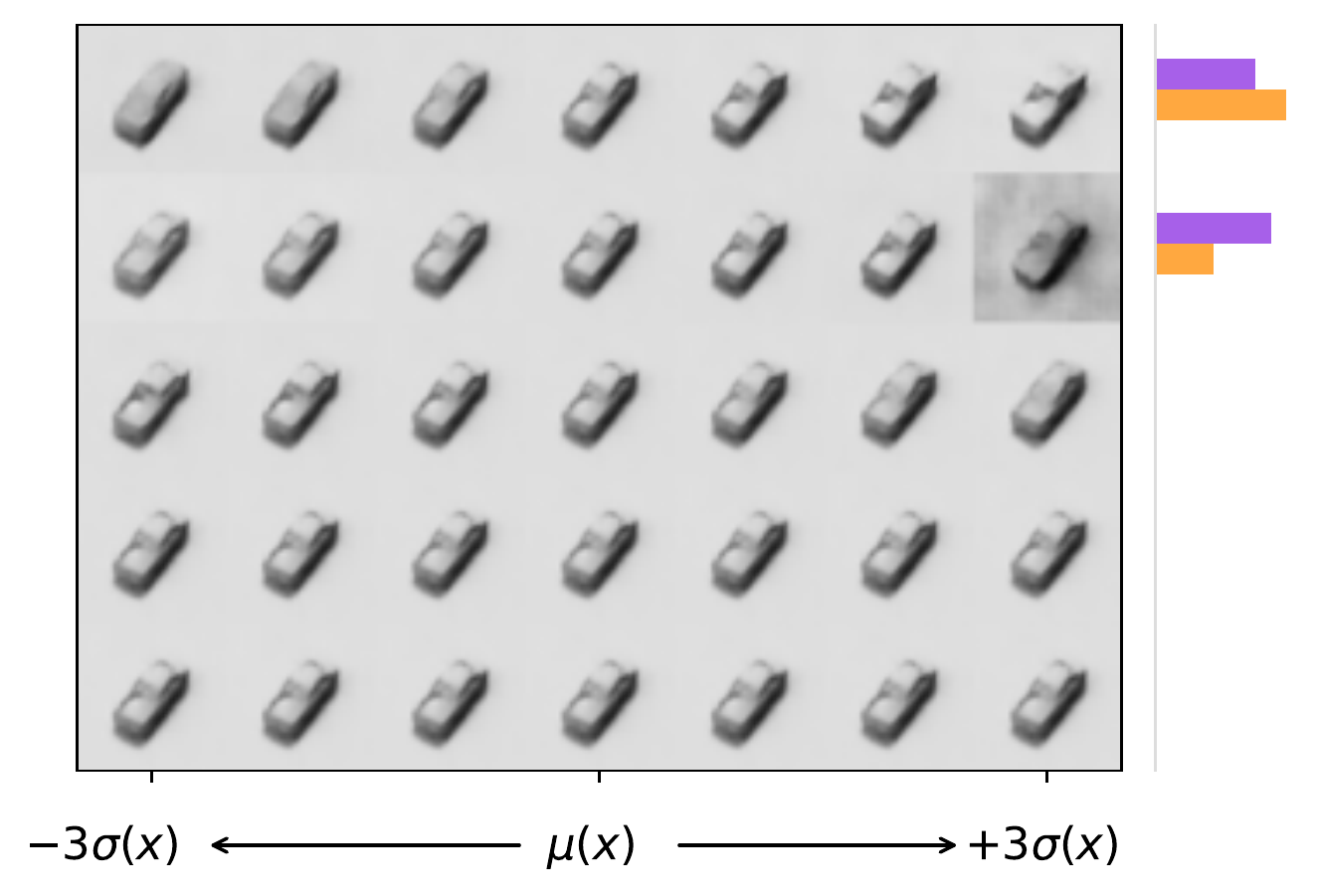}
        \end{subfigure}
\begin{subfigure}[b]{0.325\linewidth}
\centering
    \includegraphics[width=\linewidth]{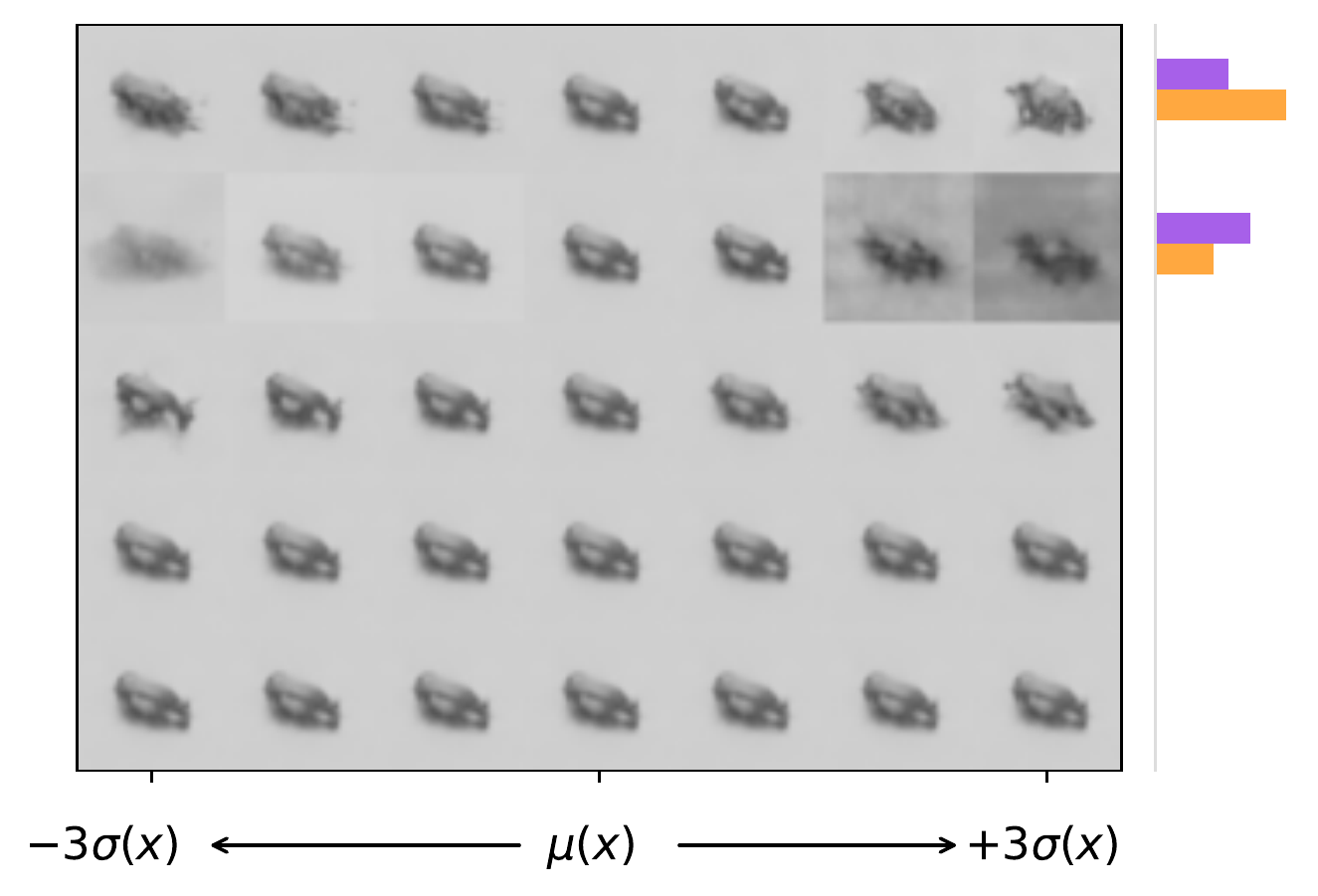}
        \end{subfigure}
\begin{subfigure}[b]{0.325\linewidth}
\centering
    \includegraphics[width=\linewidth]{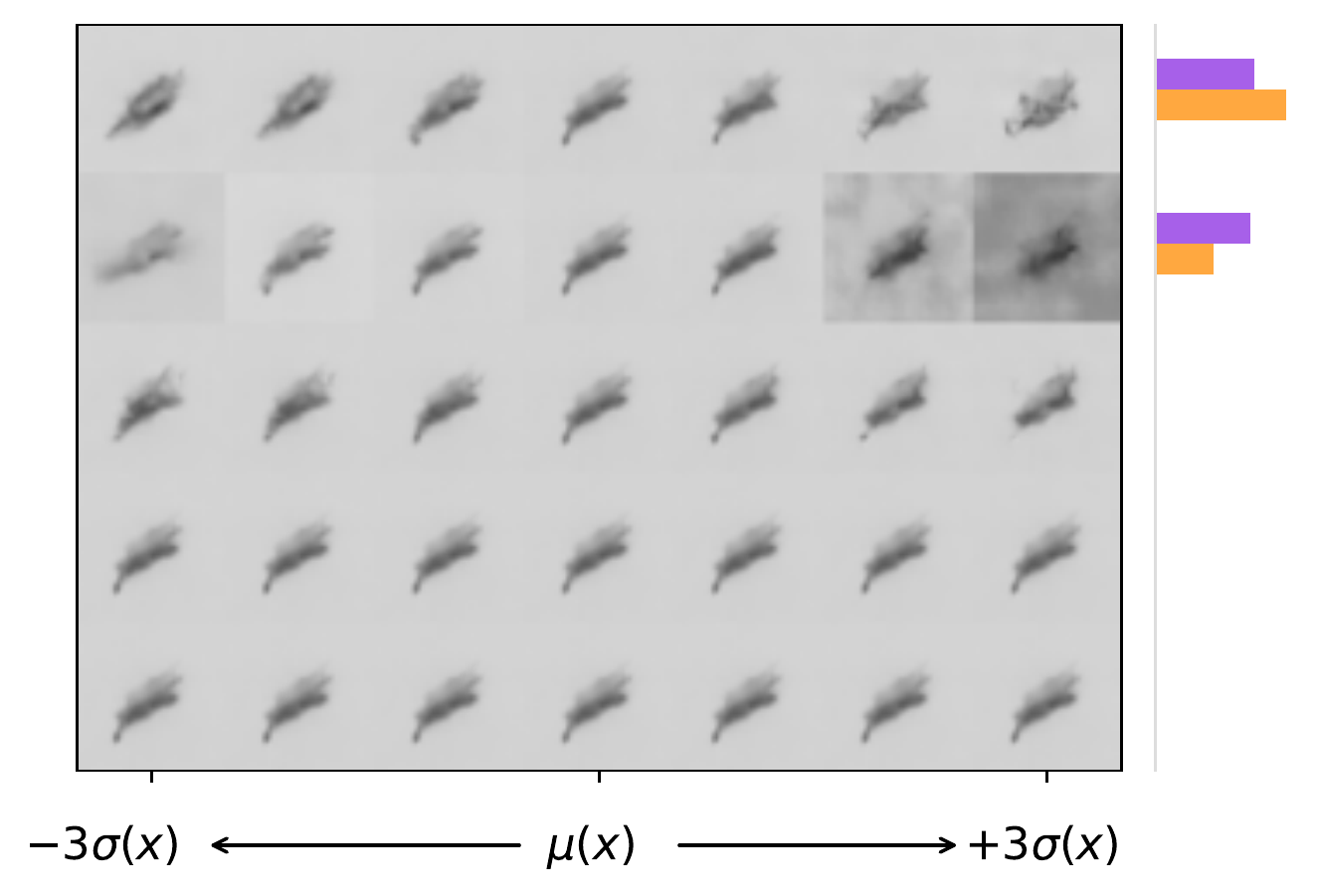}
        \end{subfigure}
\begin{subfigure}[b]{0.325\linewidth}
\centering
    \includegraphics[width=\linewidth]{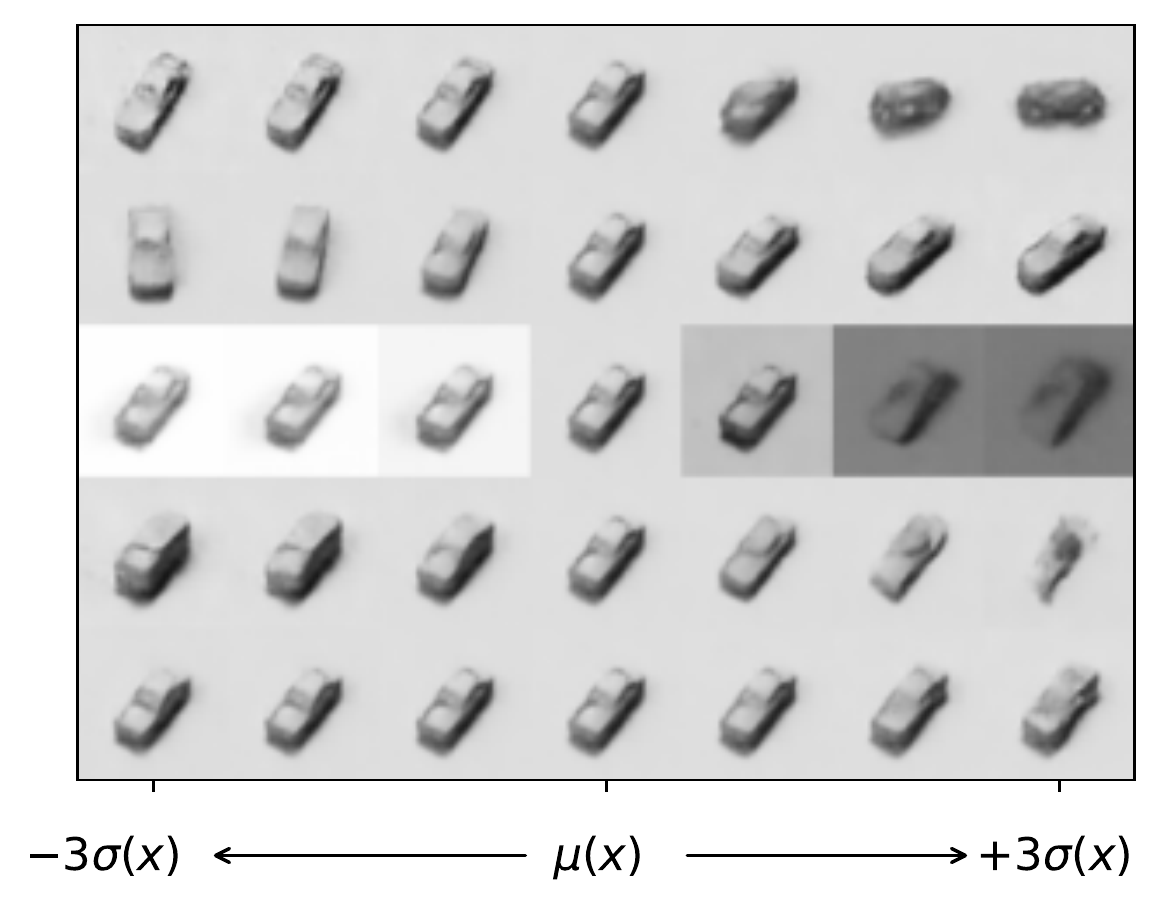}
        \end{subfigure}
\begin{subfigure}[b]{0.325\linewidth}
\centering
    \includegraphics[width=\linewidth]{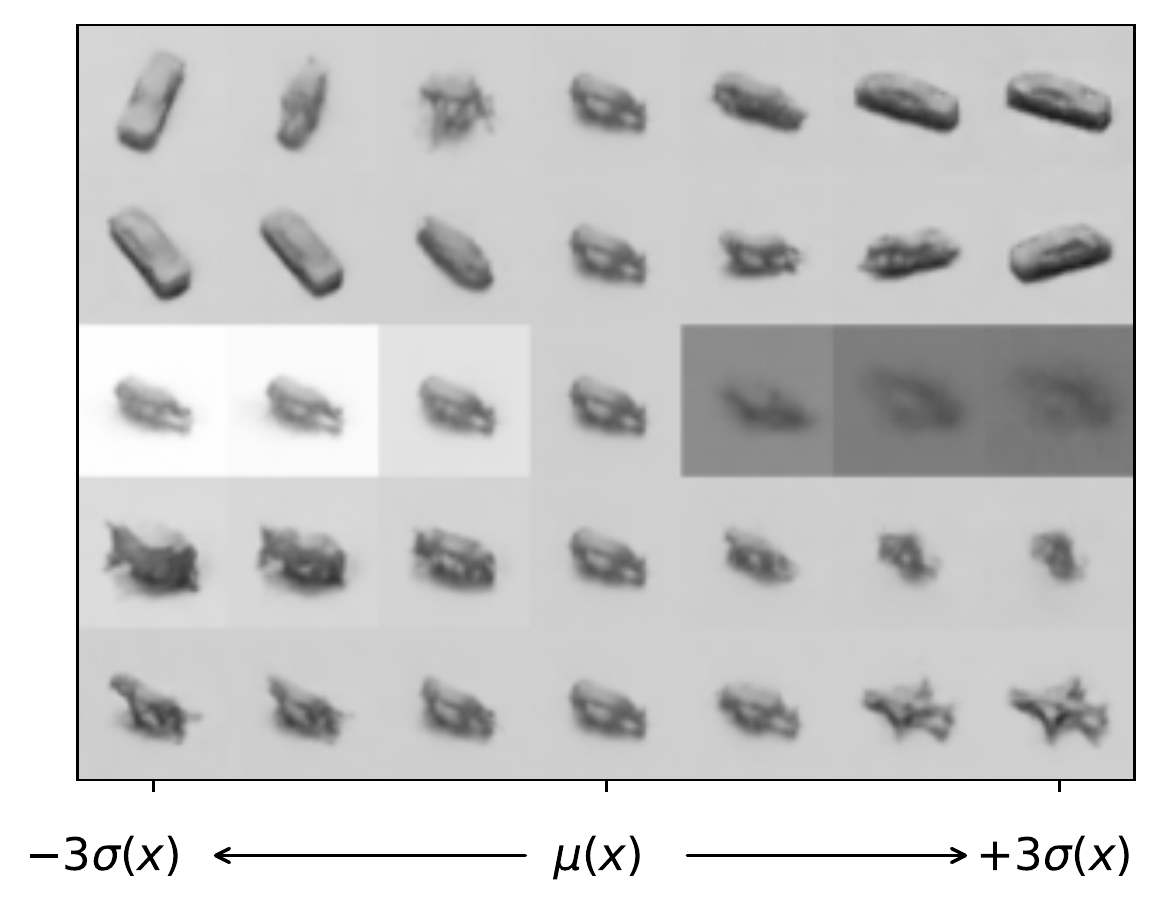}
        \end{subfigure}
\begin{subfigure}[b]{0.325\linewidth}
\centering
    \includegraphics[width=\linewidth]{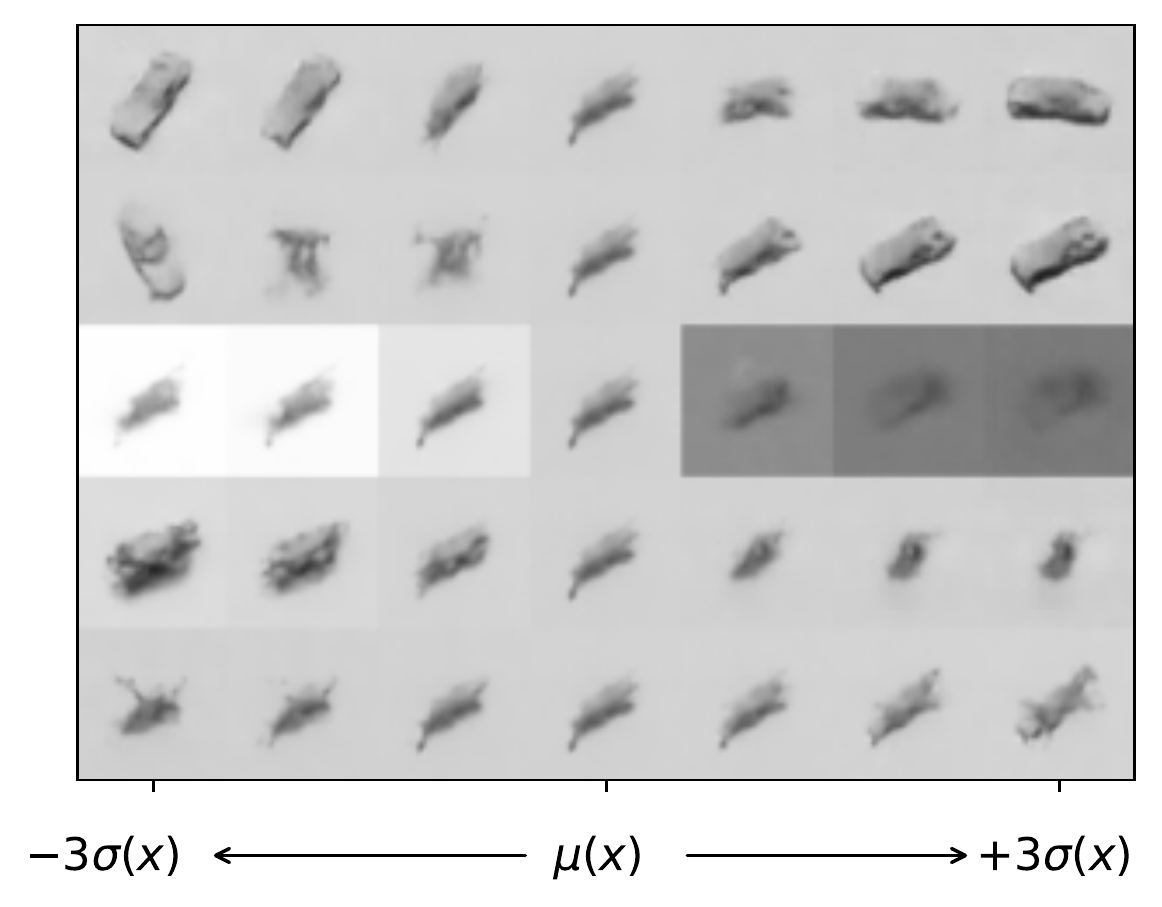}
        \end{subfigure}

\caption{\methName{} traversals of $\zc$ (first row) and $\zs$ on the SmallNORB dataset.}
\label{fig:traversals_smallnorb_clap}
\end{figure}

The SmallNORB dataset is a dataset of black and white images. In Figure~\ref{fig:smallnorb_examples}, we present some example images from the dataset.\\
The ground truth factors of variation are the object (9 classes), elevation of the camera (0 to 8), azimuth (even values from 0 to 34) and lightning condition (0 to 5). We create synthetic labels according to the following rules: 
for the first label, $y = 1$ if object type $\ge 5$ and lightning $\ge 3$, and $y = 0$ otherwise; 
for the second label, $y = 1$ if object type $\ge 5$ and lightning $< 3$, and $y = 0$ otherwise; 
for the third label, $y = 1$ if object type $< 5$ and lightning $\ge 3$, and $y = 0$ otherwise; 
for the final label, $y = 1$ if lightning $< 3$, and $y = 0$ otherwise. \\
We present \methName{} traversals on the SmallNORB dataset for $\zc$ (first row) and $\zs$ (second row) in Figure~\ref{fig:traversals_smallnorb_clap}.

\newpage
\section{Ablation studies}
\label{sec:ablations}

\subsection{No Group Sparsity}
\label{sec:no_group_sparsity}
\begin{figure}[h!]
\centering
\begin{subfigure}[b]{0.325\linewidth}
\centering
    \includegraphics[width=\linewidth]{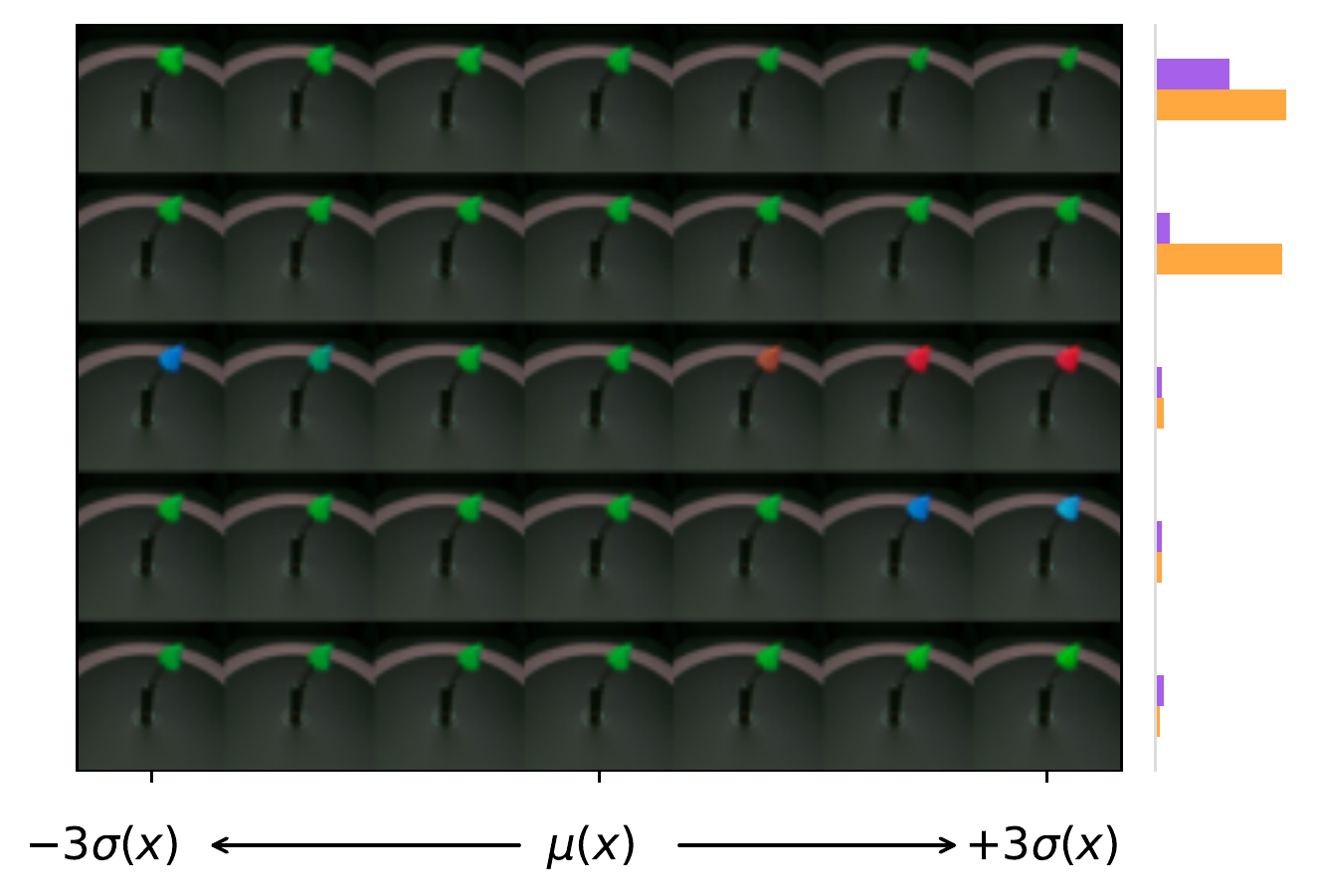}
        \end{subfigure}
\begin{subfigure}[b]{0.325\linewidth}
\centering
    \includegraphics[width=\linewidth]{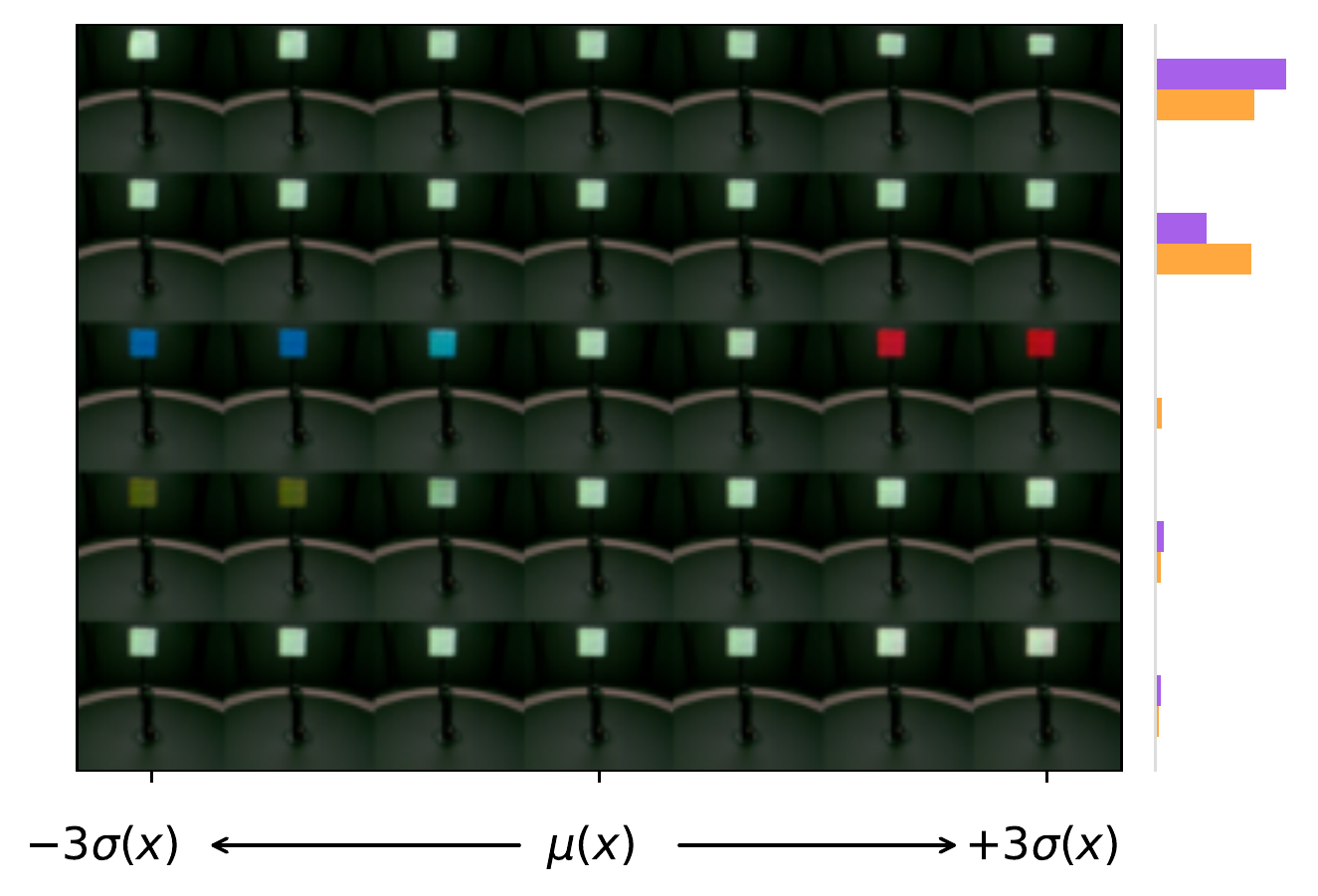}
        \end{subfigure}
\begin{subfigure}[b]{0.325\linewidth}
\centering
    \includegraphics[width=\linewidth]{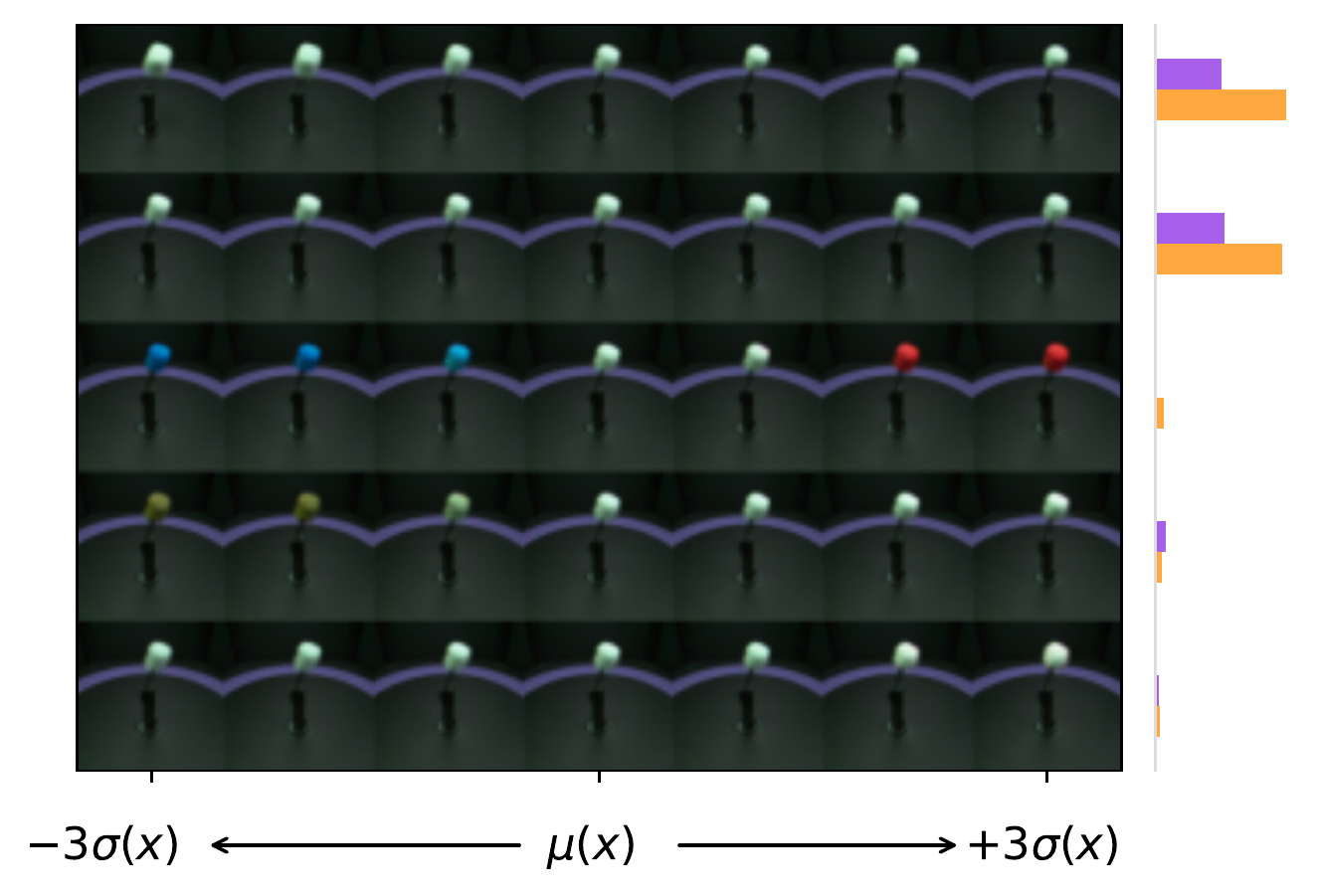}
        \end{subfigure}

\caption{No-group-sparsity \methName{} traversals of $\zc$ on the MPI3D dataset.}
\label{fig:traversals_clap_nosparsity}
\end{figure}

In Figure~\ref{fig:traversals_clap_nosparsity} we present traversals for \methName{} trained without group sparsity, i.e. $\rho_n(f, \psi) = 0$.

\subsection{Prediction-Only Model}
\label{sec:prediction_only}
\begin{figure}[h!]
\centering
\begin{subfigure}[b]{0.325\linewidth}
\centering
    \includegraphics[width=\linewidth]{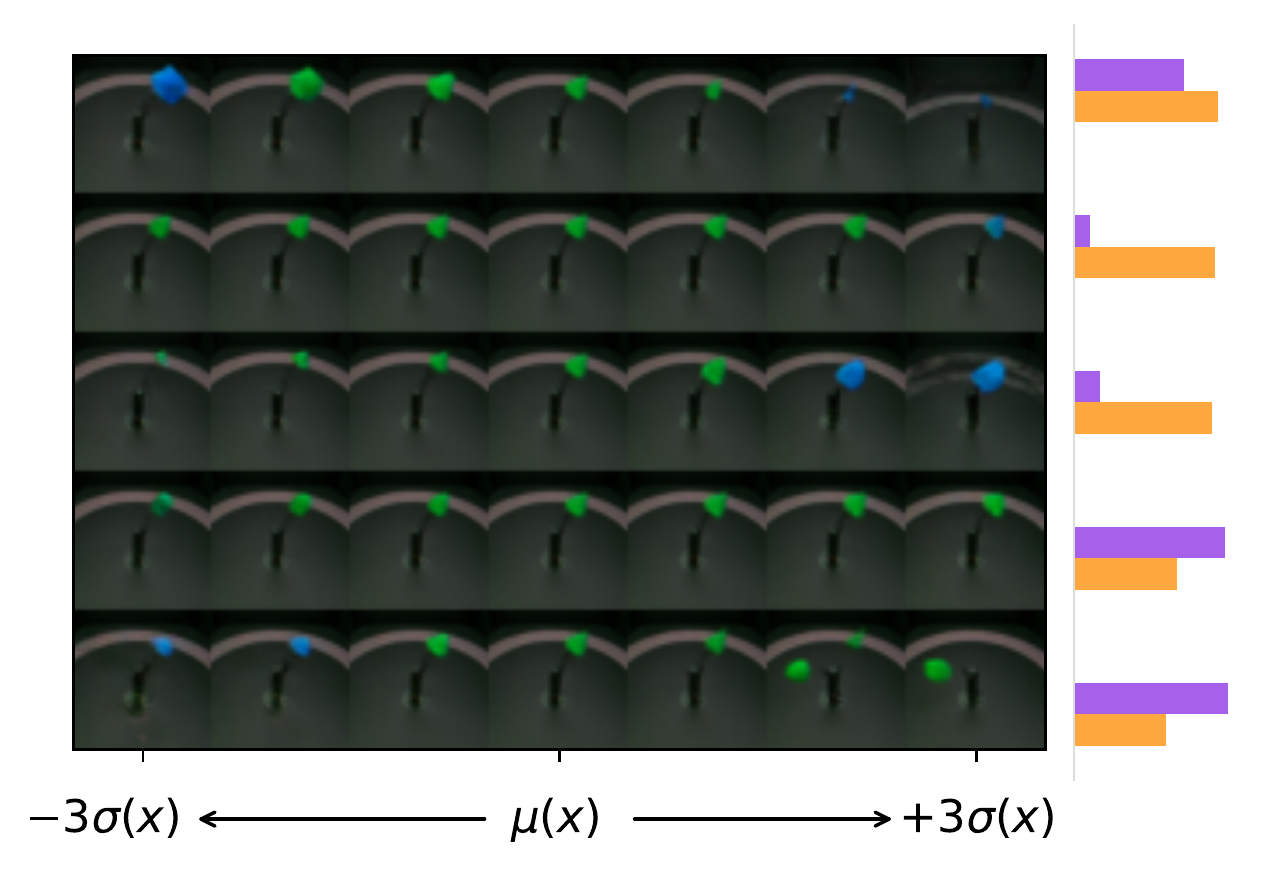}
        \end{subfigure}
\begin{subfigure}[b]{0.325\linewidth}
\centering
    \includegraphics[width=\linewidth]{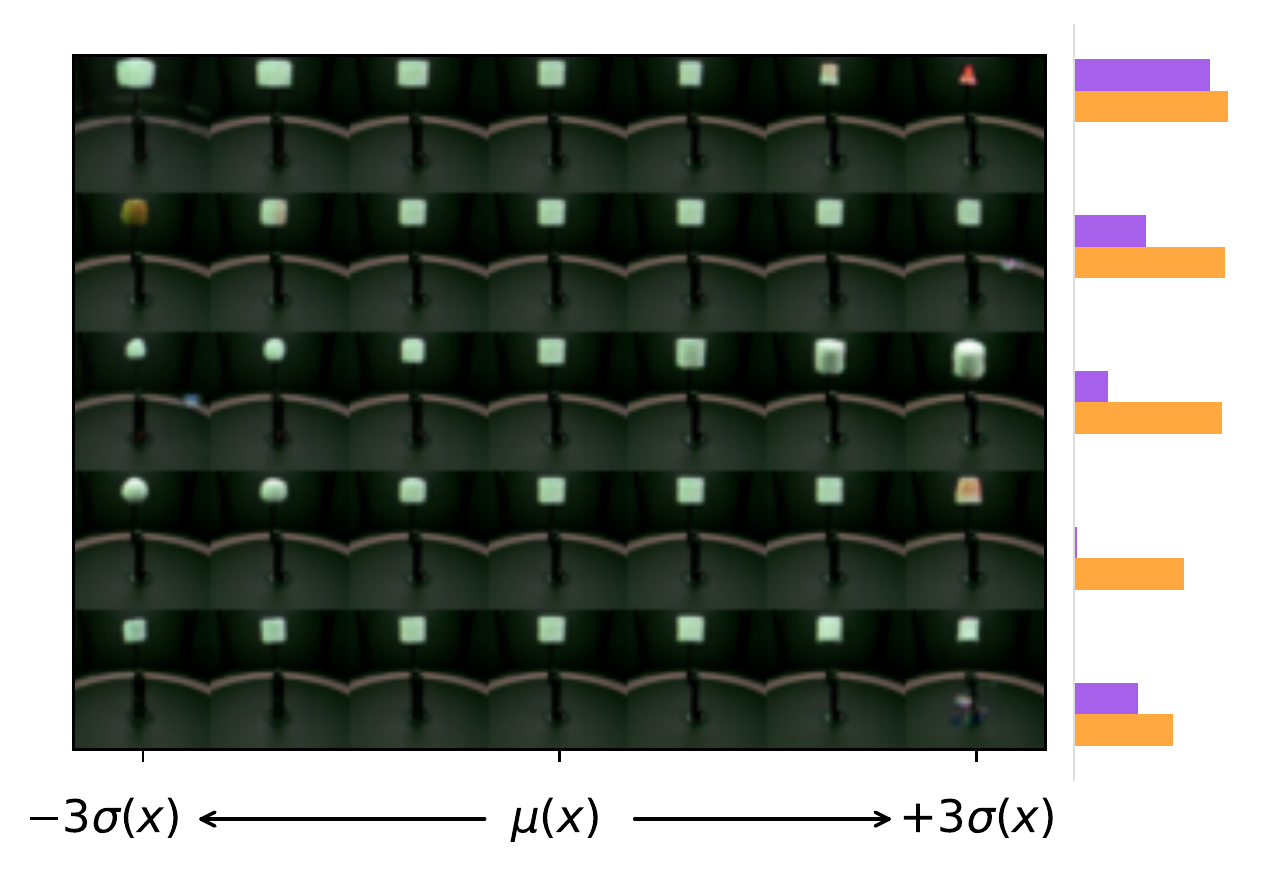}
        \end{subfigure}
\begin{subfigure}[b]{0.325\linewidth}
\centering
    \includegraphics[width=\linewidth]{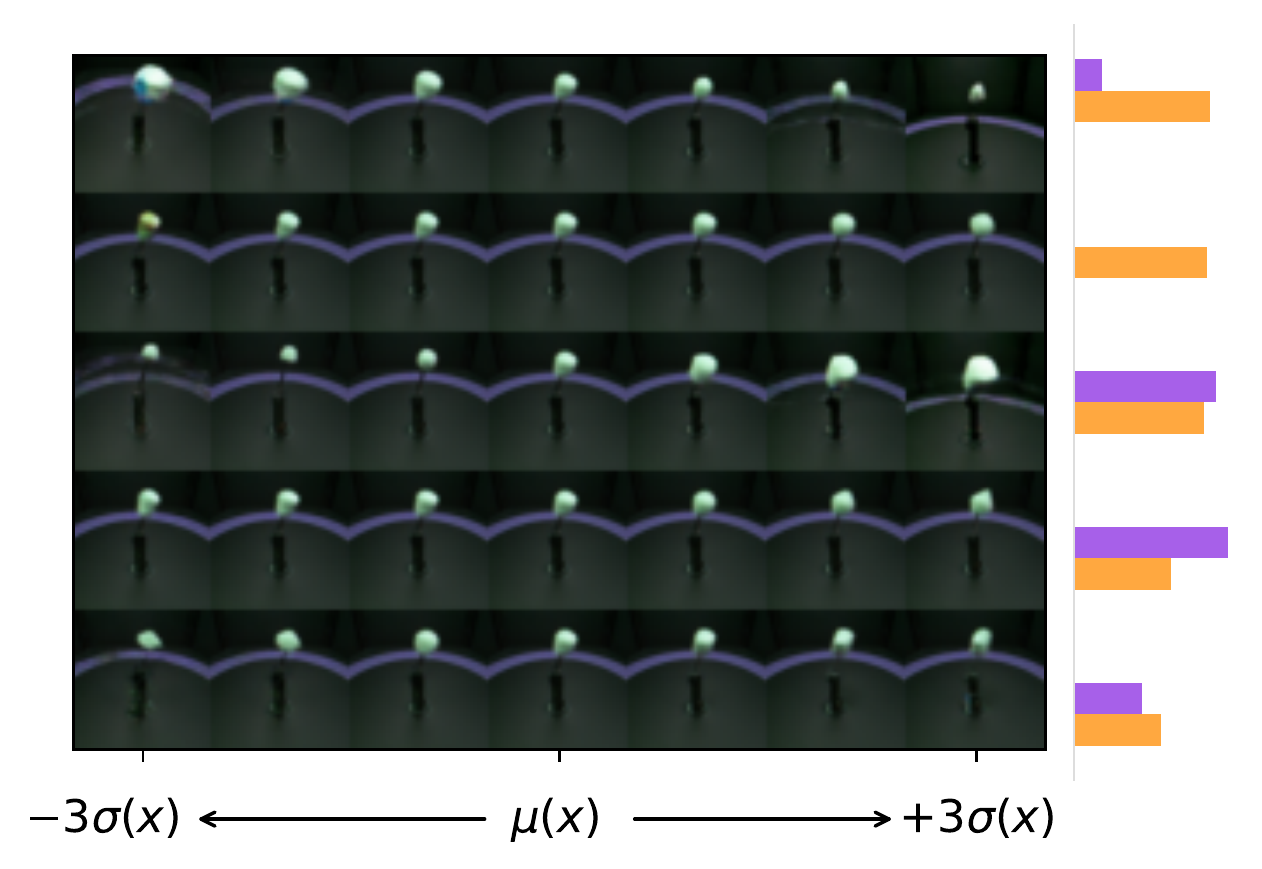}
        \end{subfigure}

\caption{Prediction-only \methName{} traversals of $\zc$ on the MPI3D dataset.}
\label{fig:traversals_clap_onlypred}
\end{figure}

In Figure~\ref{fig:traversals_clap_onlypred} we present traversals for \methName{} trained only on the prediction part of the loss in Eq.~\ref{eqn:CbPVAE}.

\subsection{MPI3D with One Label Only}
\label{sec:mpi_one_label}
\begin{figure}[h!]
\centering
\begin{subfigure}[b]{0.325\linewidth}
\centering
    \includegraphics[width=\linewidth]{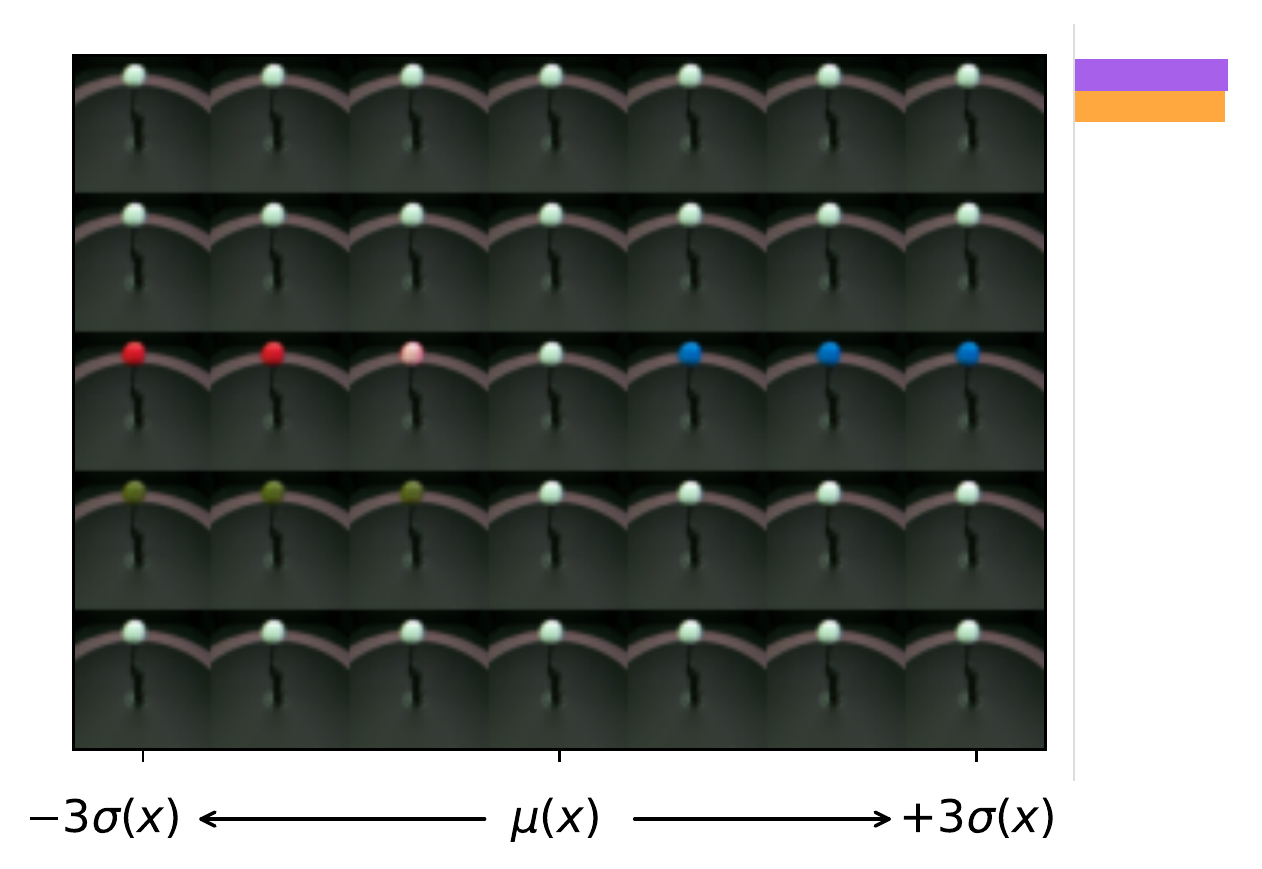}
        \end{subfigure}
\begin{subfigure}[b]{0.325\linewidth}
\centering
    \includegraphics[width=\linewidth]{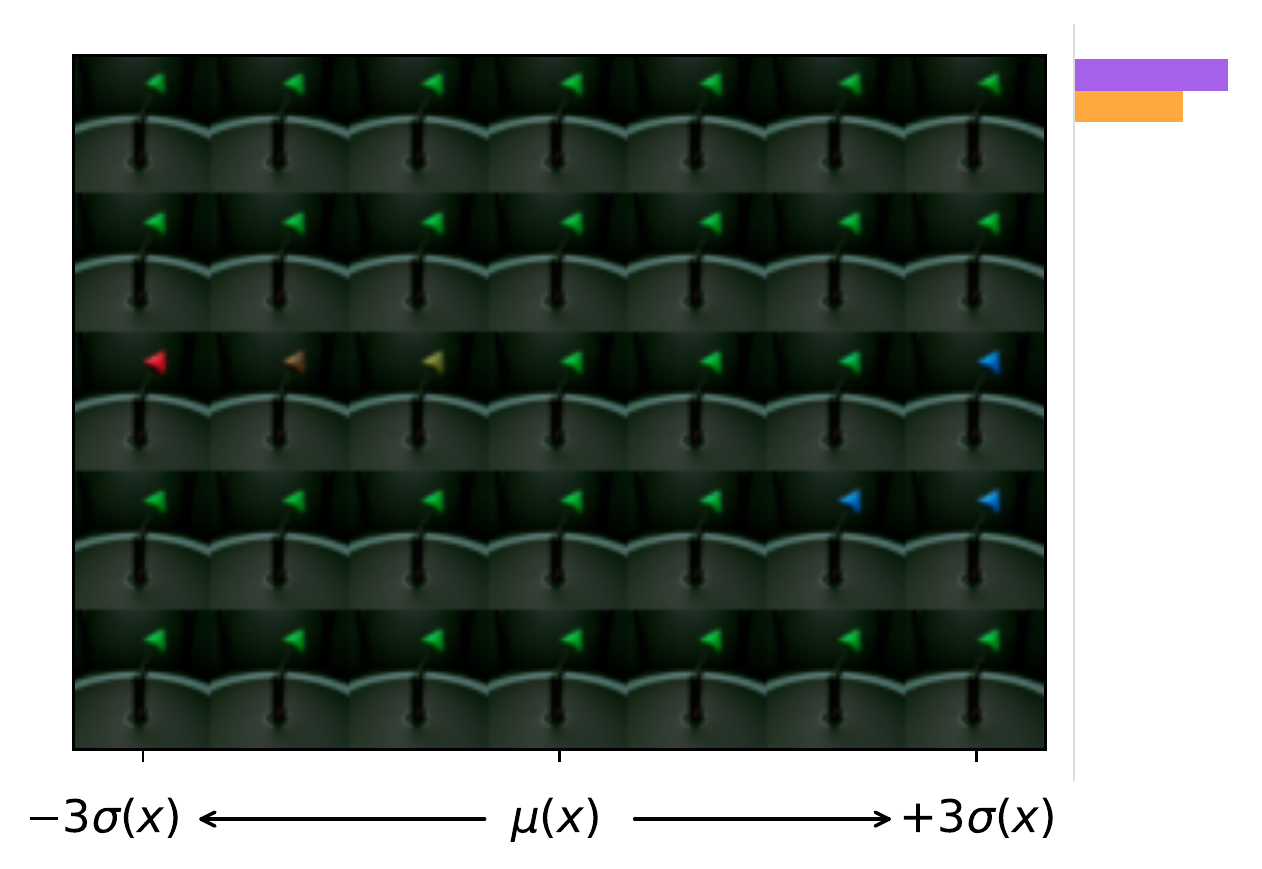}
        \end{subfigure}
\begin{subfigure}[b]{0.325\linewidth}
\centering
    \includegraphics[width=\linewidth]{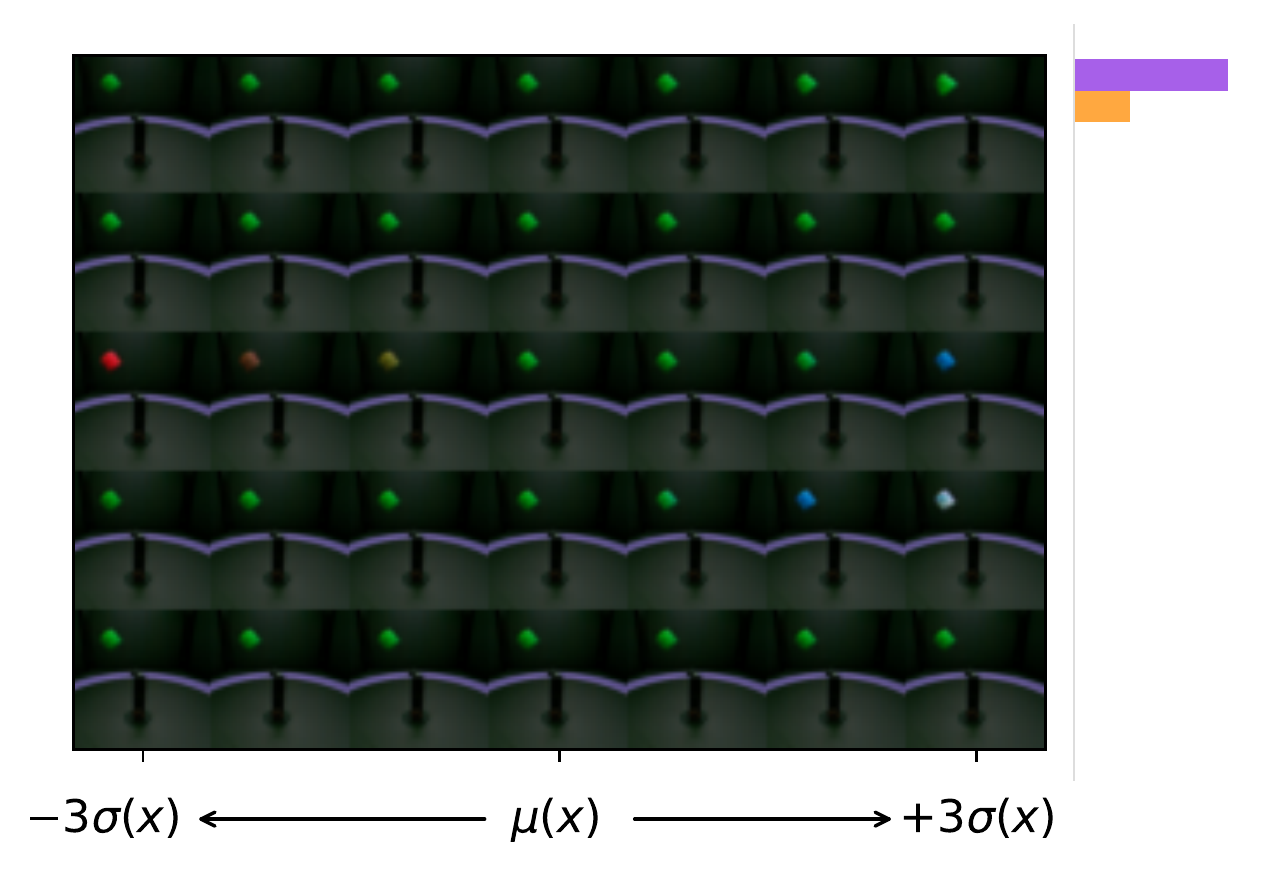}
        \end{subfigure}
\caption{Prediction-only \methName{} traversals of $\zc$ on the MPI3D dataset with only one label.}
\label{fig:traversals_clap_onelabel}
\end{figure}

In Figure~\ref{fig:traversals_clap_onelabel} we present traversals for \methName{} trained on the MPI3D dataset where only the first label is made available for supervision.

\newpage
\section{CCVAE on Shapes3D}
\label{sec:ccvae on shapes3d appendix}
\begin{figure}[h!]
\centering
\begin{subfigure}[b]{0.325\linewidth}
\centering
    \includegraphics[width=\linewidth]{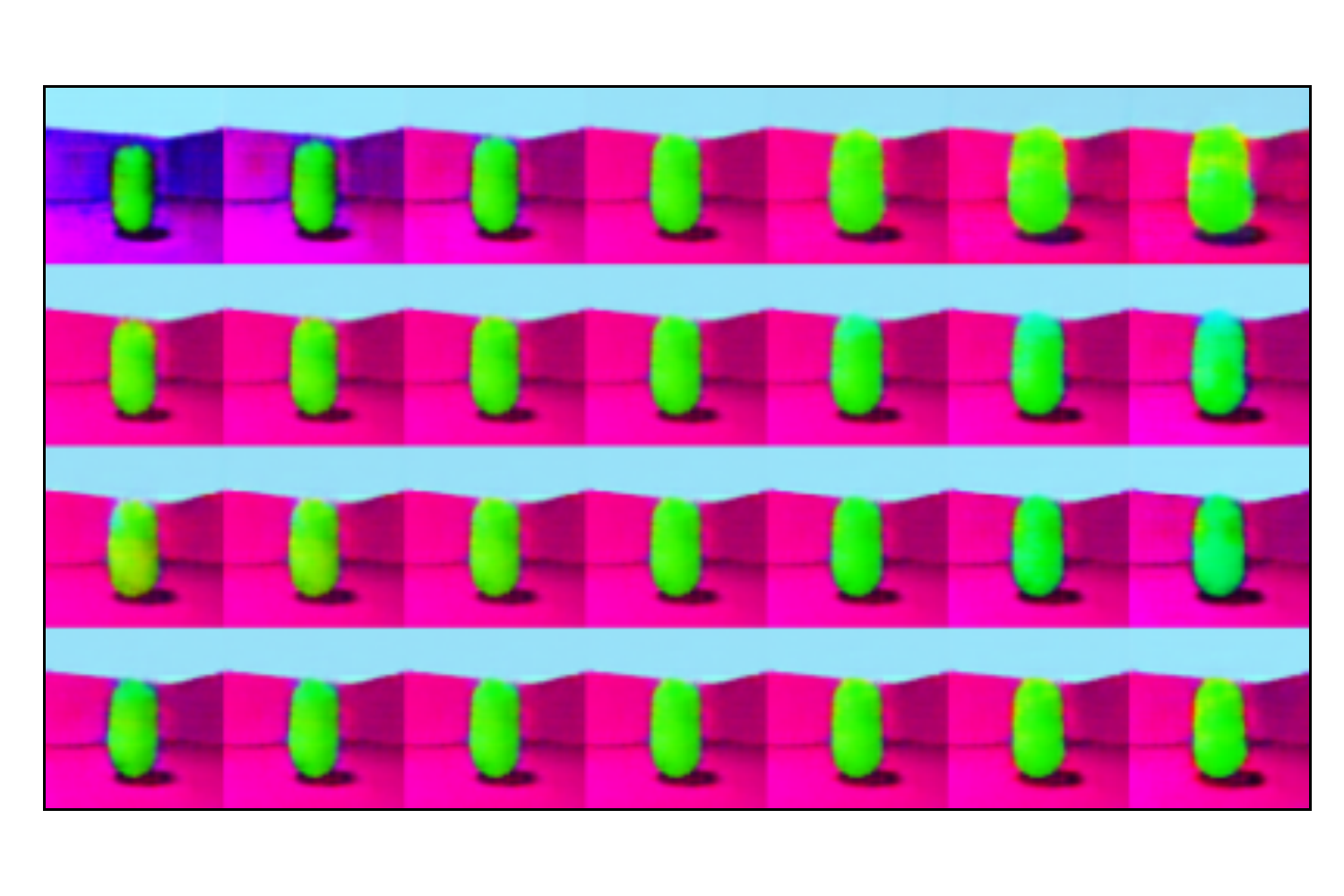}
        \end{subfigure}
\begin{subfigure}[b]{0.325\linewidth}
\centering
    \includegraphics[width=\linewidth]{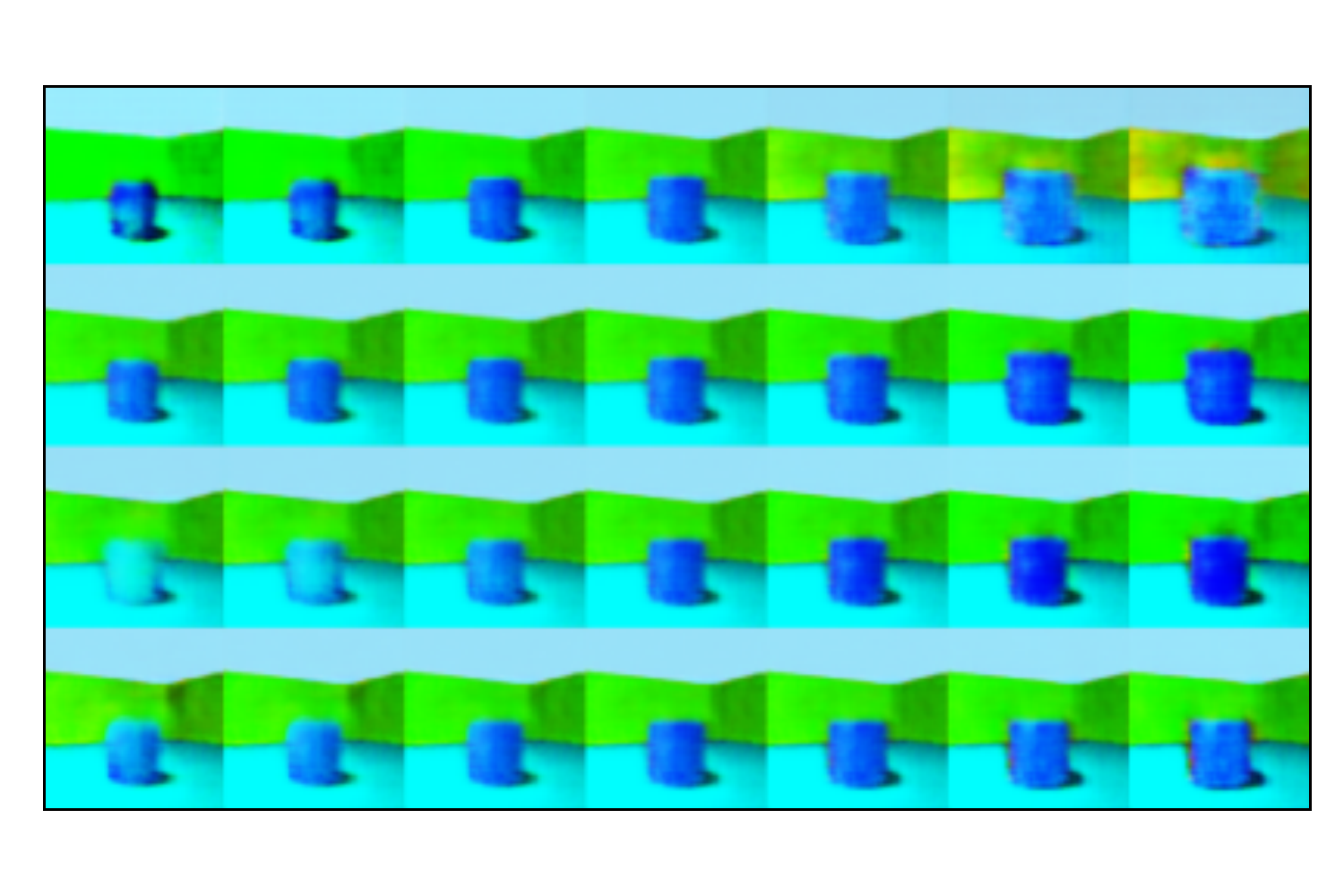}
        \end{subfigure}
\begin{subfigure}[b]{0.325\linewidth}
\centering
    \includegraphics[width=\linewidth]{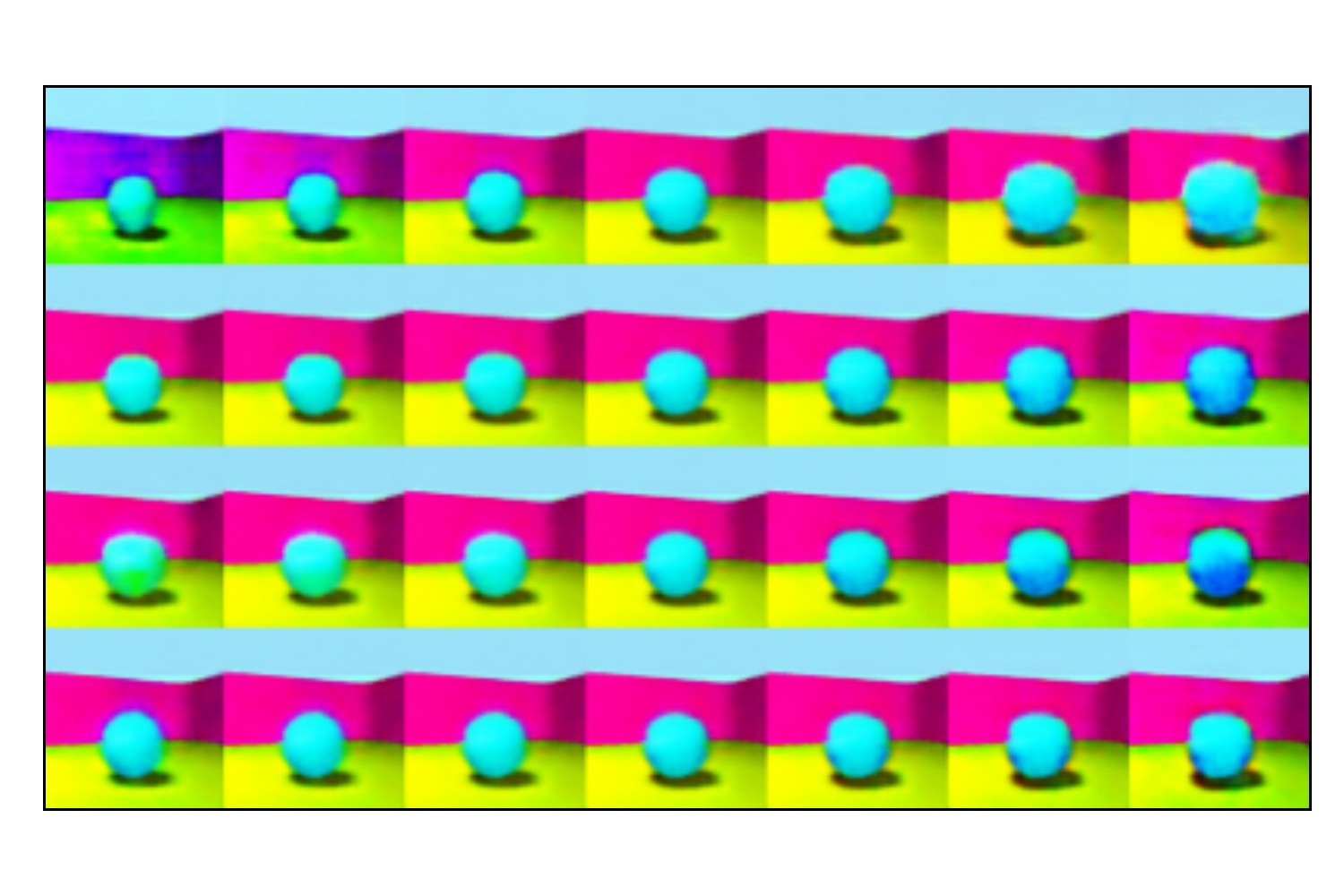}
        \end{subfigure}
\caption{CCVAE traversals of $\zc$ on the Shapes3D dataset.}
\label{fig:traversals_ccvae_shapes3d}
\end{figure}

In Figure~\ref{fig:traversals_ccvae_shapes3d} we show the traversals of CCVAE on the Shapes3D dataset.

\newpage
\section{\chest{} traversals}
\label{sec:chestxray_traversals}
\begin{figure}[h!]
\centering
\includegraphics[width=0.8\textwidth]{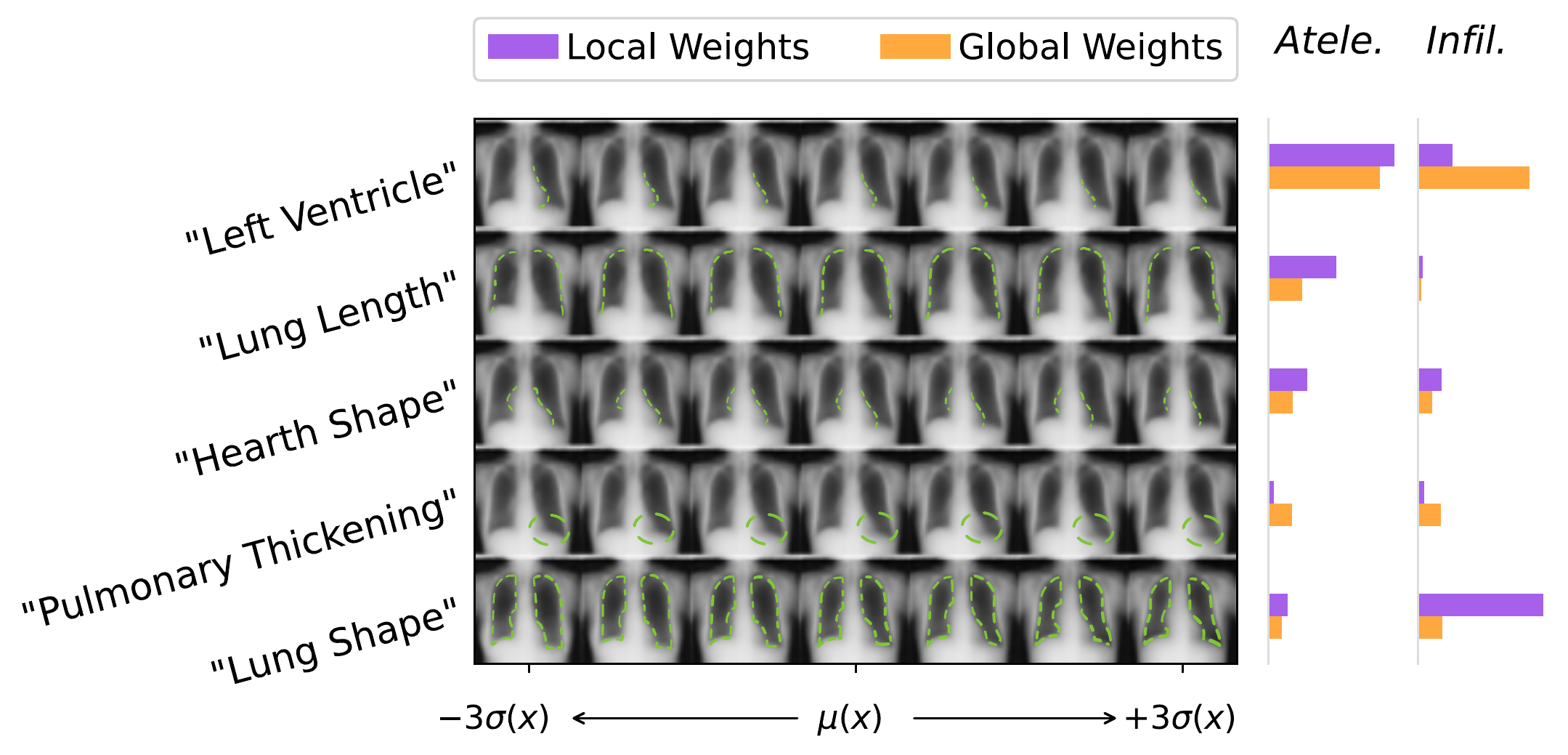}
    \caption{Complete \methName{} traversals, corresponding to those shown in Figure~\ref{fig:traversals_chestxray_clap}.}
    \label{fig:CLAP_chestxray_complete_big}
\end{figure}

We include the complete \methName{} traversals on the \chest{} dataset in Figure~\ref{fig:CLAP_chestxray_complete_big}.

\begin{figure}[h!]
\centering
\begin{subfigure}[b]{0.325\linewidth}
\centering
    \includegraphics[width=\linewidth]{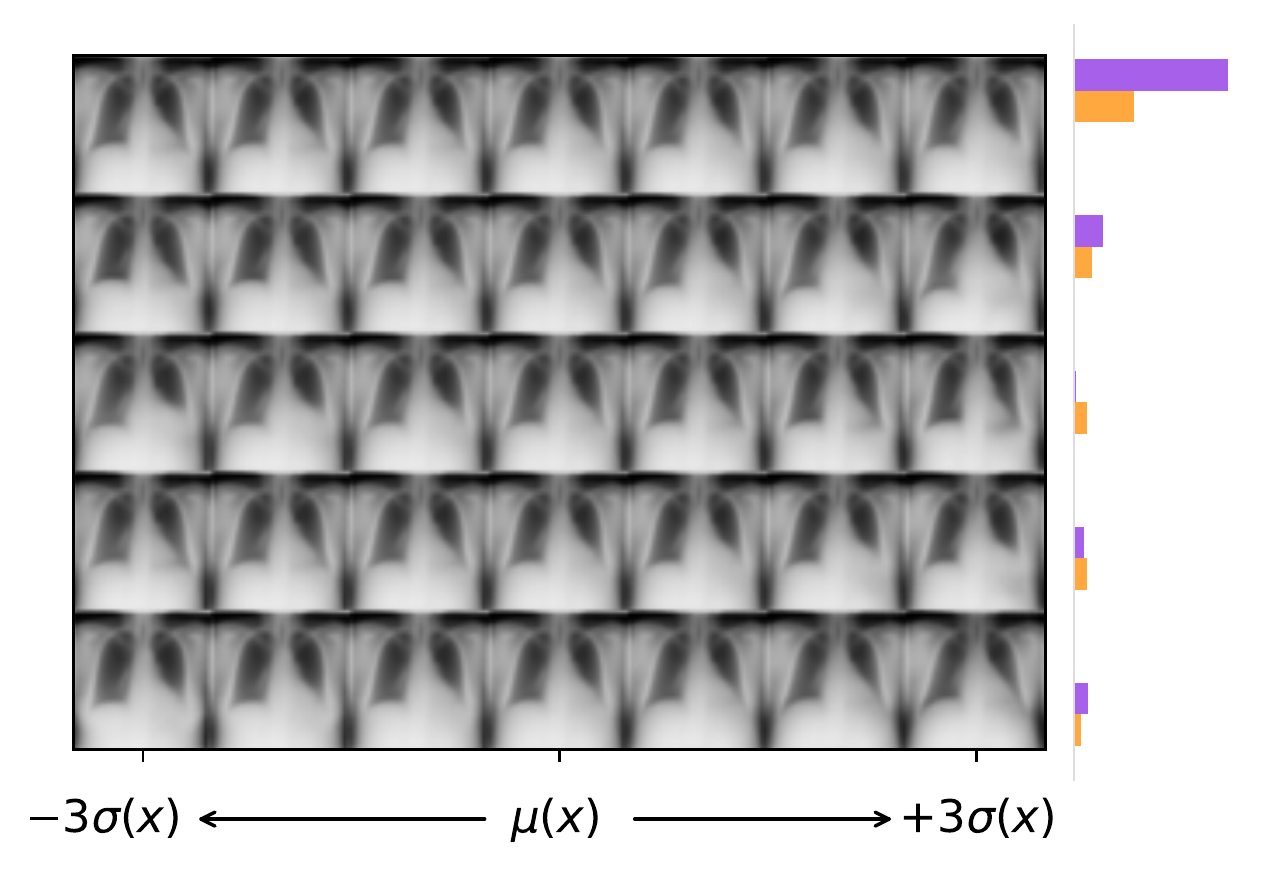}
        \end{subfigure}
\begin{subfigure}[b]{0.325\linewidth}
\centering
    \includegraphics[width=\linewidth]{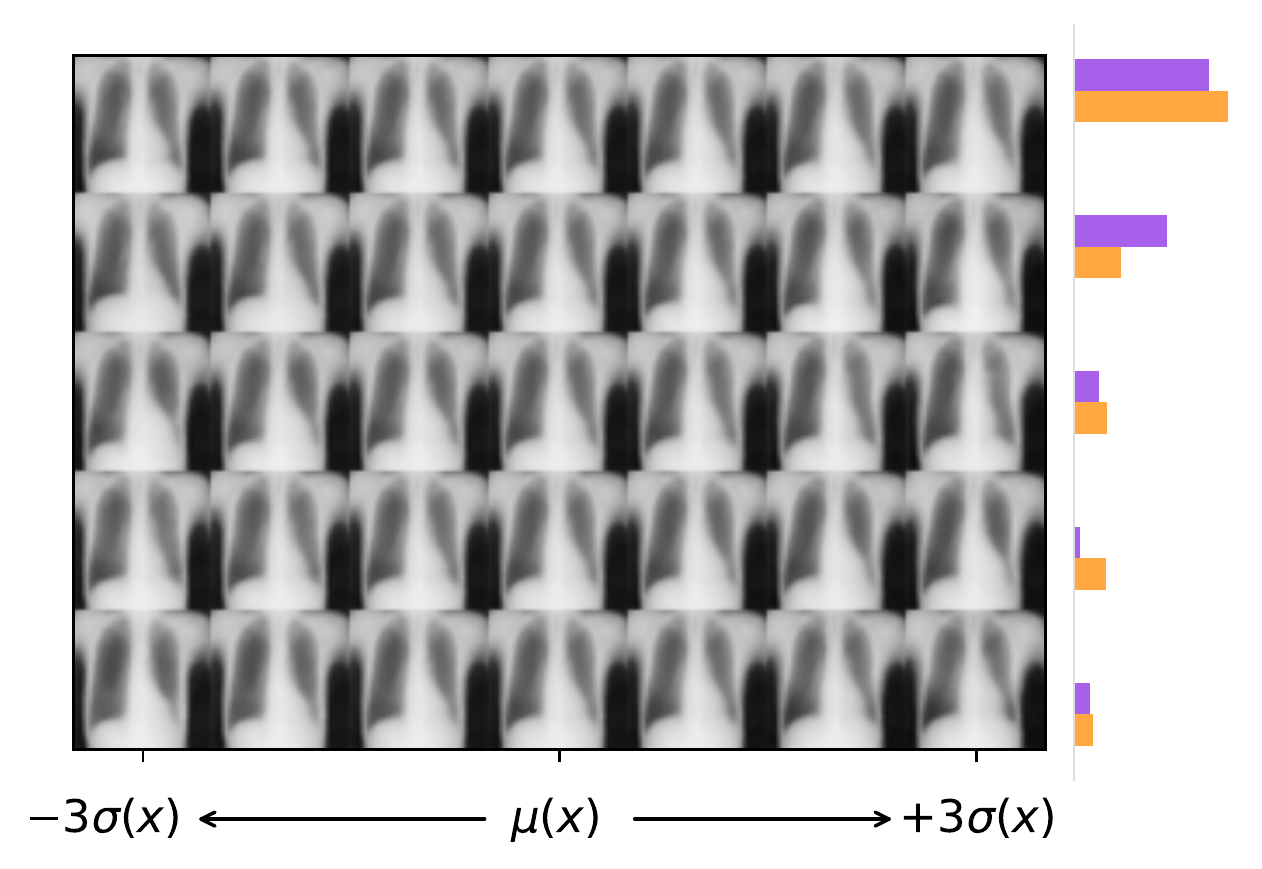}
        \end{subfigure}
\begin{subfigure}[b]{0.325\linewidth}
\centering
    \includegraphics[width=\linewidth]{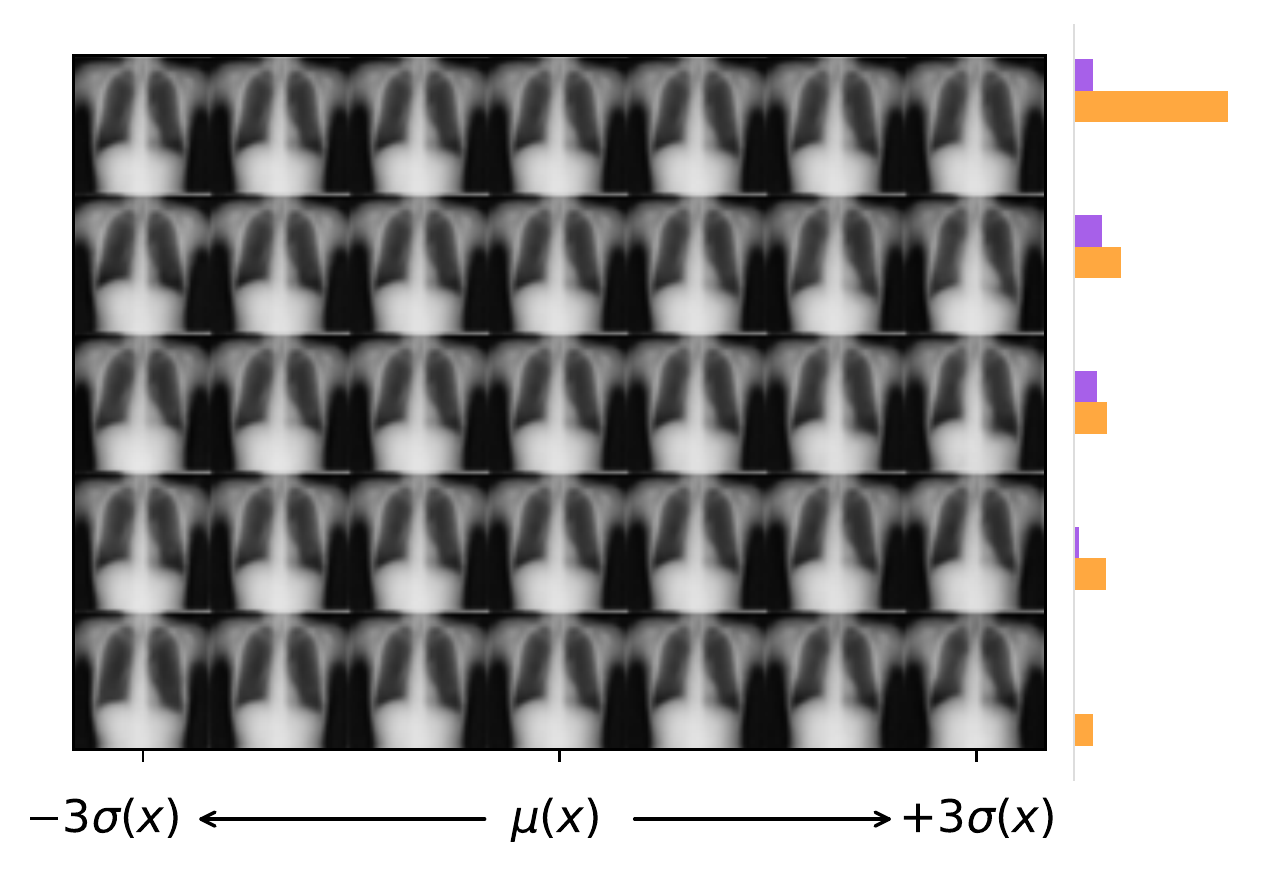}
        \end{subfigure}
\begin{subfigure}[b]{0.325\linewidth}
\centering
    \includegraphics[width=\linewidth]{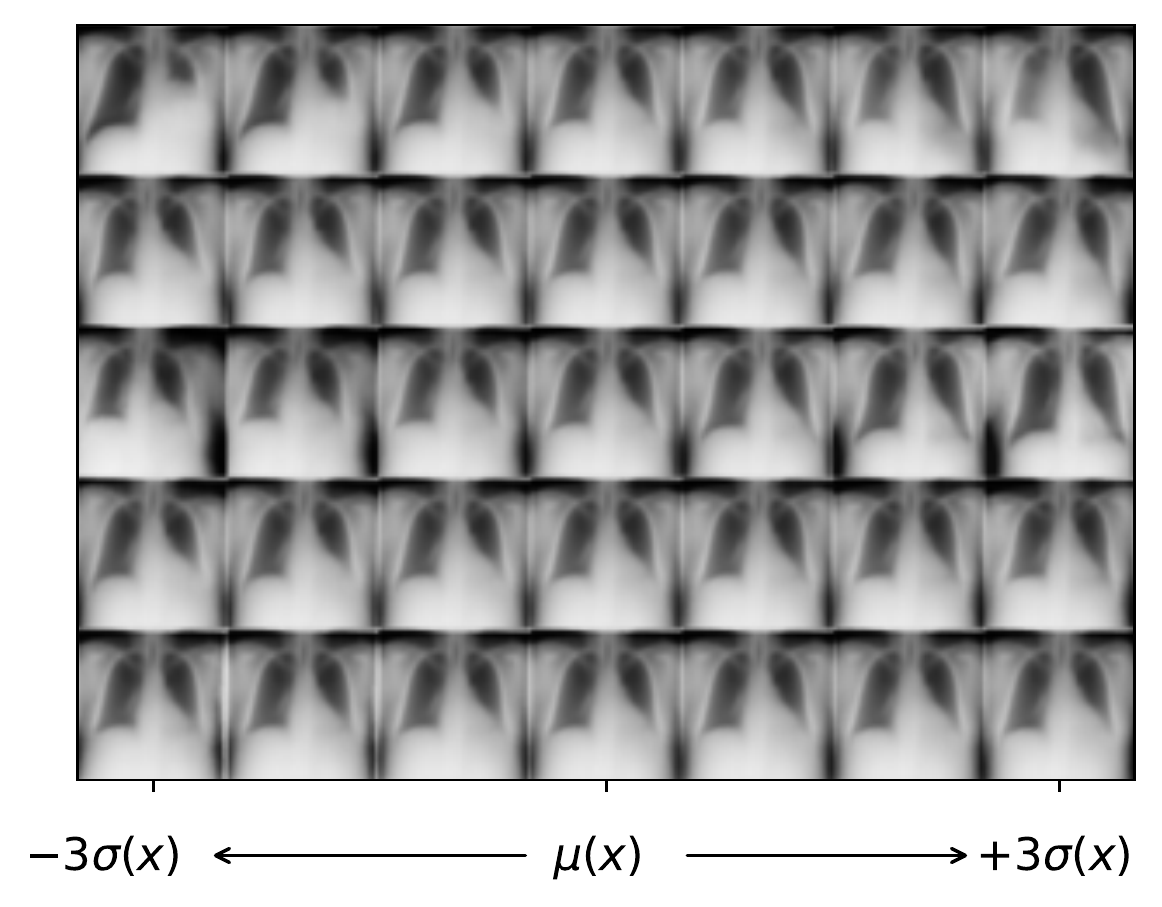}
        \end{subfigure}
\begin{subfigure}[b]{0.325\linewidth}
\centering
    \includegraphics[width=\linewidth]{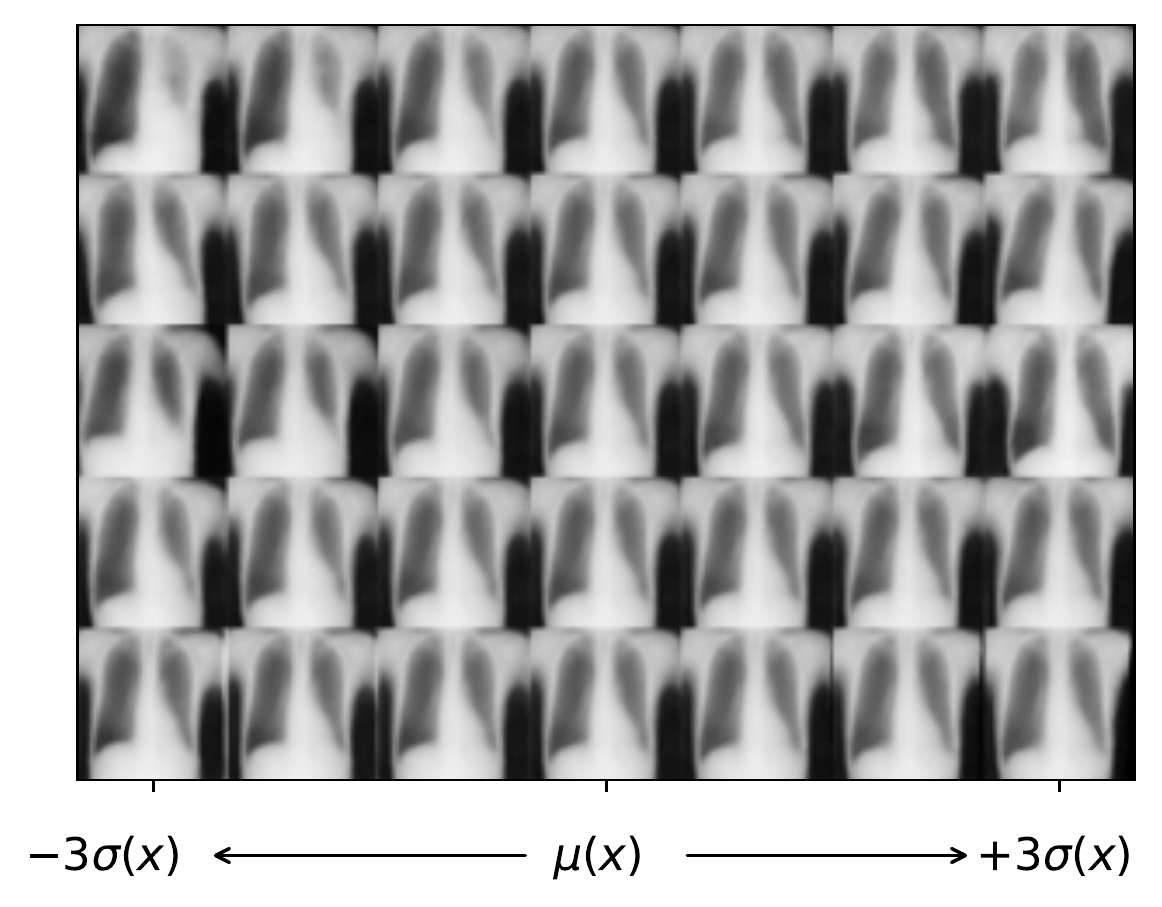}
        \end{subfigure}
\begin{subfigure}[b]{0.325\linewidth}
\centering
    \includegraphics[width=\linewidth]{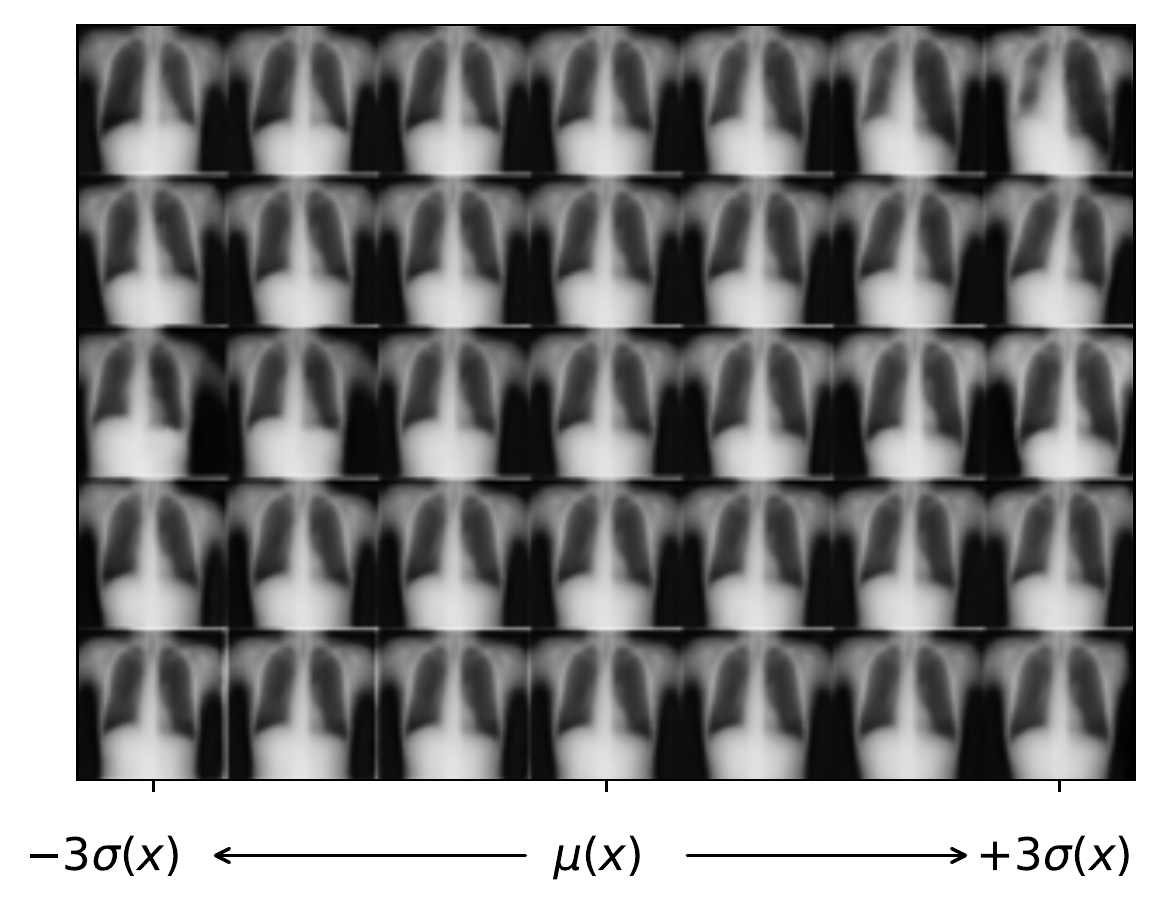}
        \end{subfigure}
\caption{\methName{} traversals of $\zc$ (first row) and $\zs$ on the \chest{} dataset.}
\label{fig:traversals_chestxray_clap_supp}
\end{figure}

We present additional \chest{} traversals of \methName{} for $\zc$ (first row) and $\zs$ (second row) in Figure~\ref{fig:traversals_chestxray_clap_supp}.

\newpage
\subsection{Glossary and Reading of \chest{}}
\label{sec:chestxray_glossary}
\begin{figure}[h!]
\centering
\begin{subfigure}[b]{0.325\linewidth}
    \includegraphics[width=\linewidth]{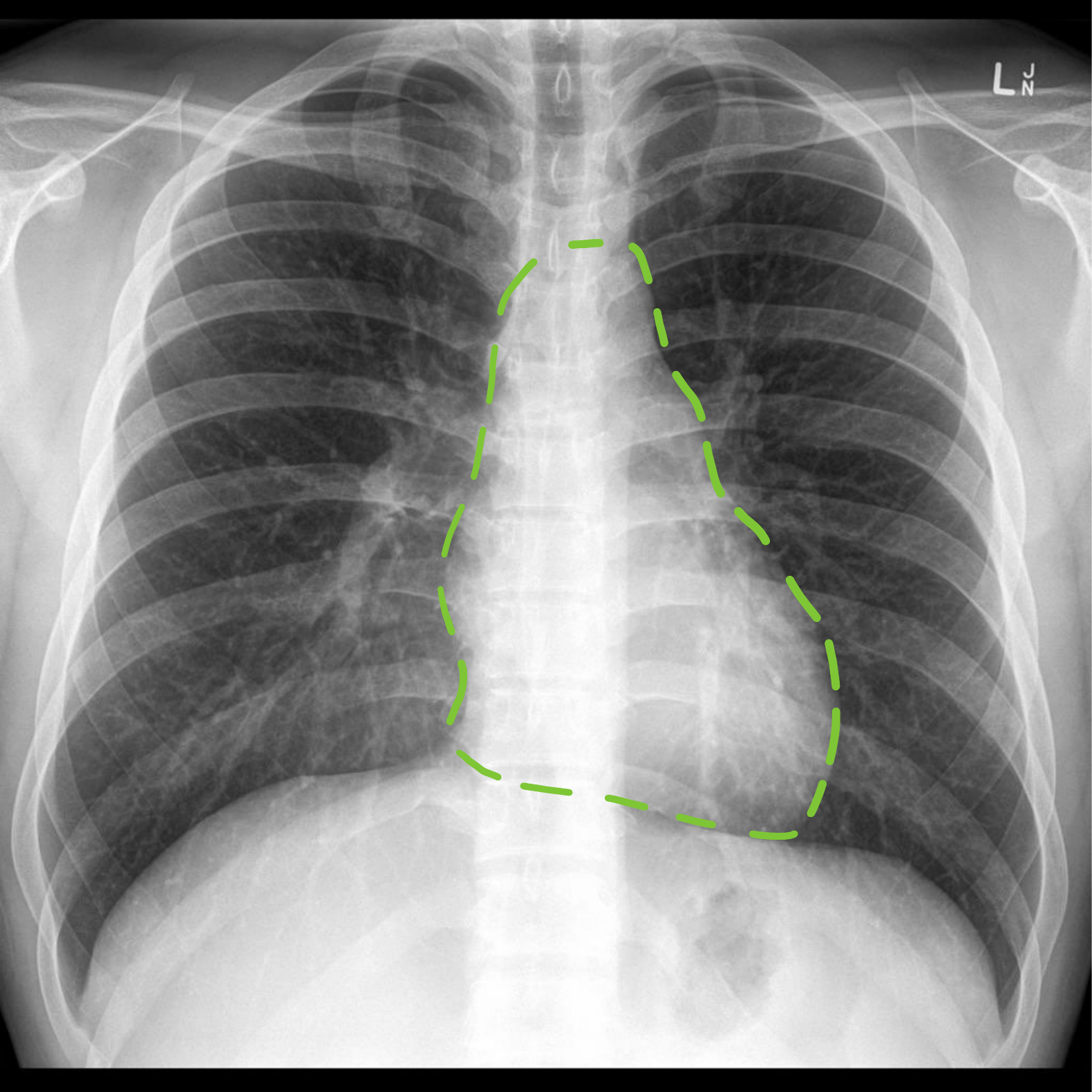}
    \caption{Heart}
\end{subfigure}
\begin{subfigure}[b]{0.325\linewidth}
    \includegraphics[width=\linewidth]{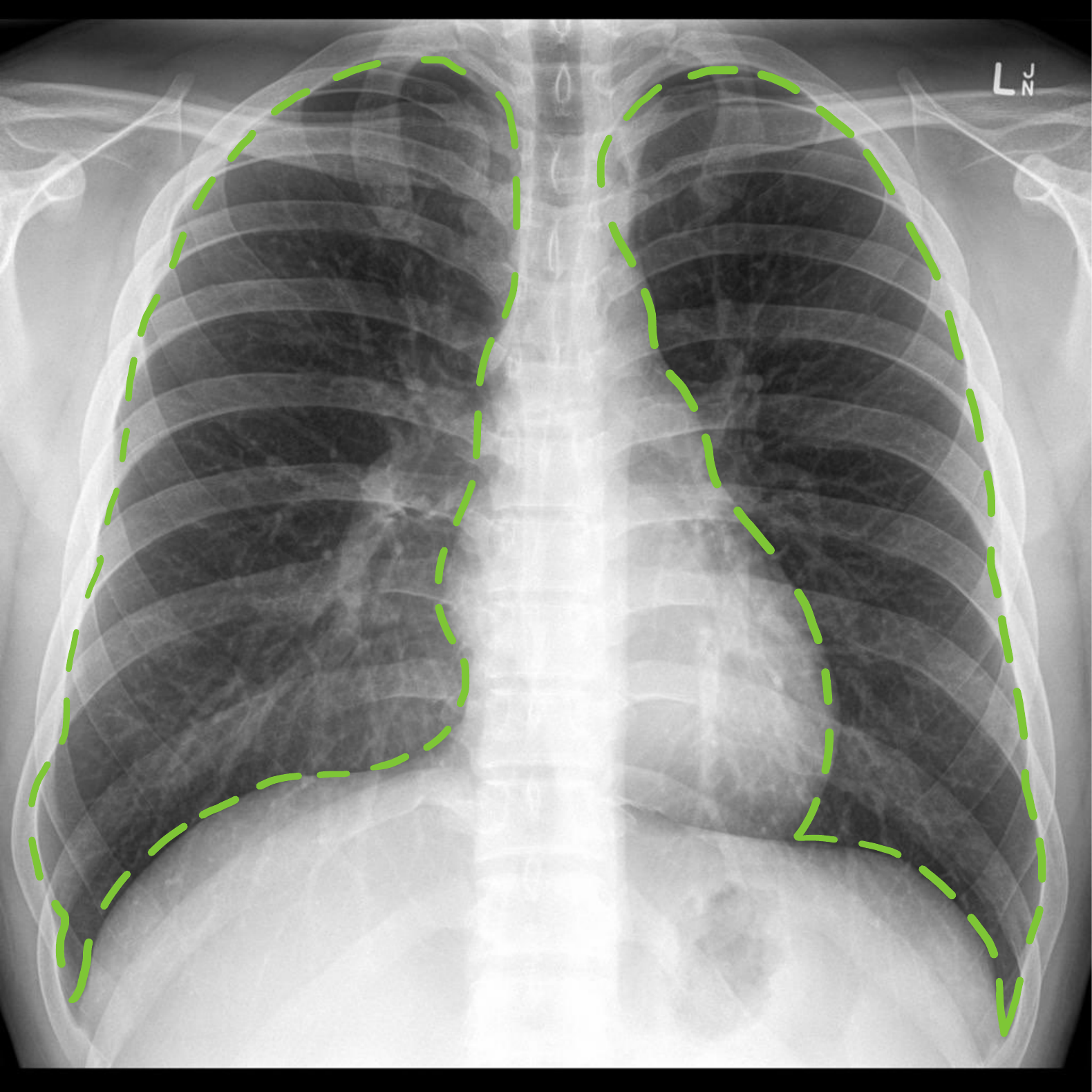}
    \caption{Lungs}
\end{subfigure}
\begin{subfigure}[b]{0.325\linewidth}
    \includegraphics[width=\linewidth]{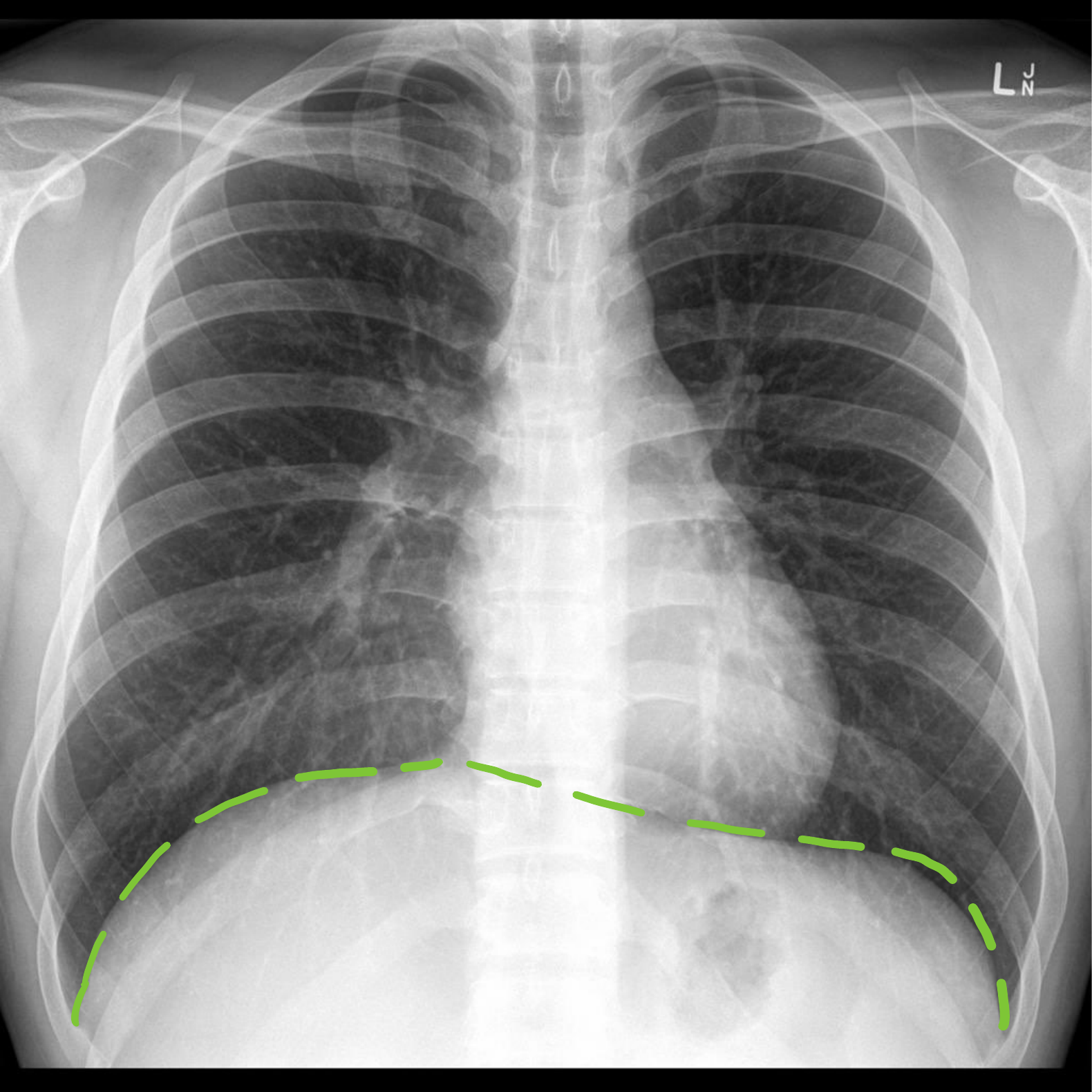}
    \caption{Liver and stomach}
\end{subfigure}
\caption{\chest{} annotations.}
\label{fig:glossary_xray}
\end{figure}

In Figure~\ref{fig:glossary_xray} we show the main thorax parts used for the analysis of the \methName{} traversals. We also remark that left and right are intended from the patient's viewpoint, which is reversed with respect to that of the reader. 

\newpage
\section{PlantVillage traversals}
\label{sec:leavestraversals}
\begin{figure}[h!]
\centering
\begin{subfigure}[b]{0.325\linewidth}
\centering
    \includegraphics[width=\linewidth]{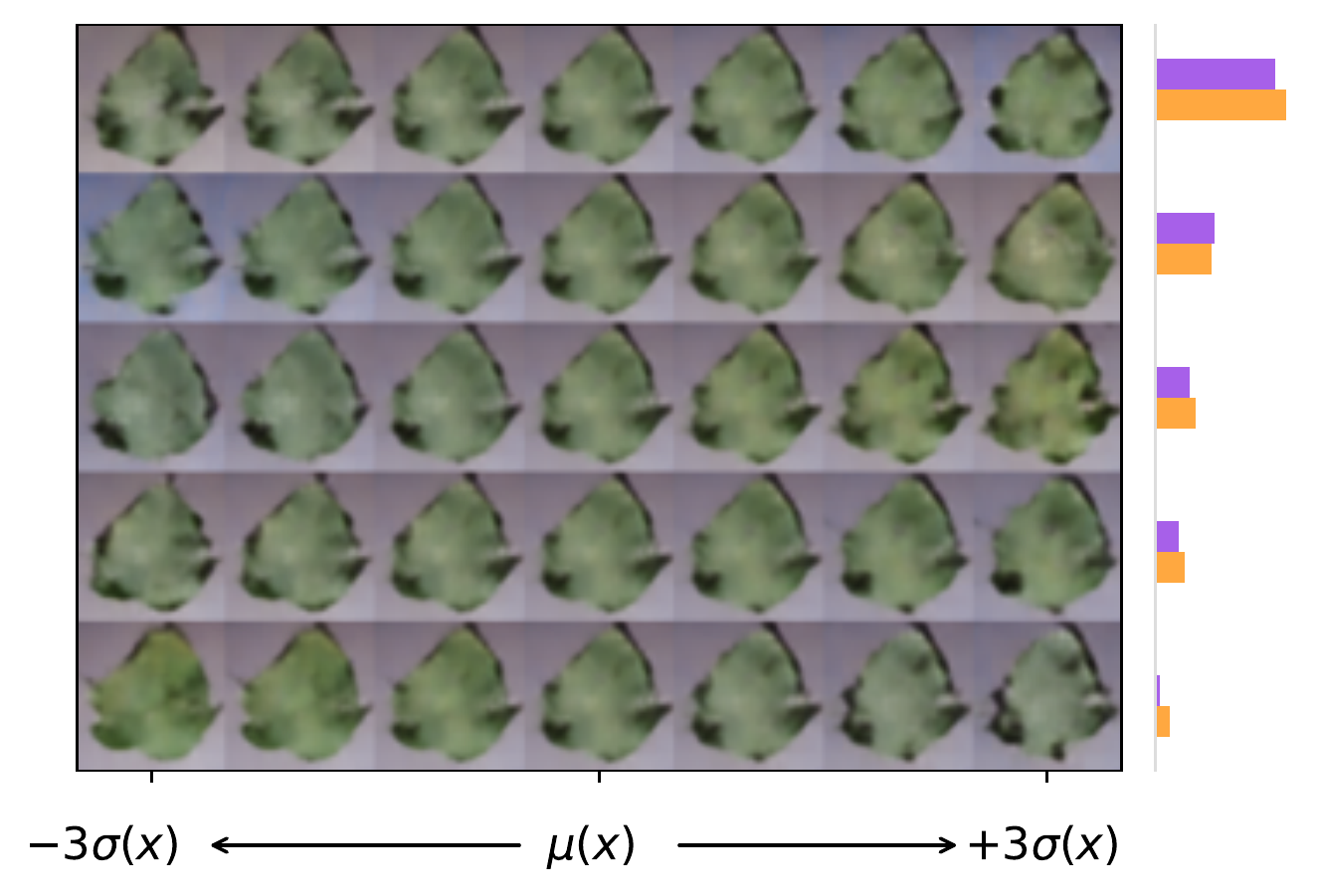}
    \end{subfigure}
\begin{subfigure}[b]{0.325\linewidth}
\centering
    \includegraphics[width=\linewidth]{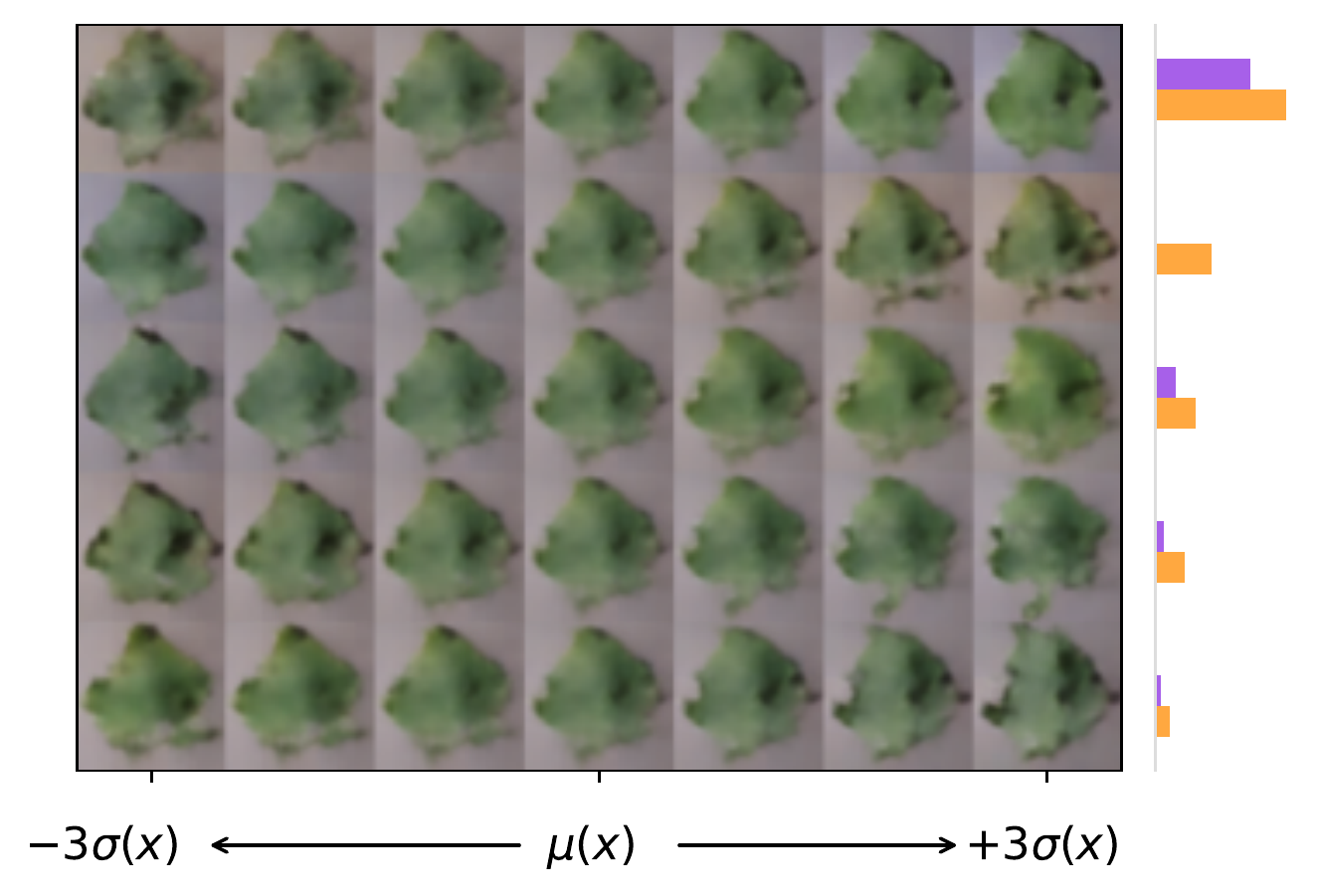}
    \end{subfigure}
\begin{subfigure}[b]{0.325\linewidth}
\centering
    \includegraphics[width=\linewidth]{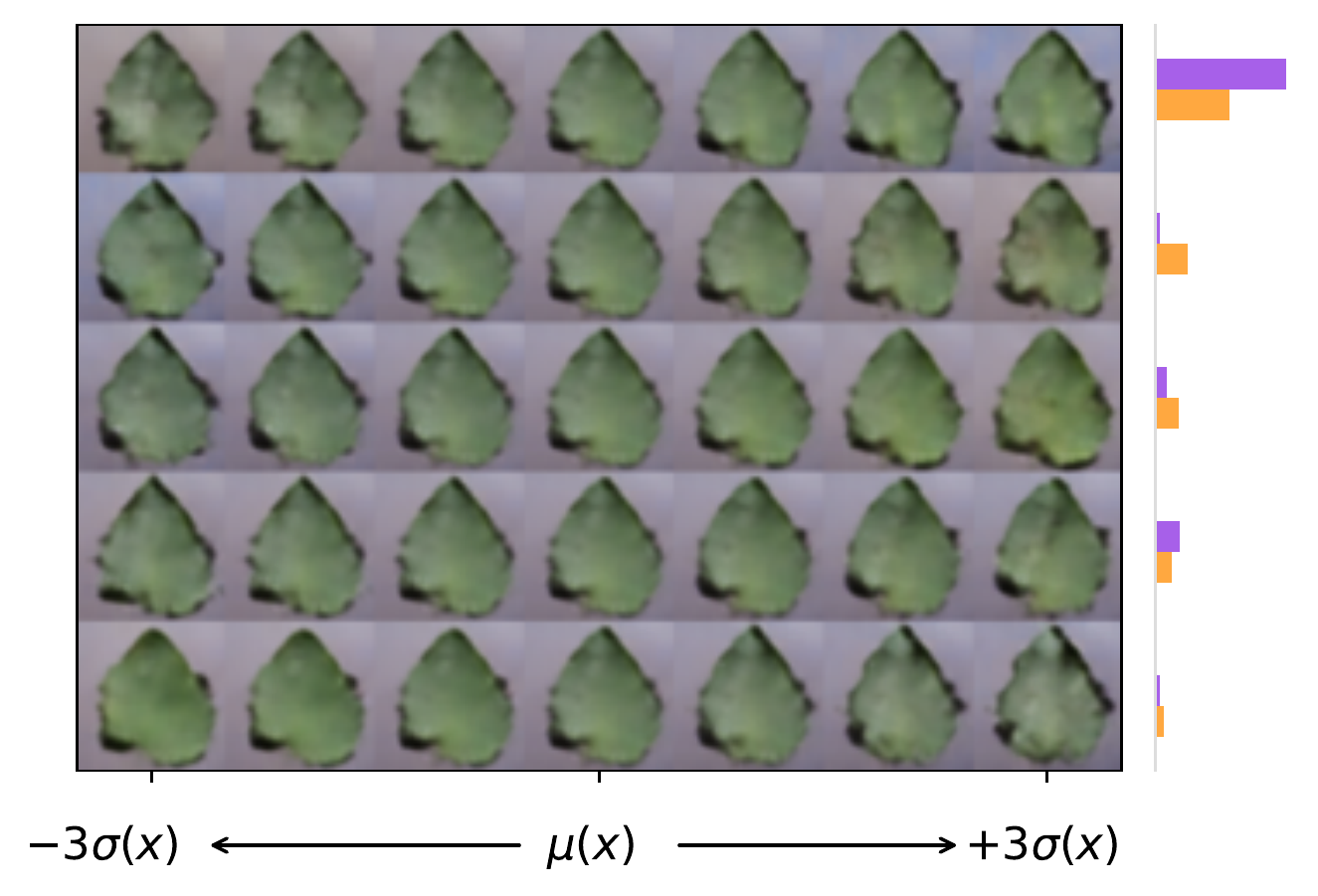}
    \end{subfigure}
\begin{subfigure}[b]{0.325\linewidth}
\centering
    \includegraphics[width=\linewidth]{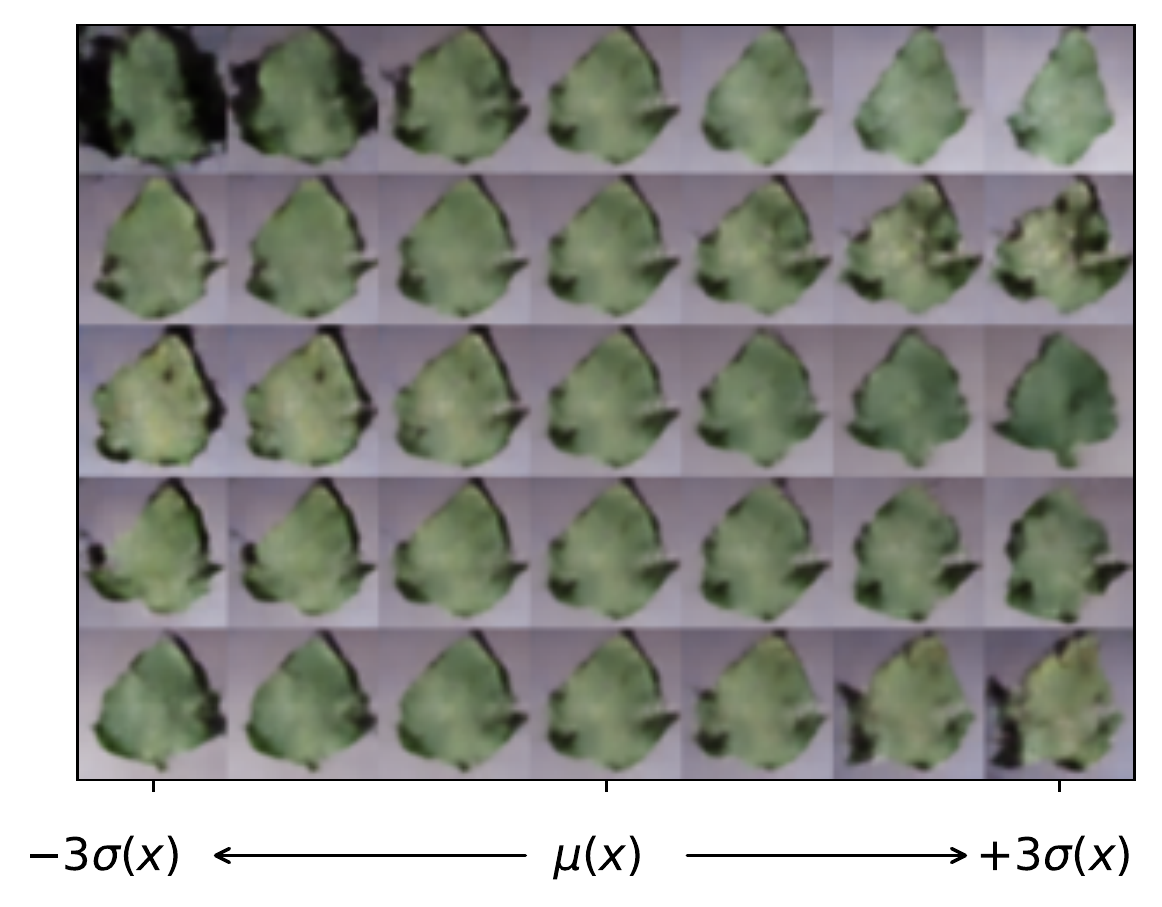}
    \end{subfigure}
\begin{subfigure}[b]{0.325\linewidth}
\centering
    \includegraphics[width=\linewidth]{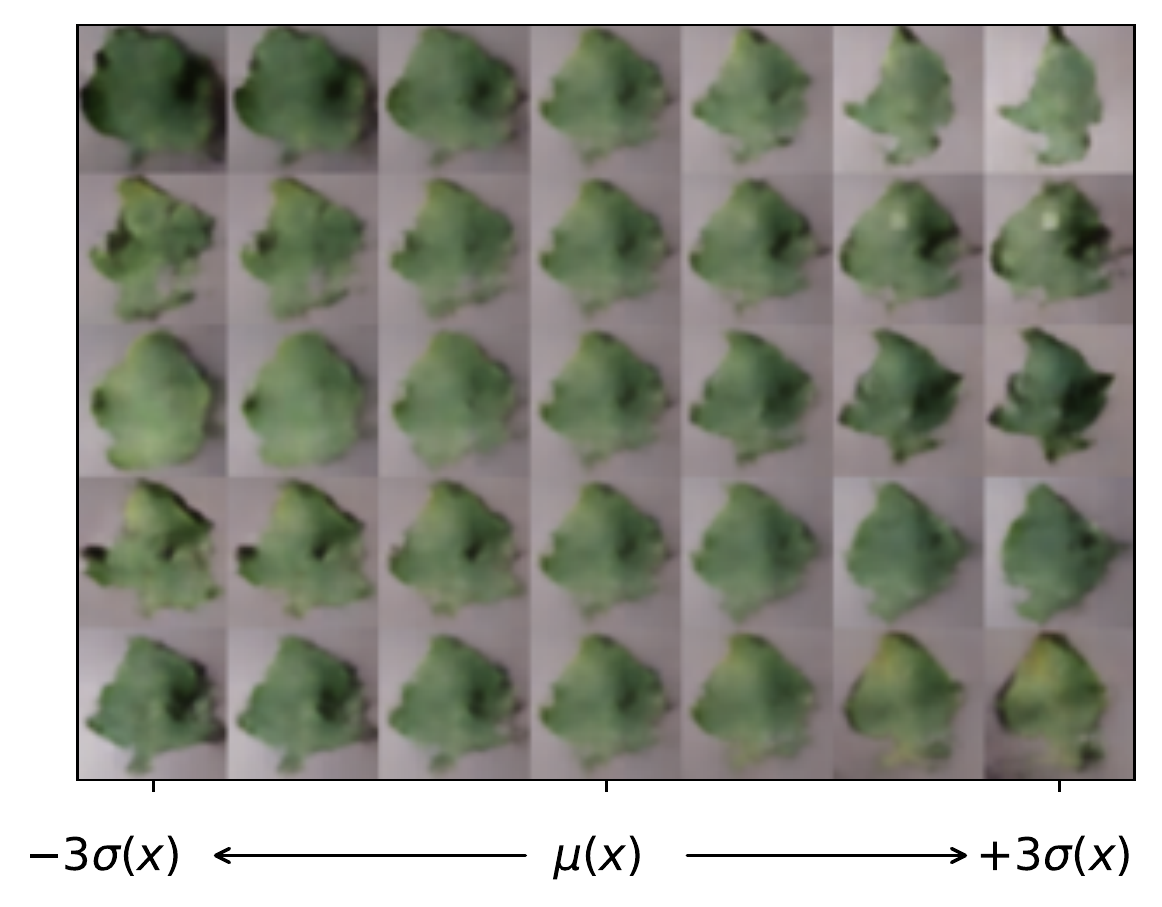}
    \end{subfigure}
\begin{subfigure}[b]{0.325\linewidth}
\centering
    \includegraphics[width=\linewidth]{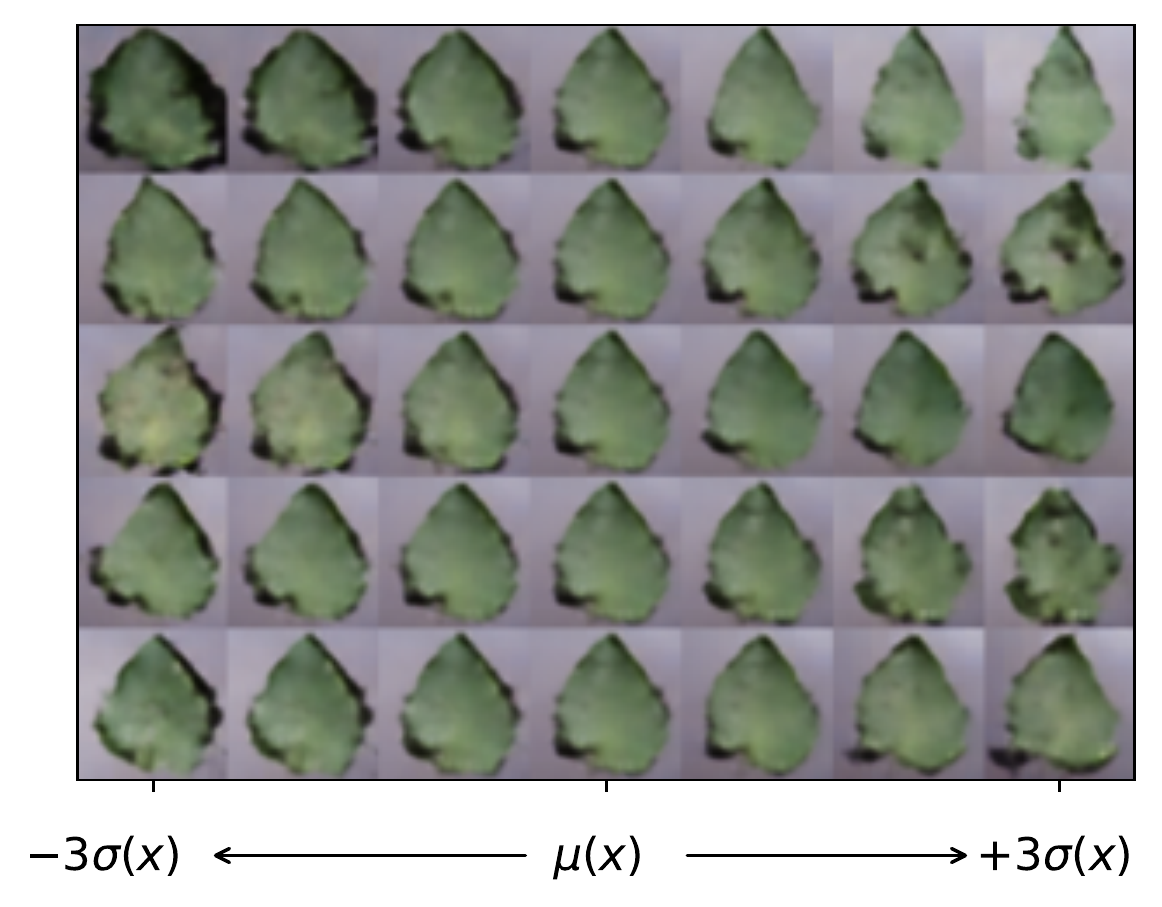}
\end{subfigure}
\caption{\methName{} traversals of $\zc$ (first row) and $\zs$ on the PlantVillage dataset.}
\label{fig:traversals_leaves_clap_supp}
\end{figure}

\begin{figure}[h!]
\centering
    \includegraphics[width=1\textwidth]{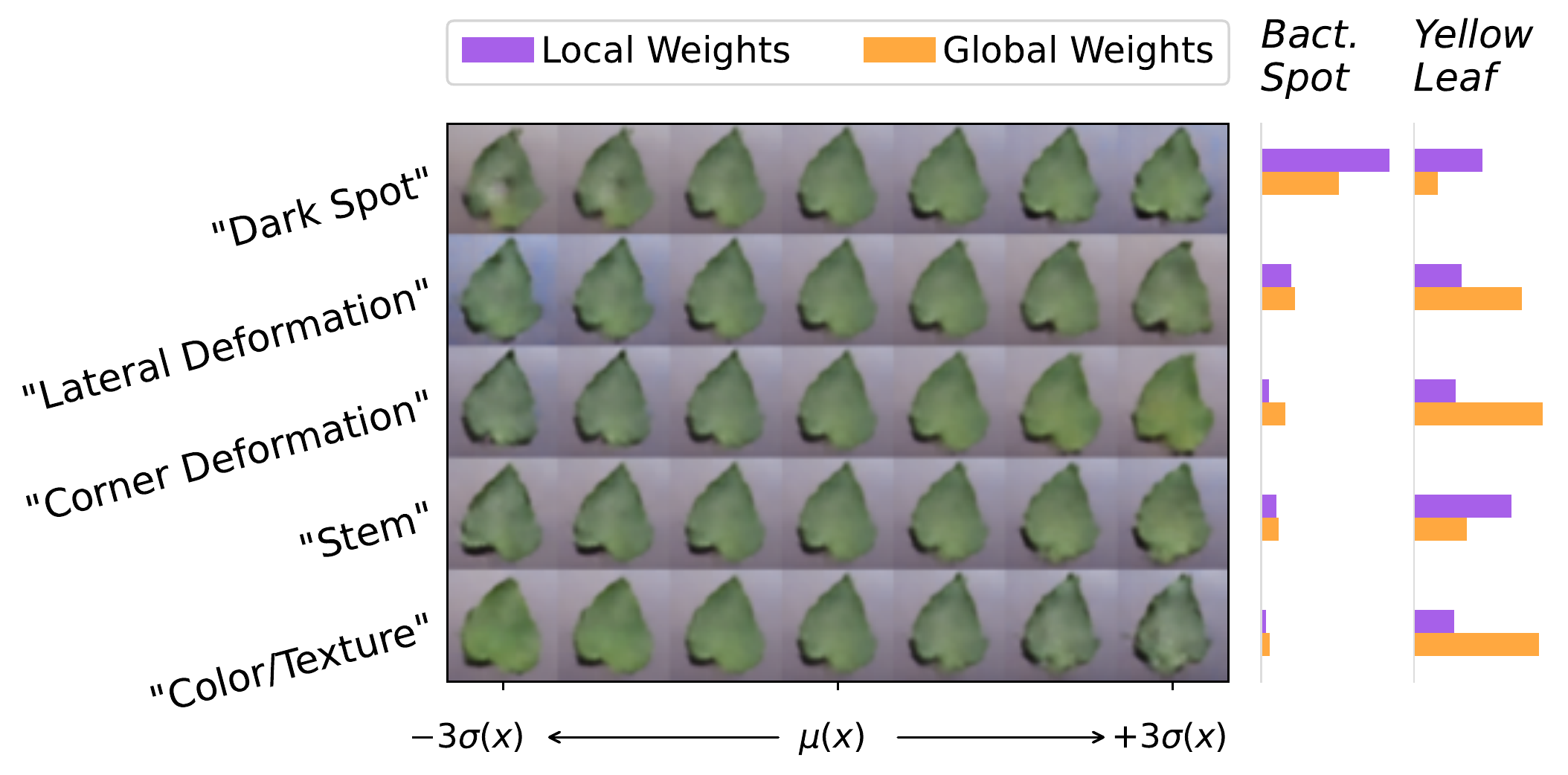}
    \caption{Traversals and human intepretations of \methName{} on the PlantVillage dataset. We include prediction weights for the \emph{Bacterial Spot} and \emph{Yellow Leaf} diseases.}
    \label{fig:clap_leaves_big}
\end{figure}

In Figure~\ref{fig:traversals_leaves_clap_supp} we present traversals of \methName{} for $\zc$ (first row) and $\zs$ (second row) on the PlantVillage dataset \citep{leafdisease}. The dataset includes various plants' leaves, we utilize the 10 binary labels (9 diseases plus healthy or not) of the tomato leaves. We include a magnified figure with human interpretations in Figure~\ref{fig:clap_leaves_big}.

\end{document}